\documentclass[10pt,twocolumn,letterpaper]{article}

\usepackage{iccv}

\usepackage{times}
\usepackage{epsfig}
\usepackage{graphicx}
\usepackage{amsmath}
\usepackage{amssymb}
\usepackage{multicol}
\usepackage{caption}
\usepackage{diagbox}
\usepackage{adjustbox}
\usepackage{rotating}
\usepackage{subcaption}
\usepackage{makecell}
\usepackage{rotating}
\usepackage{multirow}
\usepackage{pdfpages}
\usepackage{array,multirow,graphicx}
\newcommand{\mathkomma}{\quad ,}

\newcommand{\snx}[1]{\textit{SalsaNext }{#1}}

\newcommand{\ls}[1]{\textit{Lov\'{a}sz-Softmax }{#1}}

\newcommand{\sk}[1]{Semantic-KITTI {#1}}  

\def\eg{e.g.\@\xspace}
\def\ie{i.e.\@\xspace}
\def\aka{a.k.a.\@\xspace}
\usepackage[english]{babel}
\newcommand\scalemath[2]{\scalebox{#1}{\mbox{\ensuremath{\displaystyle #2}}}}
\usepackage[dvipsnames]{xcolor}
\usepackage[pagebackref=true,breaklinks=true,letterpaper=true,colorlinks,bookmarks=false]{hyperref}

\iccvfinalcopy 

\def\iccvPaperID{8871} 

\ificcvfinal\fi

\begin{document}

\title{Semantics-aware Multi-modal  Domain Translation: \\ From LiDAR Point Clouds to Panoramic Color Images}

\author{Tiago Cortinhal\\
Halmstad University\\
Sweden\\
{\tt\small tiago.cortinhal@hh.se}
%
\and
 Fatih Kurnaz \\
 Middle East Technical Univetsity\\
  Turkey\\
{\tt\small fatih@ceng.metu.edu.tr}
\and
 Eren Erdal Aksoy \\
Halmstad University\\
Sweden\\
{\tt\small eren.aksoy@hh.se}
}
\makeatletter
\let\@oldmaketitle\@maketitle
\renewcommand{\@maketitle}{\@oldmaketitle
\centering 
  \includegraphics[width=0.85\linewidth]
   {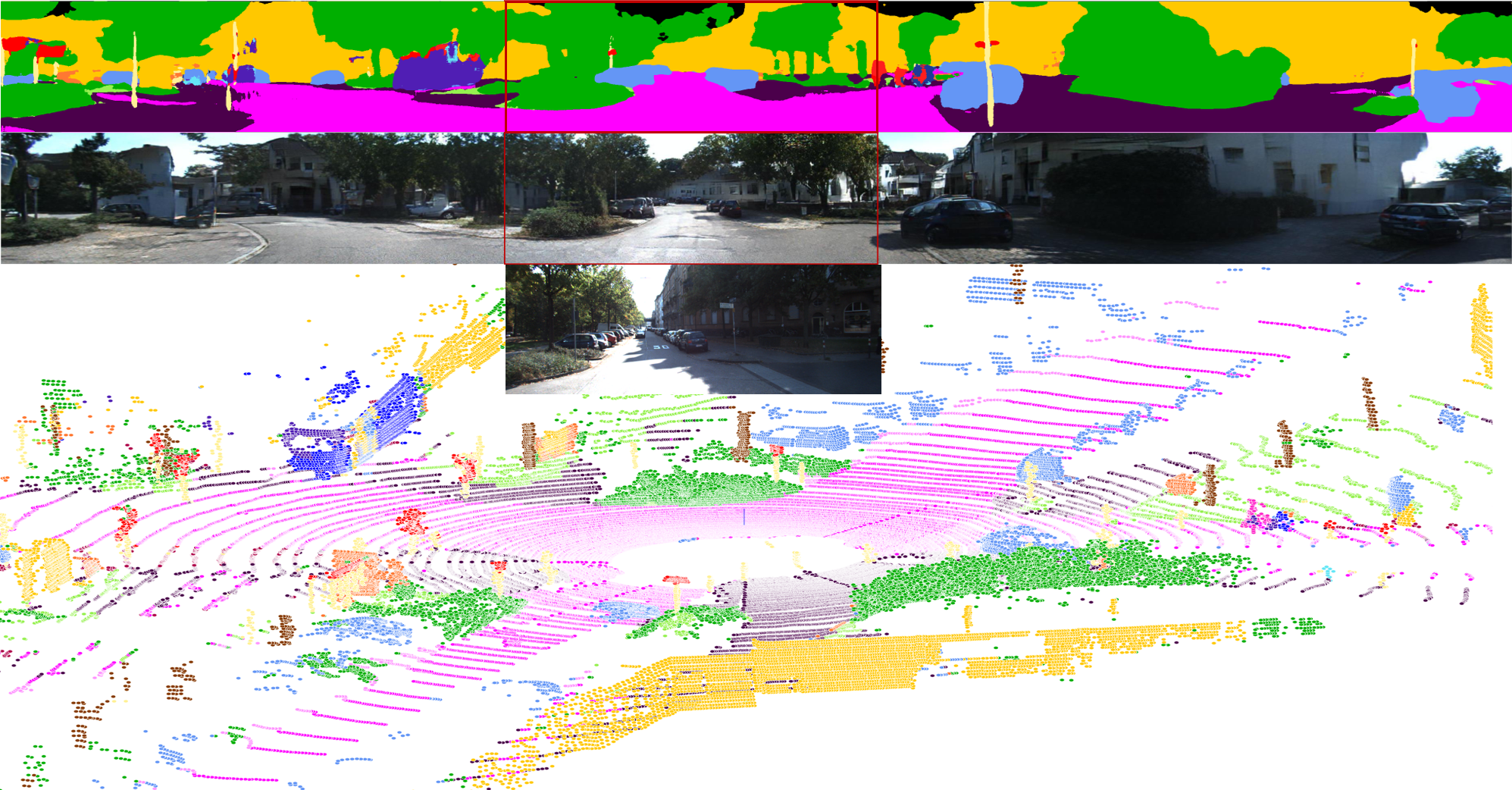}
   \captionof{figure}{We propose a modular generative neural network framework that receives a full 3D LiDAR point cloud and returns the  panoramic color image by solely relying on the semantics of the scene. The framework first applies semantic segmentation to the full LiDAR scan (the bottom image). Next, a novel generative network translates the LiDAR segments to the  camera semantic segments (the top image), which are then converted back to the panoramic color images (the second image from the top) by an additional generative model. The red frame indicates the region that the ground-truth camera image (the third image from the top) corresponds to.  Our framework, for the first time, generates a 360 degree color image   of the environment. }\bigskip
    \label{fig:introSample} 
   }
  
\makeatother

\maketitle
\ificcvfinal\thispagestyle{empty}\fi

\begin{abstract}
 
In this work, we present a simple yet effective framework to address the domain translation problem between different sensor modalities with unique data formats. By relying only on the semantics of the scene, our modular generative  framework can, for the first time, synthesize a panoramic color image  from a given full 3D LiDAR point cloud.

The framework starts with semantic segmentation of the point cloud, which is initially projected onto a spherical surface.
The same semantic segmentation is applied to the corresponding camera image. Next, our new conditional generative model adversarially learns to translate the predicted LiDAR segment maps to the camera image counterparts. Finally, generated image segments are processed to render the panoramic scene images. 
We provide a thorough quantitative evaluation on the \sk dataset~\cite{semantickitti} and show that our proposed framework outperforms  other strong   baseline models. 
Our source code is available.
\end{abstract}

\section{Introduction}


 
Domain translation can be considered a mapping of data samples from an input source domain to a different target domain. 
In computer vision and robotics, this subject has been vastly investigated to convert perceptual readings from one domain to another. For instance, translating sketches to images   or segmentation maps to images,  to name a few.   

Although there   exists an extensive literature on these kinds of image-to-image translations \cite{pix2pix2017,CycleGAN2017,Wang2017},  recent works also focus on  the multi-modal domain translation  such as synthesizing images from raw 3D   point sets \cite{Milz2019Points2Pix3P, AtienzaCVPR2019, Peters2020ConditionalAN}.  
The latter remains, unlike the former, relatively underexplored since point clouds, \eg LiDAR scans, are sparse, unstructured, and nonuniformly sampled, which makes the mapping to the structured image  space non-trivial. 

Multi-modal domain translation has practical uses, in particular for  autonomous vehicles.
Take an example of having a failure in the camera setup. The lack of a modality can severely impair the autonomous vehicle's performance since the subsequent sensor fusion  and manoeuvre planning processes solely rely on these visual readings. 
Therefore, synthesizing photo-realistic images from other functioning modality readings, \eg 3D LiDAR   clouds,  could help overcome a scenario of complete collapse.
Another application   could be generating additional annotated data in the source domain. By transferring the known labels across different domains, one can generate a new variation of the original scene from a different data  distribution  with no extra effort.

With this motivation, we propose a novel multi-modal domain translation framework leveraging the underlying semantics of the perceived scene.   
Differently from existing works~\cite{Milz2019Points2Pix3P,Kim_2019,Kim2020ColorIG}, we argue that mediating the translation between perceptually different  sensor readings  via semantic scene segments could ease the process to a great extent.

\begin{figure*}
  \includegraphics[width=\linewidth]
   {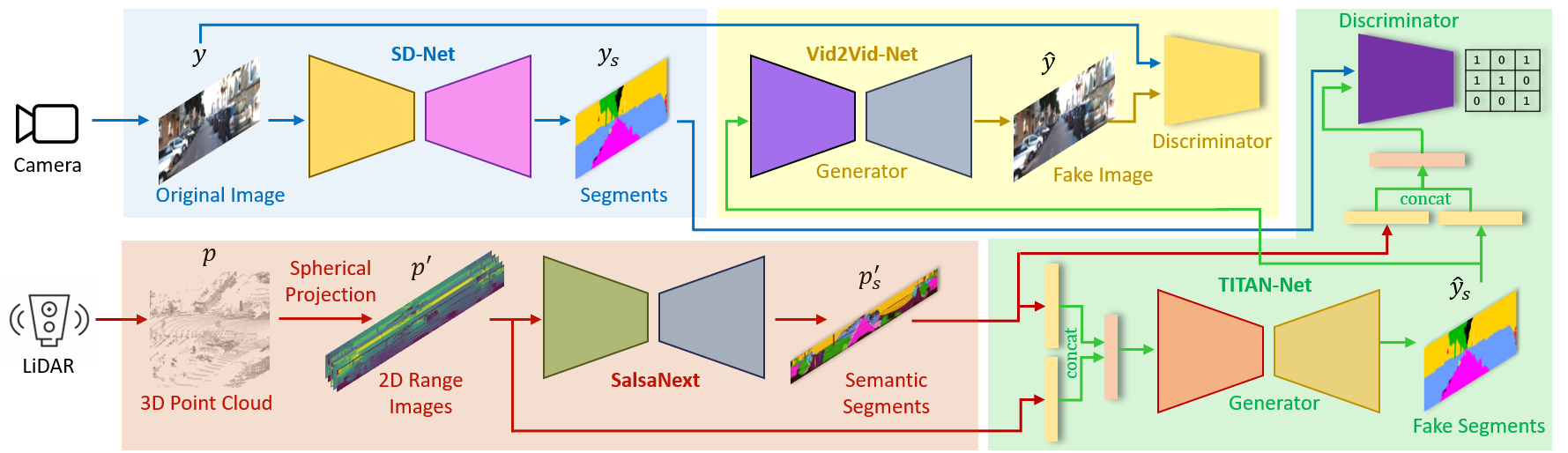}
   \caption{%
   Our proposed modular framework has four neural networks. Each is depicted by a unique background color. 
   In the red box, a captured 3D LiDAR point cloud $p$ is first projected onto the 2D range image plane $p^{\prime}$ to be further processed by   \snx~\cite{cortinhal2020salsanext} to predict semantic segments $p^{\prime}_{s}$.
   Likewise, the corresponding RGB camera image $y$ is processed by SD-Net~\cite{tao2020hierarchical}  to predict semantic segment maps $y_s$ as depicted by the blue box. 
   The green box highlights our proposed cGAN model TITAN-Net, where the \textit{Generator}  is conditioned on the concatenated $p^{\prime}_{s}$ and $p^{\prime}$ to generate the fake camera segment map   $\widehat{y}_{s}$. The TITAN-Net \textit{Discriminator} is also conditioned on  $p^{\prime}_{s}$ while comparing          
   $\widehat{y}_{s}$ with the expected $y_s$. 
   Finally, as depicted in the yellow box, the  fake segment  $\widehat{y}_{s}$ is processed by Vid2Vid-Net~\cite{VidToVid2018} to  synthesize  the realistic RGB image  $\widehat{y}$.  
}
   \label{fig:overview}
\end{figure*}

\textbf{Our contribution:} More specifically, we propose a modular generative framework  that  can, for the first time, synthesize a panoramic color image  from a   full 3D LiDAR scan. See Fig.~\ref{fig:introSample} for example. 
The framework, as shown in Fig.~\ref{fig:overview}, starts with \snx~\cite{cortinhal2020salsanext}: an off-the-shelf state-of-the-art model to semantically segment  the point cloud, which is initially projected onto a spherical surface.
The same semantic segmentation is applied to the paired camera image by employing another state-of-the-art model: SD-Net~\cite{tao2020hierarchical}.
%
As our main technical contribution, we introduce a new conditional generative model, named TITAN-Net (generaTive domaIn TrANslation Network), which adversarially learns to translate the predicted LiDAR segment maps to the camera image counterparts. 
Finally, generated image segments are processed to render the panoramic scene images by a state-of-the-art model. 
%

To the best of our knowledge, our framework is the first approach that relies on sensor-independent semantic context information to achieve semantically consistent translation between multi-modal   domains. 
This opens a rich new vein of opportunity. We can handle  possible camera failures by, for instance,  rendering a raw 3D point set into an image. Without any additional effort, we can further generate realistically looking variants of the scene image from the very same input point cloud. Hence, the available image datasets can simply be augmented with no extra cost.    
%

We provide extensive quantitative and qualitative evaluations of our framework.    
Obtained results on the \sk dataset~\cite{semantickitti} show that our framework  outperforms all   evaluated strong baselines by a large margin.

%




\section{Related Work}

Although there is a large corpus of work on scene image generation~\cite{VidToVid2018, nvidia_vid2vid, surfelgan} and point cloud rendering \cite{Ouyang2017ACS, Milz2019Points2Pix3P, AtienzaCVPR2019, Peters2020ConditionalAN, Kim_2019, Kim2020ColorIG, SongECCV2020},  studies that combine both are lackluster. 

\subsection{Image-to-Image Translation} 
 
Translating a   scene image from one domain to another can be challenging but rewarding. Various methods  \cite{pix2pix2017,CycleGAN2017,Wang2017} showed promising results in image to image translation.
Some works \cite{slmarchildon2021,Roy2019,Tomei_2019_CVPR,mo2018} also leveraged the image semantics to address the image translation and
domain transfer problems.
The same applies to image sequences where temporal cues need to be considered. For instance, \cite{VidToVid2018} created the \textit{vid2vid} network consisting of a carefully designed conditional Generative Adversarial Network  (cGAN) with a spatio-temporal learning objective. The network can create temporally consistent and photorealistic image sequences. Later the \textit{vid2vid} model was improved in \cite{nvidia_vid2vid} by introducing memories of past frames to solve the long-term temporal consistency problem.  

Furthermore, scene generation has also been addressed in the simulation domain.  SurfelGAN~\cite{surfelgan} uses texture-rich surfels  to  generate different trajectories  from the same simulated environment.

\subsection{Point Cloud Rendering}
Most of the recent works converting point clouds to RGB images heavily rely on conditional GANs to force the network to generate more realistic images. In \cite{Ouyang2017ACS} the focus is on generating scene images from upsampled LiDAR data by using a simple cGAN, in which real images were used as the conditions for the cGAN. Other relevant works altered only the conditional part of the GAN. For example, \cite{Milz2019Points2Pix3P} used predefined  background image patches and viewpoint dependent projection to bias the network while generating images in compliance with 3D specifications. As an alternative,  auxiliary conditional GAN was used in \textit{pc2pix}~\cite{AtienzaCVPR2019} to render given a point cloud to its real-life shape from the desired camera angle while skipping the surface reconstruction process entirely. After applying arithmetic operations in the latent space, \textit{pc2pix} can render various newly generated point clouds to images. Lastly, there are works \cite{Peters2020ConditionalAN} which employ cGAN for rendering point clouds without incorporating camera images in the rendering process.

In addition to GANs, other generative models such as asymmetric encoder-decoder network~\cite{Kim_2019} or U-NET~ \cite{Kim2020ColorIG} are also used to generate RGB images directly from point clouds. Similar to them, an end-to-end pipeline was developed in~\cite{SongECCV2020}, which contains a point cloud encoder and RGB decoder in addition to a refinement network to improve photo quality. They can, thus, generate realistic RGB images from the novel viewpoints of a point cloud. 

The closest works to ours are  \cite{Milz2019Points2Pix3P,Kim_2019,Kim2020ColorIG}. 
These approaches, however, neither have the capacity to process the full LiDAR point clouds nor exploit the scene semantics across   modalities with different characteristics. Our framework differs in that the scene semantics acts as a bridge between the full-scan 3D point cloud and 2D image spaces to boost the domain translation.

\section{Method}
 
GANs~\cite{2014GAN} aim to learn a mapping from a noise vector $z$ to a data type $y$, such that the \textit{Generator} ($G$)   learns $G_{z \rightarrow y}: z \rightarrow y$. 
Conditional GANs~\cite{2014arXiv1411.1784M}, on the other hand, condition the generation of data solely on an additional vector  of information  $s$ as $G_{\{z,s\} \rightarrow y}: \{z,s\} \rightarrow y$. 
The additional vector highly depends on the task at hand, but it can hold any feature set (\eg sketches or semantic segments) from an image in a given domain to facilitate domain translation to a vector of classes expected to appear in the generated data. 

Following the works of~\cite{pix2pix2017,2016DeepMultiScale}, we omit the use of the noise vector $z$ as the \textit{Generator} will learn to ignore it and produce deterministic outputs. Nevertheless,  several recent works on conditional generative models already addressed the stochastic data generation \cite{2020StochasticCGAN,2018Diverse}.
We focus only on the domain translation task with an intermediary representation between image and point cloud domains. 
 
Given a LiDAR point cloud $p \in \mathbb{R}^{n \times 4}$, where $n$ represents the number of points and each point has $x,y,z$ coordinates and $i$ intensity values, our goal is to generate an RGB image $y \in \mathbb{R}^{w \times h \times 3}$ with a fixed image size:  $w$ (width) and $h$ (height).
We can define our domain translation  as $G_{\{p,s\} \rightarrow y}: \{p,s\} \rightarrow y$, which is conditioned on the semantic segment maps $s$.

To solve this domain translation problem from $p$ to $y$, we propose a modular framework consisting of four independently trained neural networks. Fig.~\ref{fig:overview} shows the overall framework. Note that our proposed solution is valid for $G_{\{p,s\} \rightarrow y}$, but is yet to be applied to $G_{\{y,s\} \rightarrow p}$. 
In the following, we give a detailed description of each network in the framework.

\subsection{LiDAR Point Cloud Segmentation}
\label{sec:pcseg} 

As depicted in the red box in Fig.~\ref{fig:overview}, our framework starts with the semantic segmentation of 3D LiDAR point clouds.
 
To ease the correspondence problem between the unstructured point cloud $p$ and structured image data $y$, we first apply a spherical projection  \cite{rangenetpp,salsanet2020} to $p$ and create the native LiDAR range view image $p^{\prime}$. In this manner, each  point in $p$ is mapped to an image coordinate $(u,v)$ as:
\begin{equation}
\scalemath{1.0}{
\begin{pmatrix}
\mathit{u}  \\
\mathit{v}
\end{pmatrix}
=
\begin{pmatrix}
\frac{1}{2} [1-\arctan(y,x)\pi^{-1}]\mathit{w^{\prime}}  \\
 [1-(\arcsin(z,r^{-1})+f_{down})f^{-1}]\mathit{h^{\prime}}
\end{pmatrix}
\mathkomma
}
\end{equation}
where $\mathit{w^{\prime}}$ and $\mathit{h^{\prime}}$ denote the width and height    of the projected image, $r$ denotes the range of each point as $r = \sqrt{x^2+y^2+z^2}$ and $f$ the vertical field of view as $f = |f_{down} + | f_{up}|$.
The final output of this transformation will be $p^{\prime} \in \mathbb{R}^{\mathit{w^{\prime}} \times \mathit{h^{\prime}} \times 5}$, i.e. an image $\mathit{w^{\prime}} \times \mathit{h^{\prime}}$ with $(x,y,z,i,r)$ as channels. 
Note that the projection of $p$  to $p^{\prime}$  does not lead to an enormous information loss in the point cloud since the depth information is still kept as an additional channel in $p^{\prime}$.

The projected LiDAR data $\mathit{p^{\prime}}$ is then fed to an off-the-shelf semantic segmentation network  \snx~\cite{cortinhal2020salsanext} which has an encoder-decoder structure extended with an early context module capturing the global context  information. The encoder unit consists of a stack of residual dilated convolution layers fusing receptive fields at various scales.  The decoder part is composed of pixel-shuffle layers, which directly leverage the learned feature maps to  upsample  them with high accuracy and  less computation.  
The final \snx output is a 2D image $p^{\prime}_{s} \in \mathbb{R}^{\mathit{w^{\prime}} \times \mathit{h^{\prime}} }$   storing the predicted point-wise semantic segment  labels.

\subsection{Camera Image Segmentation} 
\label{sec:camseg} 

A similar segmentation treatment is also applied to the paired RGB camera images synchronized with the LiDAR point clouds as highlighted in the blue box in Fig.~\ref{fig:overview}.

For this purpose, we employ  another off-the-shelf state-of-the-art  semantic segmentation network SD-Net~\cite{tao2020hierarchical} which receives the original RGB images $\mathit{y}$ and returns the single-channel segment  maps,  $y_{s} \in \mathbb{R}^{\mathit{w} \times \mathit{h}}$ with a fixed   width ($w$) and height ($h$).
SD-Net has a hierarchical attention architecture and learns to predict attention between adjacent scale pairs. Such multi-scale predictions are then combined at a pixel level to infer the semantic segments. 
SD-Net only operates during training, not in inference. 
   
\subsection{Translation of the LiDAR Semantic Segments} 
\label{titannet}

As highlighted in the green box in Fig.~\ref{fig:overview}, the translation from 3D LiDAR point clouds to RGB camera images is triggered once both synchronized paired modality data, \ie $p$ and $y$, are represented by their corresponding  semantic segments, \ie $p^{\prime}_{s}$ and $y_{s}$. 
To convert the full-scan point cloud segments to their counterparts in the camera image space, we introduce a new semantics-aware conditional GAN model, named TITAN-Net.
  
\textbf{TITAN-Net:} Our cGAN involves two models: \textit{Generator} and \textit{Discriminator}.
The \textit{Generator} architecture is adapted from \snx~\cite{cortinhal2020salsanext} since it is already designed to process range-view projected data, which is compatible with the input that TITAN-Net receives. In addition, \snx  has a lightweight model, thus, exhibits a high runtime performance (reaching up to 24 Hz) which allows fast TITAN-Net training.
As shown in  Fig.~\ref{fig:overview}, the \textit{Generator} receives as input both the range-view projections $p^{\prime}$ and the semantic segmentation maps $p^{\prime}_{s}$ coming from \snx (see Sec.~\ref{sec:pcseg}). 
To exploit both inputs more efficiently, we apply a $1\times 1$ convolution  to each input before concatenating them together. The   merged inputs  then pass through another $1\times 1$ convolutional layer. We further introduce a final upsampling layer to the \snx model to match the actual RGB image dimensions. Finally, the  TITAN-Net \textit{Generator} returns a fake camera image segment map,  $\widehat{y}_{s} \in \mathbb{R}^{\mathit{w} \times \mathit{h}}$.
Note that  LiDAR and camera have different fields of views of the same scene. To create the pairing between both during the  training of TITAN-Net, we restrain the LiDAR projection to the approximate area corresponding to the scene in the camera image.  The full range view image will be only used during the inference to create the panoramic   images.

The TITAN-Net \textit{Discriminator} network is based on the Pix2Pix \textit{Discriminator}, commonly known as PatchGAN~\cite{pix2pix2017}.  
PatchGAN is an extension of \cite{2016Precomputed} and assumes that the most relevant dependencies in an image are present at the patch level, usually called as Markov Random Fields (MRF). 
PatchGAN acts as a patch-wise classifier and outputs a 2D array corresponding to the image patches. 
This allows  us to compute the \textit{Discriminator} loss related to each region, assuming each non-overlapping area is independent. 
To compare with the \textit{Generator} output, the   \textit{Discriminator} also receives the output of SD-Net (see Sec.~\ref{sec:camseg}), \ie $y_{s}$, as the expected RGB image segmentation map. 
Like the TITAN-Net \textit{Generator}, the \textit{Discriminator} is also conditioned on the point cloud segments $p^{\prime}_{s}$, and the same concatenation operation is applied before feeding the \textit{Generator} output $\widehat{y}_{s}$ and $p^{\prime}_{s}$  to the \textit{Discriminator}.

\textbf{Loss Function:} The TITAN-Net loss function is a linear combination of the  Wasserstein GAN with Gradient Penalty (WGAN-GP)~\cite{2017ImprovedWassG} ($\mathcal{L}_{wgan-gp}$)   and  \ls~\cite{Berman2018TheLL}  ($\mathcal{L}_{ls}$) losses:
$\mathcal{L} =  \mathcal{L}_{wgan-gp} + \mathcal{L}_{ls} $. 


As shown in~\cite{2017ImprovedWassG}, penalizing the gradient with WGAN-GP  can stabilize  the training procedure to reduce the mode collapse scenarios and ensure that a robust \textit{Discriminator} can still pass relevant information back to the \textit{Generator}. 
The WGAN-GP based \textit{Discriminator} and \textit{Generator} losses are defined as: 

\begin{equation}
\begin{aligned}
\mathcal{L}_{D}^{WGANGP} = \mathbb{E}_{\tilde{x}\sim \mathbb{P}_g}[D(\tilde{x})] - \mathbb{E}_{x\sim \mathbb{P}_r}[D(x)] + \\ 
\lambda \mathbb{E}_{\hat{x}\sim \mathbb{P}_{\hat{x}}}[(|| \nabla_{\hat{x}} D(\hat{x})||_2 -1)^2] ~\mathkomma
\end{aligned}
\end{equation}
\begin{equation}
\begin{aligned}
&\qquad    \mathcal{L}_{G}^{WGANGP} = - \mathbb{E}_{\tilde{x} \sim \mathbb{P}_g}[D(\tilde{x})] \quad ,&
\end{aligned}
\end{equation}
where $\mathcal{P}_r$,$\mathcal{P}_g$,$\mathcal{P}_{\hat{x}}$ represent the real, generated, and sampling probabilities, respectively. 
The sampling probabilities are uniformly sampled along straight lines between $\mathcal{P}_r$ and $\mathcal{P}_g$ as in \cite{2017ImprovedWassG} and correspond to the gradient penalty.

We include the   \ls~\cite{Berman2018TheLL} loss   to directly optimize the Jaccard index, which is the main metric to evaluate the quality of semantic segments (see Sec.~\ref{sec:jacindex}). Thus, during learning, we aim at maximizing the intersection-over-union  score between the predicted and expected segments. 
The term $\mathcal{L}_{ls}$  acts as a guiding loss and  is defined as:
\begin{equation}
\qquad \scalemath{0.63}{
\mathcal{L}_{ls} = \frac{1}{|C|}\sum_{c\in C} \overline{\Delta_{J_c}}(m(c)) ~,~~ and ~~~
m_i(c) = \left\{
\begin{array}{l l}
  1-x_i(c) &   \text{if ~ $c = y_i(c)$    } \\
  x_i(c) &   \text{otherwise}\\
\end{array}
\right.
~,
}
\label{eq:quidingLoss}
\end{equation}
where $|C|$ represents the class number, $\overline{\Delta_{J_c}}$ defines the Lov\'{a}sz extension of the Jaccard index, $x_i(c) \in [0,1]$ and $y_i(c) \in \{-1,1\}$ hold the predicted probability and ground truth label of pixel $i$ for class $c$, respectively.

\subsection{Segment to RGB Image Translation}
As depicted in the yellow box in Fig.~\ref{fig:overview}, our framework finally employs the   camera segments  $\widehat{y}_{s}$ generated by TITAN-Net to  synthesize  realistic RGB images   $\widehat{y}$. 

For this purpose, we employ an off-the-shelf state-of-the-art cGAN model Vid2Vid-Net~\cite{VidToVid2018} which is coupled with spatio-temporal adversarial objectives to generate  temporally consistent and photorealistic image sequences. Consequently, given the generated segment masks $\widehat{y}_{s}$,   Vid2Vid-Net returns the final response $\widehat{y}$ of our modular framework depicted in  Fig.~\ref{fig:overview}.

\section{Experiments}

\subsection{Implementation Details}

Except for Vid2Vid-Net, which is retrained  using the default configurations from the available source code, we use  the publicly available pre-trained weights for all  other models in our proposed framework.

Regarding TITAN-Net, as an optimizer, we use  Adam\cite{Kingma2015AdamAM} with a learning rate of $1\times 10^{-4}$, $(0.5,0.999)$  as $(\beta_1,\beta_2)$.
The batch size and dropout probability  are fixed at $10$ and   $0.2$, respectively.
To avoid overfitting, we perform data augmentation by flipping randomly around the y-axis and randomly dropping points before creating the projection. Both augmentations are applied independently of each other with a probability of $0.5$.

The entire proposed framework is trained with point clouds of size ranging from 10-13k points per scan and images of  size $1241 \times 376$. After applying the spherical projections, we obtain the range view images with the  size of $2048 \times 64 \times 5$ centered on the view of the camera.

Our TITAN-Net model is implemented in PyTorch, and the source code is released for public use~
\href{https://github.com/Halmstad-University/TITAN-NET}{https://github.com/Halmstad-University/TITAN-NET}. 

\subsection{Dataset}
We evaluate the performance of the proposed framework and compare it to the other state-of-the-art domain translation approaches by using the large-scale challenging  \sk dataset~\cite{semantickitti}. There exist over 43K point-wise annotated full 3D LiDAR scans and in total $19$ different classes in the \sk dataset. 
By following  the same protocol introduced in \cite{rangenetpp}, we  divide the dataset  into training  (sequences 00-10), validation (sequence 08), and test splits (sequences 11-21).    

Note that although \snx is already trained on the \sk dataset, the SD-Net model is trained using the Cityscapes   dataset~\cite{Cityscapes} which has fewer classes, \ie $14$, with different  labels. To cope with the incompatibilities between the class numbers and labels in   two   datasets, we define a   mapping table (see the supplementary material) that returns $14$ unique class labels matched in both datasets.


\begin{table*}[!t]
\centering
\resizebox{0.9\hsize}{!}{
\begin{tabular}{l||llllllllllllll|l}
Approach &  \rotatebox{90}{Car} & \rotatebox{90}{Bicycle} & \rotatebox{90}{Motorcycle} & \rotatebox{90}{Truck}    & \rotatebox{90}{Other-Vehicle} & \rotatebox{90}{Person} & \rotatebox{90}{Road} & \rotatebox{90}{Sidewalk} & \rotatebox{90}{Building} & \rotatebox{90}{Fence} & \rotatebox{90}{Vegetation} & \rotatebox{90}{Terrain}  & \rotatebox{90}{Pole} & \rotatebox{90}{Traffic-Sign} & mIoU $\uparrow$ \\ 
\hline
Pix2Pix~\cite{pix2pix2017}           & 8.8                                & 0                                 & 0                                 & 0                                  & 0                               & 0                                 & 57.7                               & 15.7                               & 32.8                               & 12.5                               & 32.7                               & 14.8                               & 0.5                             & 0                                  & 12.5                                  \\
TITAN-Net (Ours) &   \multicolumn{1}{c}{\textbf{68.2 }} & \multicolumn{1}{c}{\textbf{9.9 }} & \multicolumn{1}{c}{\textbf{7.6 }} & \multicolumn{1}{c}{\textbf{7.6 }} & \multicolumn{1}{c}{0\textbf{ }} & \multicolumn{1}{c}{\textbf{7.3 }} & \multicolumn{1}{c}{\textbf{75.4 }} & \multicolumn{1}{c}{\textbf{48.3 }} & \multicolumn{1}{c}{\textbf{62.9 }} & \multicolumn{1}{c}{\textbf{33.6 }} & \multicolumn{1}{c}{\textbf{60.1 }} & \multicolumn{1}{c}{\textbf{49.9 }} & \multicolumn{1}{c}{\textbf{2 }} & \multicolumn{1}{c|}{\textbf{3.7 }} & \textbf{31.1 }                       
\end{tabular}
}
\caption{Quantitative results for the generated semantic segment  images on the test sequences.  
 $\uparrow$ denotes that   higher is better.  }
   \label{tab:gen_sem_seg}
\end{table*}



\begin{table*}[!t]
\centering
\resizebox{\textwidth}{!}{
\begin{tabular}{l||l|l|l|l|l|l|l|l|l|l}
\multicolumn{3}{l}{} & \multicolumn{8}{c}{SWD $\times 10^3$ $\downarrow$} \\ 
\cline{4-11}
Approach & SSIM $\uparrow$ & FID $\downarrow$ & 1024$\times$1024 & 512$\times$512 & 256$\times$256 & 128$\times$128 & 64$\times$64 & 32$\times$32 & 16$\times$16 & avg \\ 
\hline
SC-UNET\textasciitilde{} & 0.3158 & 261.282 & 2.65 & 2.56 & 2.36 & 2.20 & 2.14 & 2.14 & 4.07 & 2.59 \\
Pix2Pix\textasciitilde{} $\rightarrow$ Vid2Vid & 0.2543 & 73.476 & 2.29 & 2.22 & 2.18 & 2.15 & 2.11 & \textbf{2.13} & 3.95 & 2.59 \\
TITAN-Net (Ours) $\rightarrow$ Vid2Vid & 0.2610 & \textbf{61.914} & \textbf{2.22} & \textbf{2.18} & \textbf{2.15} & \textbf{2.11} & \textbf{2.08} & \textbf{2.13} & \textbf{3.78} & \textbf{2.38} \\
Pix2Pix\textasciitilde{} $-$ w/o SegMap & 0.2006 & 209.150 & 2.43 & 2.35 & 2.31 & 2.31 & 2.35 & 2.42 & 4.76 & 2.71 \\
TITAN-Net $-$ w/o SegMap & \textbf{0.3692} & 326.298 & 3.24 & 3.14 & 2.98 & 2.61 & 2.23 & 2.10 & 3.94 & 2.89 \\
TITAN-Net $-$w/o Rangeview $\rightarrow$Vid2Vid & 0.2442 & 76.932 & 2.31 & 2.23 & 2.18 & 2.16 & 2.15 & 2.22 & 4.47 & 2.53 \\ 
\hline\hline
\multicolumn{1}{l}{SD-Net -\textgreater{} Vid2Vid} & \multicolumn{1}{l}{0.4089} & \multicolumn{1}{l}{20.3694} & \multicolumn{1}{l}{2.10} & \multicolumn{1}{l}{1.99} & \multicolumn{1}{l}{1.86} & \multicolumn{1}{l}{1.71} & \multicolumn{1}{l}{1.56} & \multicolumn{1}{l}{1.41} & \multicolumn{1}{l}{1.70} & 1.76
\end{tabular}
 
}
\caption{Quantitative results for the synthesized RGB images using the test sequences. 
Each level of the Laplacian Pyramid corresponds to a given resolution. The distances are shown per level and the average gives us the overall distance between both distributions. Due to the nature of SWD, both images are resized to $1024\times1024$ before calculating the distance.
 $\downarrow$ denotes that lower scores are better and $\uparrow$ that higher is better. 
 }
   \label{tab:gen_rgb_im}
\end{table*}	
 

\subsection{Evaluation Metrics}

We use the following metrics to measure the quality of the generated segment maps and synthesized images:
 
\textbf{Jaccard Index:}
\label{sec:jacindex} 
For the quality evaluation of the predicted segment maps, we use the Jaccard Index, \aka the mean intersection-over-union (mIoU), over all classes. 
A higher mIoU score indicates better segmentation results.

\textbf{Structural Similarity Index Measure (SSIM):}
SSIM is a perception metric that measures image similarity by exploiting three different image components: Luminance, Contrast, and Structure. 
Both images are normalized, and we   compare the covariance of both images. In our evaluations, we use SSIM with a window size of 11 as in the original paper \cite{1284395}.
%
The higher the SSIM value, the better. 

\textbf{Fr\'echet Inception Distance (FID):}
%
%
%
%
%
The lower the FID value, the closer is the generated image to the real counterpart. As described in \cite{GANsTB}, FID is consistent with human judgment. Any noise or artifacts present in the generated image decrease  the FID value. Generally, FID is a reliable metric as it correlates consistently with the visual quality of generated images.

\textbf{Sliced Wasserstein Distance (SWD):}
Another metric allowing us to weigh the distribution of generated   and real images is the Sliced Wasserstein Distance (SWD) proposed in \cite{karras2018progressive}.  This metric assumes that a successful \textit{Generator}   produces images that have structural similarities at different scales.  We  extract image patches from a Laplacian Pyramid~\cite{laplace} starting with a low-pass resolution of 16$\times$16, which is doubled until the desired resolution is reached. 


After the respective normalization w.r.t. the mean and standard deviation, the SID value between both sets of patches is computed following the work in \cite{Rabin2012}. Generally, the lower the SWD value, the better.

\subsection{Baselines}

We compare the performance of our framework to  two different baselines trained on the same dataset using the same training protocol. 

\textbf{Pix2Pix}~\cite{pix2pix2017} is the state-of-the-art generative model for  image-to-image translation. In our framework in Fig.~\ref{fig:overview}, we replace TITAN-Net with Pix2Pix to diagnose  the contribution of our generative model TITAN-Net in  the generated segment maps and synthesized images.

\textbf{SC-UNET}~\cite{Kim2020ColorIG} is a recent generative model based on Selected Connection U-Net (SC-UNET)  specifically designed for generating RGB images directly from point clouds. Unlike our approach, SC-UNET neither incorporates the   segment maps nor involves adversarial training.

\setlength{\columnsep}{0.1cm}
\begin{figure*}[!t]
\begin{multicols}{3}
    \includegraphics[width=1.0\linewidth]{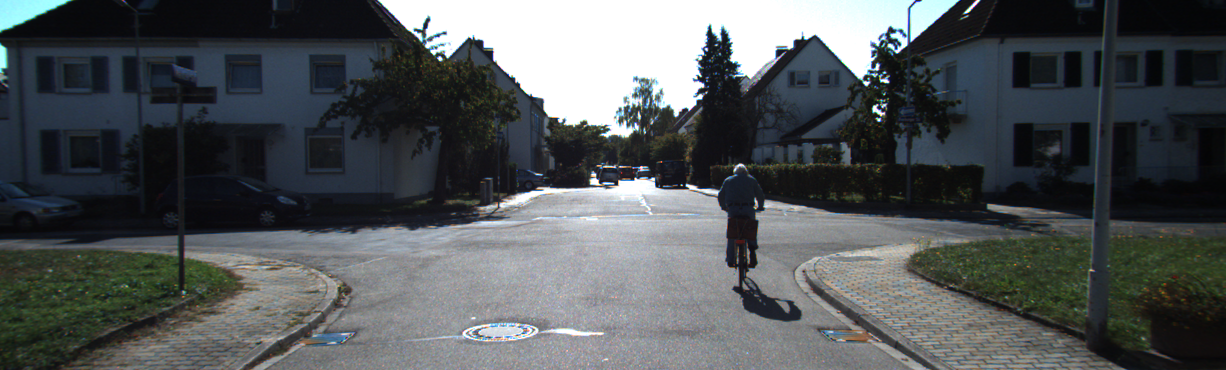} 
    \includegraphics[width=1.0\linewidth]{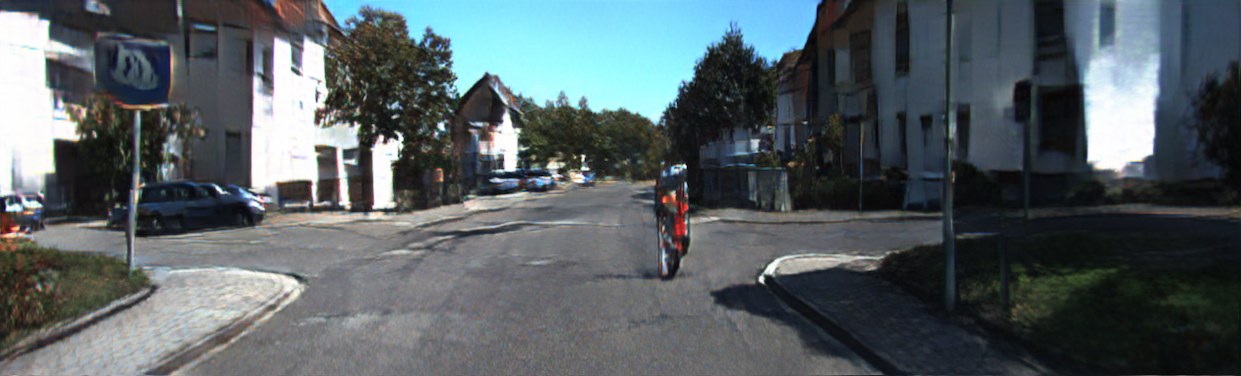}
    \includegraphics[width=1.0\linewidth]{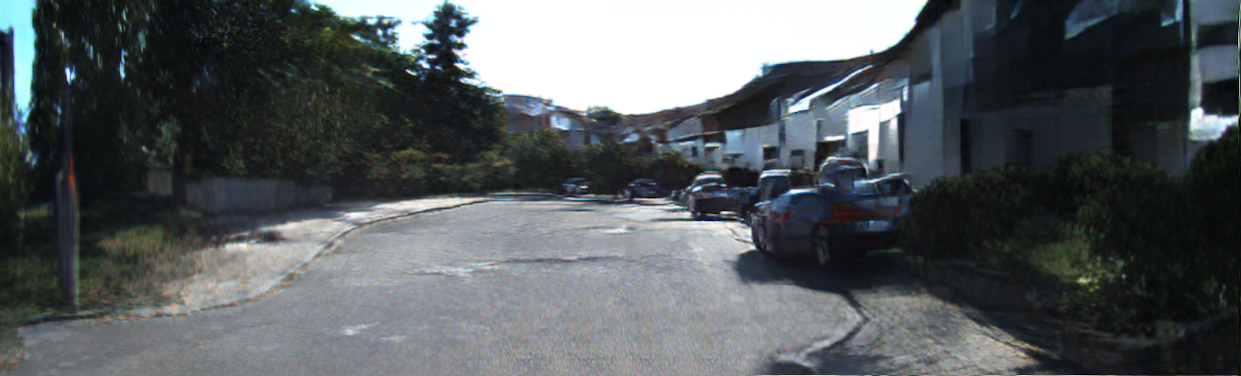}
\end{multicols}
\vspace{-0.95cm}
\begin{multicols}{3}
   \includegraphics[width=1.0\linewidth]{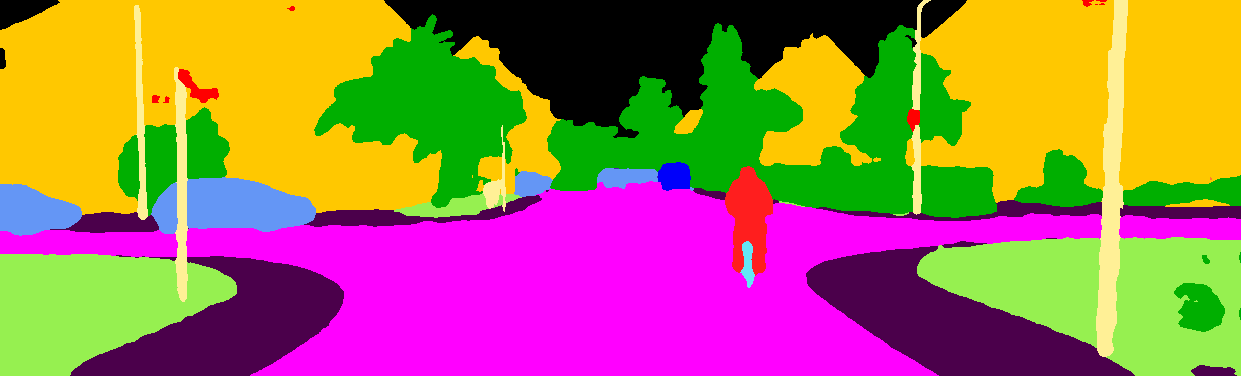} 
    \includegraphics[width=1.0\linewidth]{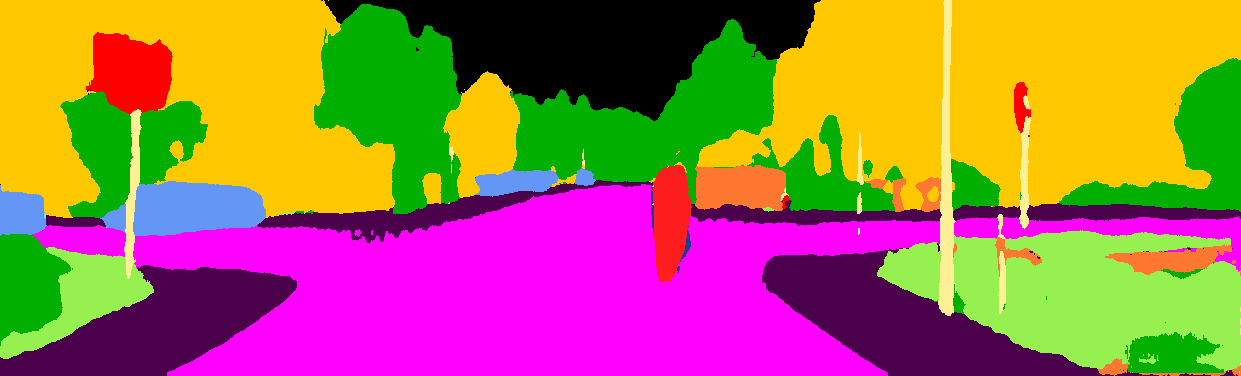}
    \includegraphics[width=1.0\linewidth]{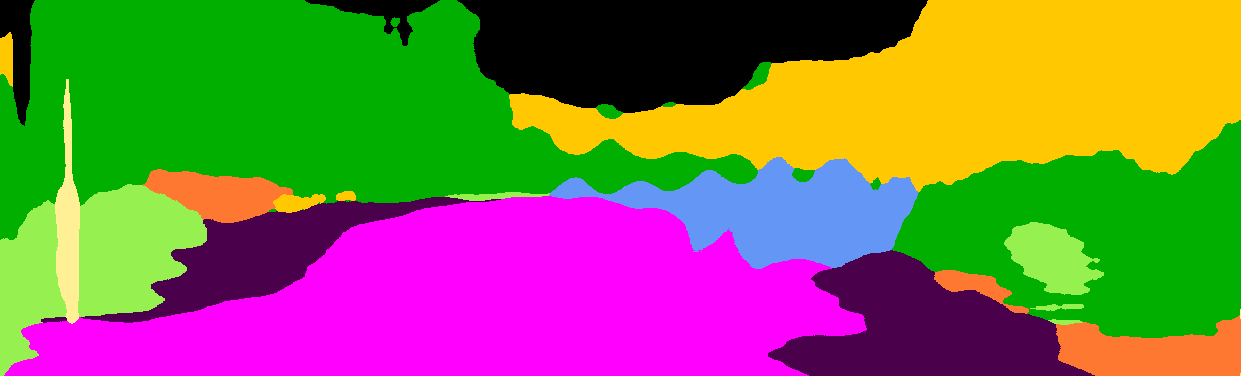}
\end{multicols}

\vspace{-0.8cm}

\begin{multicols}{3}
    \includegraphics[width=1.0\linewidth]{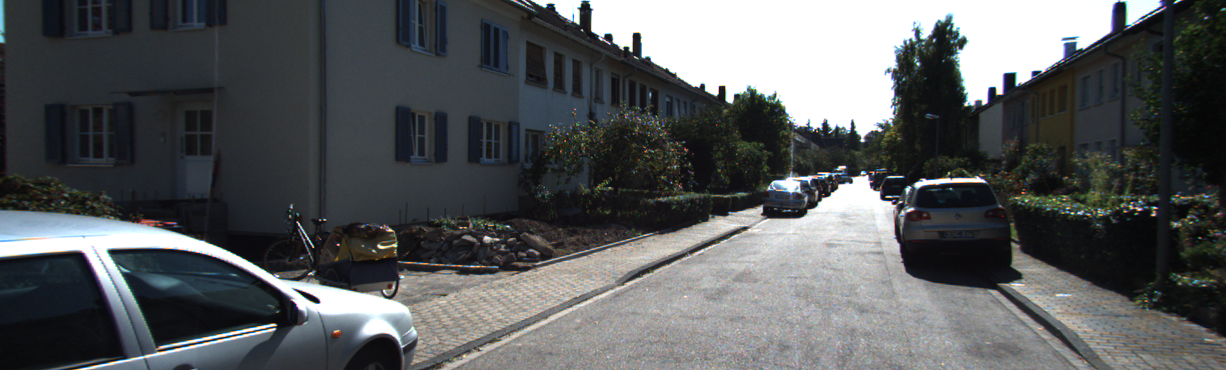} 
    \includegraphics[width=1.0\linewidth]{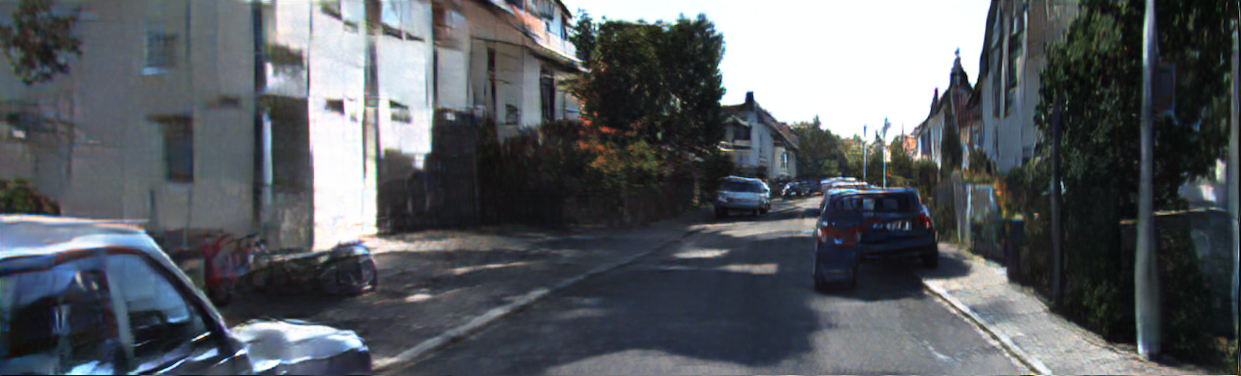}
    \includegraphics[width=1.0\linewidth]{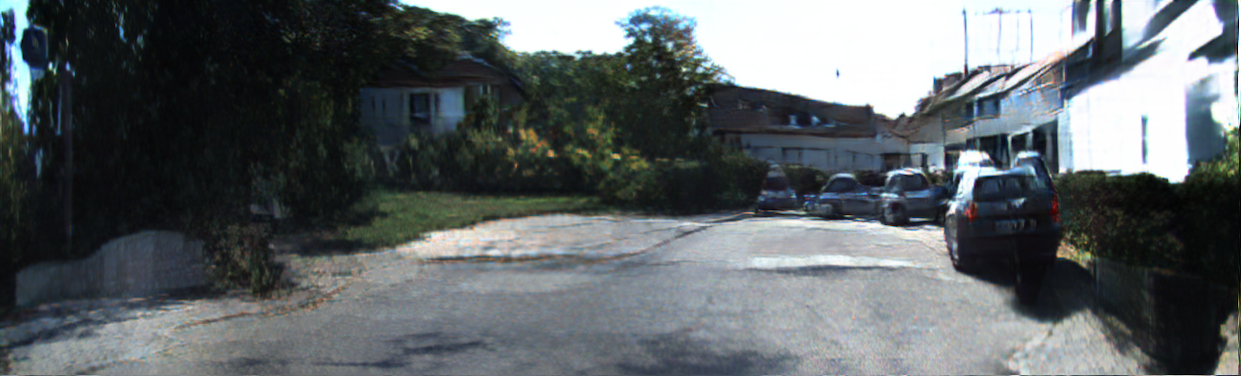}
\end{multicols}
\vspace{-0.95cm}
\begin{multicols}{3}
   \includegraphics[width=1.0\linewidth]{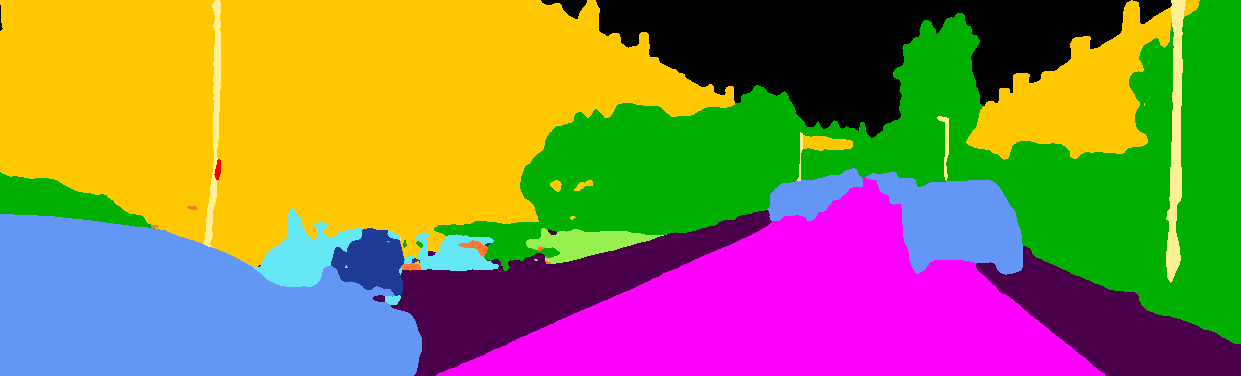} 
    \includegraphics[width=1.0\linewidth]{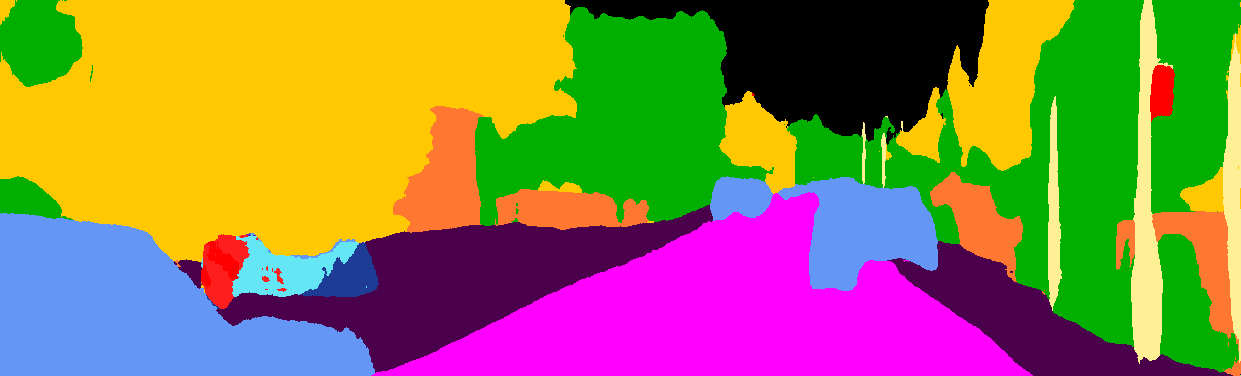}
    \includegraphics[width=1.0\linewidth]{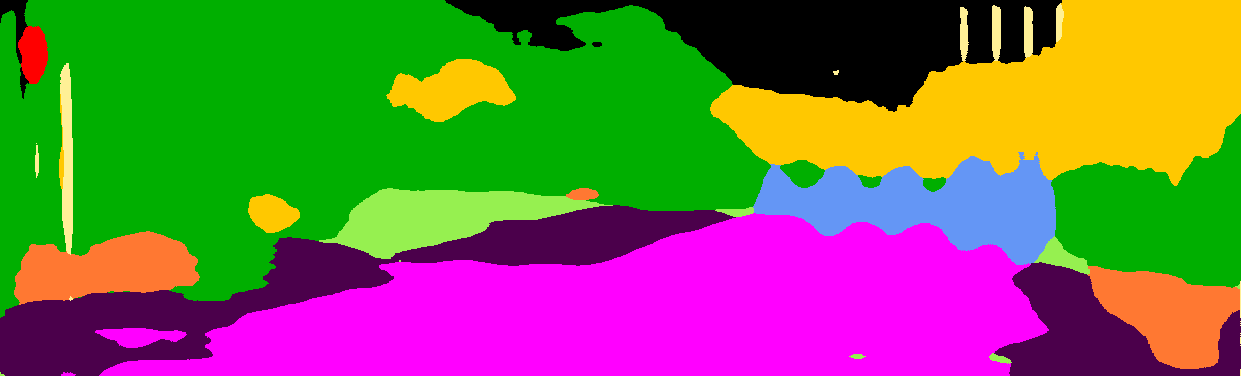}
\end{multicols}
  

\caption{Sample qualitative results are showing the synthesized images at the top together with the corresponding generated   segment maps at the bottom. From left to right, we have the ground-truth images, the TITAN-Net results (Ours)  and the Pix2Pix~\cite{pix2pix2017} outputs. Note that TITAN-Net and Pix2Pix are combined with Vid2Vid to translate segments to RGB images. 
}
\label{fig:synt_img_results}
\end{figure*}


\begin{figure*}[!t]
\begin{multicols}{4}
    \includegraphics[width=1.0\linewidth]{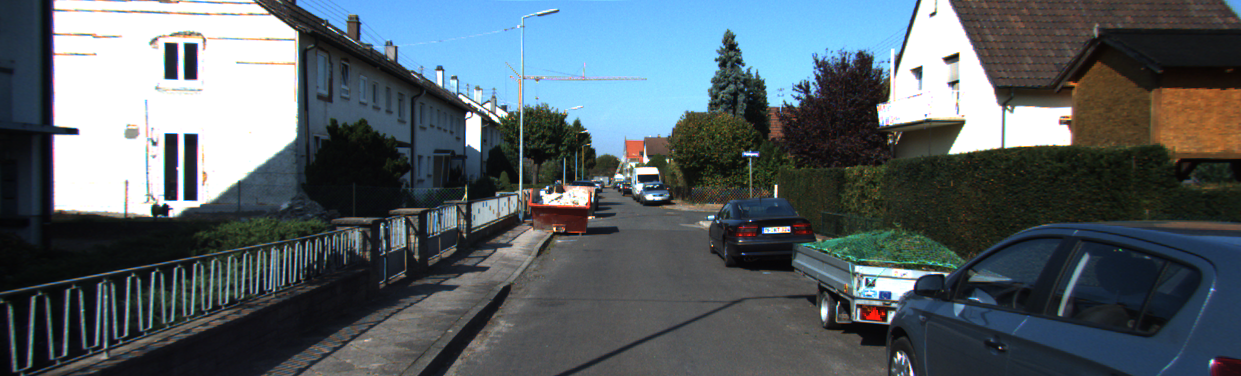}\par 
    \includegraphics[width=1.0\linewidth]{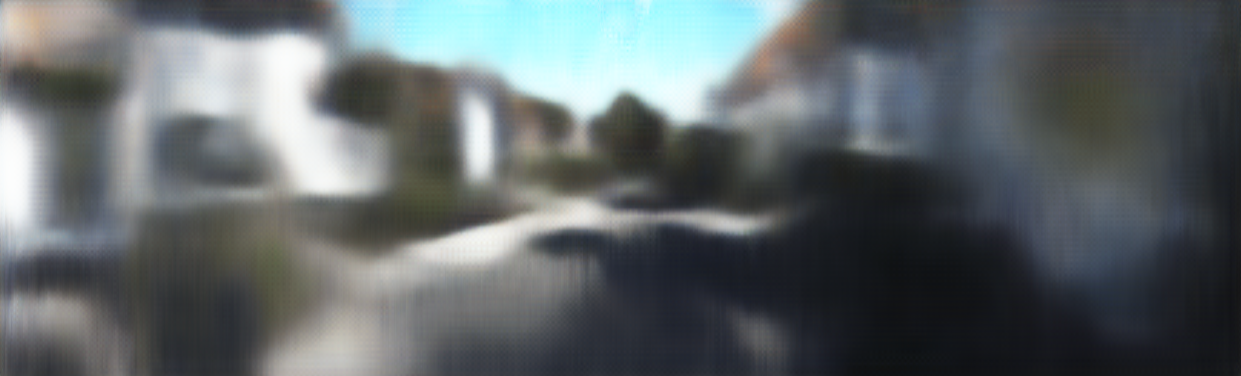}\par               
    \includegraphics[width=1.0\linewidth]{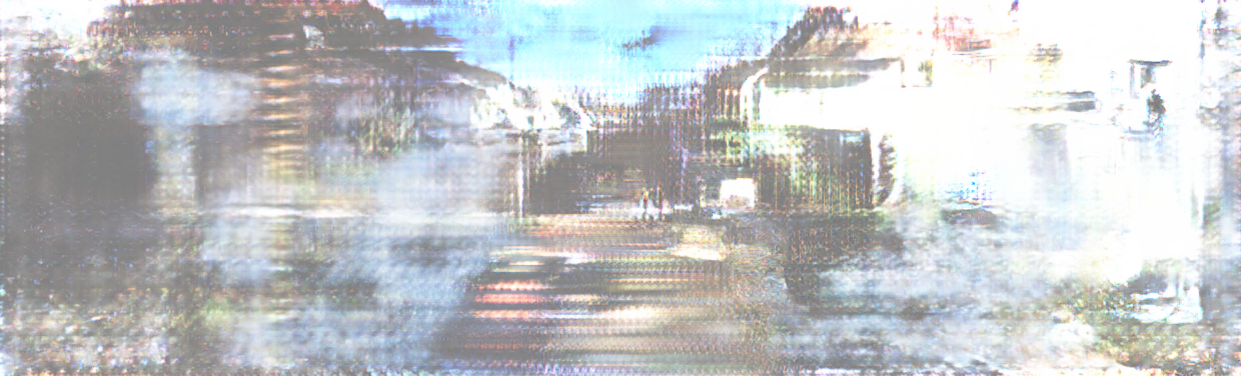}\par 
    \includegraphics[width=1.0\linewidth]{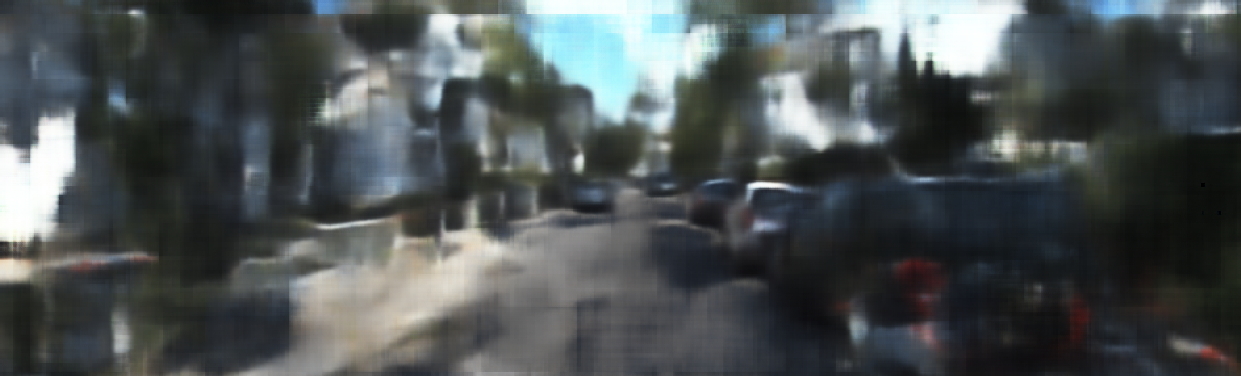}\par
\end{multicols}

%
%
%

\caption{Sample images generated directly from the projected point cloud images without employing the segmentation maps. From left to right we have the ground-truth image and the results from TITAN-Net (Ours), Pix2Pix~\cite{pix2pix2017}, and SC-UNET~\cite{Kim2020ColorIG}.  
}
\label{fig:synt_img_results_wo_segmap}
\end{figure*}

\subsection{Quantitative \& Qualitative Results}

We start with evaluating the quality of the generated semantic segmentation masks since the remaining image synthesis solely relies on these segments in our proposed framework. 
Table~\ref{tab:gen_sem_seg} shows the obtained mIoU scores on the validation set for the  segment maps $\widehat{y}_{s}$ generated  from   $p^{\prime}$ and $p^{\prime}_{s}$ (see Fig.~\ref{fig:overview}). 
Note that the output of SD-Net~\cite{nvidia_vid2vid}, \ie $y_s$, is here considered as the ground-truth since it is also employed by the TITAN-Net and Pix2Pix \textit{Discriminators}. 
Table~\ref{tab:gen_sem_seg} shows that the replacement of our proposed TITAN-Net model  with the Pix2Pix~\cite{pix2pix2017} counterpart in Fig.~\ref{fig:overview} leads to a substantial drop in the segmentation accuracy, without having any exception in the individual classes.  

Obtained quantitative results  on the quality of the final synthesized RGB images, $\widehat{y}$,   are reported in Table~\ref{tab:gen_rgb_im}.  
We, here, compare the performance of TITAN-Net combined with Vid2Vid (\ie TITAN-Net $\rightarrow$ Vid2Vid) to the other approaches, \ie SC-UNET~\cite{Kim2020ColorIG} and Pix2Pix~\cite{pix2pix2017}  combined with Vid2Vid (\ie Pix2Pix $\rightarrow$ Vid2Vid). 
Table~\ref{tab:gen_rgb_im} clearly shows that our proposed approach (\ie TITAN-Net $\rightarrow$ Vid2Vid) considerably outperforms the others by leading to the lowest FID and SWD scores. 

When it comes to the SSIM metric, SC-UNET~\cite{Kim2020ColorIG} performs better than the other two methods. 
We will elaborate more on this result in section~\ref{sec:ablationstudy}.

Fig.~\ref{fig:synt_img_results} shows   sample  segment maps and   RGB  images generated by our framework  (\ie TITAN-Net $\rightarrow$ Vid2Vid) in comparison with      Pix2Pix~\cite{pix2pix2017}   and Vid2Vid (\ie Pix2Pix $\rightarrow$ Vid2Vid). 
This figure clearly shows that our TITAN-Net model can reconstruct more accurate segment maps, thus, has much better image synthesis capability compared to Pix2Pix. 
For instance, the semantically important classes (such as buildings, roads, and vehicles) are reconstructed with high fidelity, as depicted in Fig.~\ref{fig:synt_img_results}.  
Note that since SC-UNET~\cite{Kim2020ColorIG} does not rely on the  segment masks, it is omitted in this figure.

\subsection{Ablation Study}
\label{sec:ablationstudy} 

We conduct ablation studies to better understand the contribution of different components in our     framework.

\textbf{Effect of the semantic segmentation maps:}  
We assess the contribution of semantic segments to the image reconstruction process.
Therefore, we measured the performance of TITAN-Net and Pix2Pix when the intermediate semantic segmentation step is completely bypassed as in the case of SC-UNET~\cite{Kim2020ColorIG}. Here, the RGB images $\widehat{y}$ are  directly recovered from the projected point cloud images  $p^{\prime}$. 
We followed the same training protocol defined for the previous experiments. However, we used Mean Squared Error (MSE) as our guiding loss in Eq.~\ref{eq:quidingLoss} since  the \ls loss only applies to segments. 

Obtained results without accessing the segmentation maps (\textit{w/o SegMap}) for both TITAN-Net and Pix2Pix are respectively reported in the fourth and fifth rows in Table~\ref{tab:gen_rgb_im}.
Qualitative synthesized images are also depicted in Fig.~\ref{fig:synt_img_results_wo_segmap}.
These results convey the fact that excluding the semantics drastically diminishes the quality of synthesized images.
The reason for obtaining a better SSIM score in Table~\ref{tab:gen_rgb_im} is that the model rather learns how the luminance and contrast behave but fails to capture   abstract context information as shown in Fig.~\ref{fig:synt_img_results_wo_segmap}.


\textbf{Effect of having   $p^{\prime}$  as a  condition in TITAN-Net:}  
In the previous ablation study, we implicitly investigated the  role of LiDAR segments $p^{\prime}_{s}$ in image synthesis. We now   diagnose the contribution of $p^{\prime}$  in the quality of generated RGB images. The last row in Table~\ref{tab:gen_rgb_im} shows that when the condition on $p^{\prime}$ is removed, the obtained results get worse in contrast to the results in the third row.

%

\begin{figure*}[!t]
\centering 
\includegraphics[width=1.0\linewidth, height=0.12\linewidth]{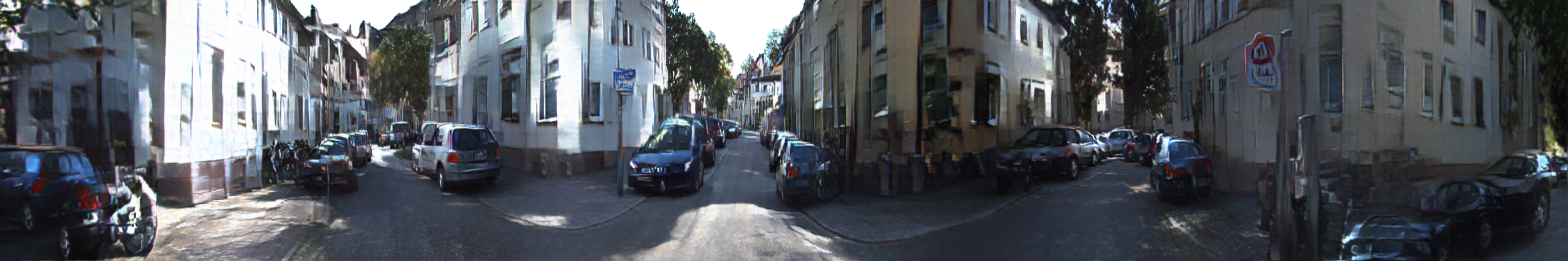}\par
\vspace{0.05cm}
\includegraphics[width=1.0\linewidth, height=0.12\linewidth]{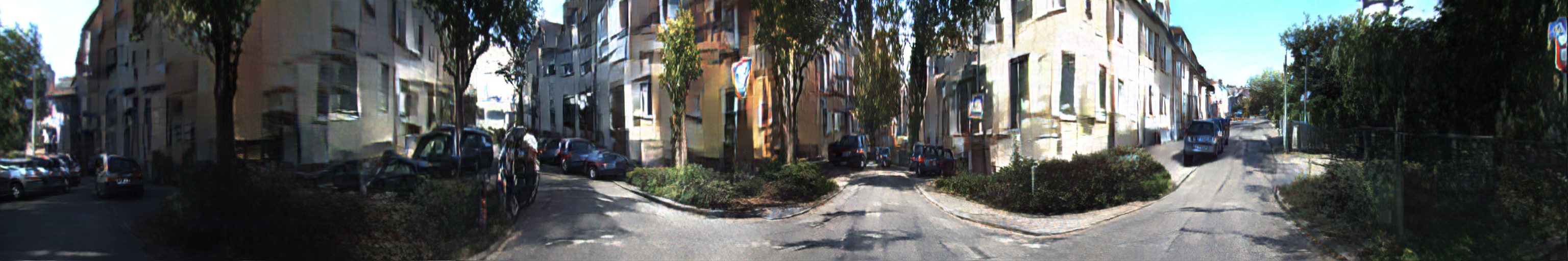}\par     
\caption{Two sample panoramic images  synthesized by our proposed TITAN-Net model on the \sk test set.}
\label{fig:panoramic}
\end{figure*}

\begin{figure}[!b]
\centering 
\includegraphics[width=0.9\linewidth]{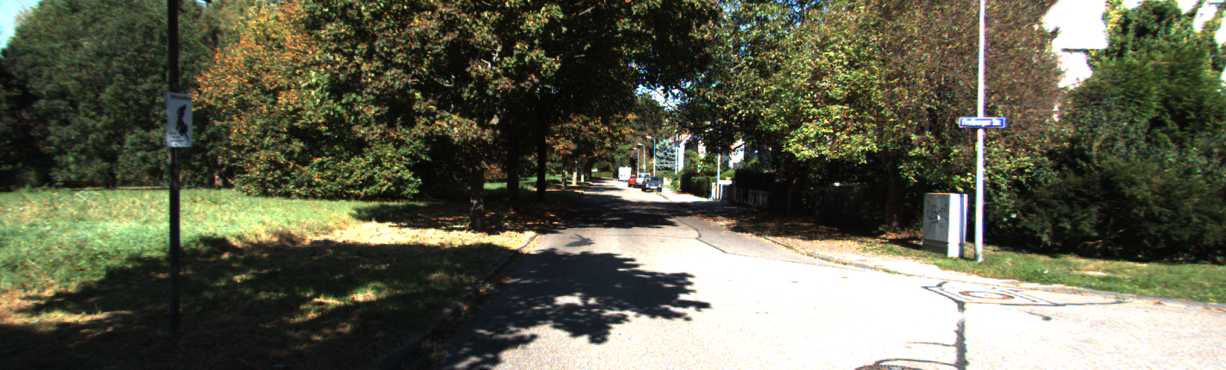}\par 
\includegraphics[width=0.9\linewidth]{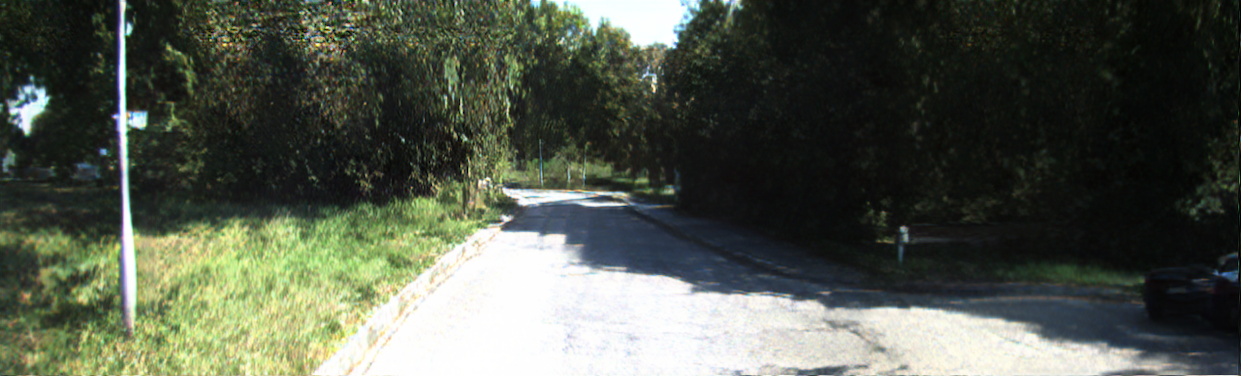}\par     
\includegraphics[width=0.9\linewidth]{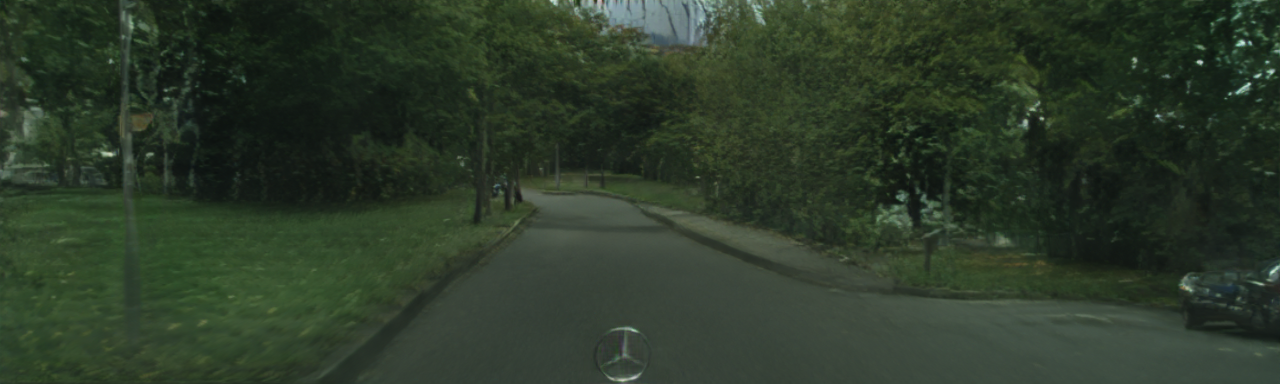}\par 
  
\caption{Different variants generated by our framework. From top to bottom, they are the ground truth camera image,  generated image by TITAN-Net $\rightarrow$ Vid2Vid trained on the \sk dataset, and synthesized image by TITAN-Net $\rightarrow$ Vid2Vid trained on     Cityscapes, respectively.}
\label{fig:kitti_vs_cityscape}
\end{figure}

\subsection{Runtime}
The runtime for  training on two Quadros RTX 6000 GPUs is about six days for 19K training samples and 118 epochs. Regarding the  inference time, on a single Quadros RTX 6000 GPU it takes $40.79,$ $198.75,$ $42.15$,  $123.72$ msecs for point cloud segmentation, image segmentation, translation between segment maps, and synthesizing a $376 \times 1241$ image.  When it comes to our baseline models, translation between segment maps takes about $22.48 $ and $87.72$ msecs for the Pix2Pix~\cite{pix2pix2017} and  SC-UNET~\cite{Kim2020ColorIG} models, respectively. These values were calculated on the validation split.

\section{Limitations and Discussion}

In this work, we argue that employing scene semantics is of utmost importance in translating features between the domains that have unique data formats such as 3D point clouds and 2D images. Findings provided in Table~\ref{tab:gen_rgb_im} and Figs.~\ref{fig:synt_img_results}-\ref{fig:synt_img_results_wo_segmap} clearly support  our hypothesis that the semantic data representation can, to a great extent, alleviate the domain translation problem across different sensor modalities.

Unlike other relevant works~\cite{Milz2019Points2Pix3P,Kim_2019,Kim2020ColorIG}, our proposed framework has the capacity to process the full $360^\circ$ LiDAR scan. This gives us a unique chance to synthesize panoramic camera images from  the projected point cloud range images $p^{\prime}$. Fig.~\ref{fig:panoramic} illustrates two sample panoramic images generated by our framework using the test split of the dataset. Note that due to lack of ground truth, it is non-trivial to evaluate the quality of these rendered panoramic images.  
Therefore, the results presented so far in Figs.~\ref{fig:synt_img_results}-\ref{fig:synt_img_results_wo_segmap}  involve  images synthesized only from a restricted region in the  LiDAR projection, which approximates the original camera view, \ie the only available ground truth.
We emphasize that this novel contribution plays a crucial role in handling possible sensor failures in autonomous vehicles. Take an example of having a failed camera sensor. The missing scene images can then be translated from other functional sensors, \eg LiDAR. Thus, the vehicle can employ these generated images as \textit{initial beliefs}  to bootstrap the subsequent sensor fusion and maneuver planning processes  instead of simply having a sudden emergency stop. 

Furthermore,   such a smooth translation  between different   modalities  can allow us to  gather, for instance, additional annotated data with no extra effort. The top row in  Fig~\ref{fig:kitti_vs_cityscape} shows an original camera image   from the \sk dataset. By using the corresponding LiDAR scan,   our framework   trained on this dataset  can already produce a variation of this scene  as depicted in the middle row in Fig~\ref{fig:kitti_vs_cityscape}. We can now simply replace the Vid2Vid  head with the version trained on the Cityscapes dataset~\cite{Cityscapes} to produce  a different variant as shown in the bottom row in Fig~\ref{fig:kitti_vs_cityscape}. 
Producing such different variants  without additional effort can help us to augment the available image datasets needed to efficiently regularize the neural networks.
%

 
We are aware of the fact that some   vehicle samples in the generated images may have visual artifacts, \eg vehicle boundaries are not preserved, in particular, when the scene has  multiple vehicle  samples (see Fig.~\ref{fig:panoramic}). The main reason is that our framework relies on \snx and SD-Net, which are not instance-aware segmentation approaches. We believe that segmenting individual instances can largely mitigate this problem. 
Thanks to having a modular framework, the segmentation networks can easily be replaced with the instance-aware counterparts in Fig.~\ref{fig:overview}.

Another limitation in our   framework is that  each LiDAR and camera data is treated individually. Thus there is no  temporal  consistency  between synthesized   images. 
We plan to extend our approach by incorporating   temporal cues to overcome this issue.

In the supplementary  material, we provide more images together with a video\footnote{\href{https://youtu.be/eV510t29TAc}{https://youtu.be/eV510t29TAc}} showing the performance of TITAN-Net on the validation and test splits.

\section{Conclusion}

In this work\footnote{The research leading to these results has received funding from the
Vinnova FFI project SHARPEN, under grant agreement no. 2018-05001.}, we introduce a novel semantics-aware domain translation framework to synthesize panoramic color images from a given     point cloud. Our  framework is a modular approach and involves four different models  trained individually. The framework relies on   our new  cGAN model, TITAN-Net,  which translates the LiDAR semantic maps to camera image formats to boost the image generation.

{
\clearpage
\small
\bibliographystyle{ieee_fullname}
\bibliography{salsaBIB}
}

\clearpage

\twocolumn[\section*{\textit{Supplementary Material} \\ Semantics-aware Multi-modal  Domain Translation: \\ From LiDAR Point Clouds to Panoramic Color Images}]
\%title{\textit{Supplementary Material} \\ Semantics-aware Multi-modal  Domain Translation: \\ From LiDAR Point Clouds to Panoramic 
%
%


\ificcvfinal\thispagestyle{empty}\fi

We here provide more additional material to support our main submission. In the first section, we provide a detailed description of the  TITAN-Net \textit{Generator} and \textit{Discriminator} architectures. Next, we provide a table showing the corresponding label matches between the \sk and   Cityscapes   datasets. We also present an additional ablation study. Finally, we provide more qualitative experimental results, \eg more synthesized images and  videos showing the performance of TITAN-Net on the validation and test splits.

\begin{figure*}[!t]
  \includegraphics[width=0.95\linewidth]
   {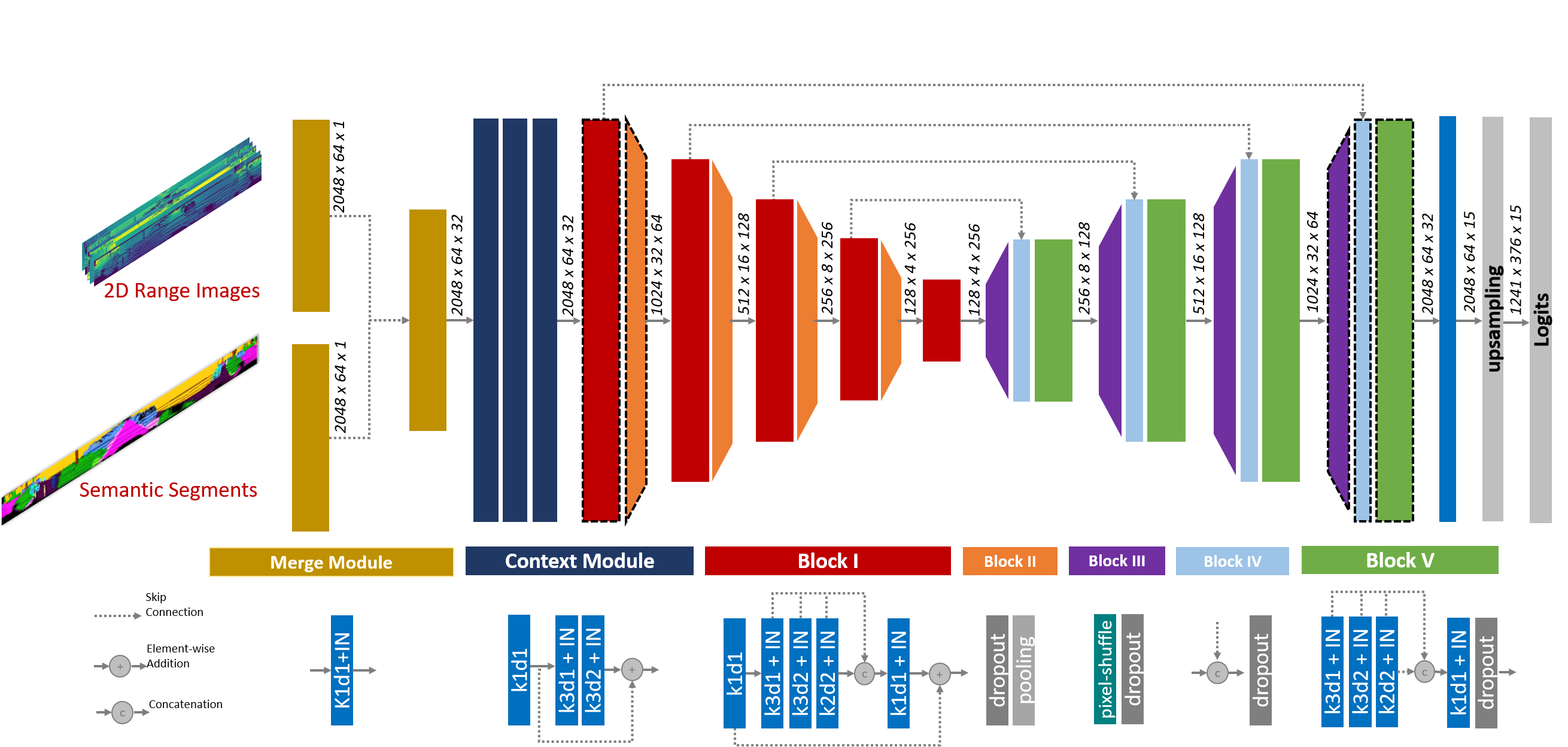}
   \caption{TITAN-Net Generator. Blocks with dashed edges are not involving dropout. The abbreviations \textit{k}, \textit{d}, and \textit{IN} stand for the kernel size, dilation rate and instance normalization, respectively.}
   \label{fig:G}
\end{figure*}

\begin{figure}[!b]
  \includegraphics[width=\linewidth]
   {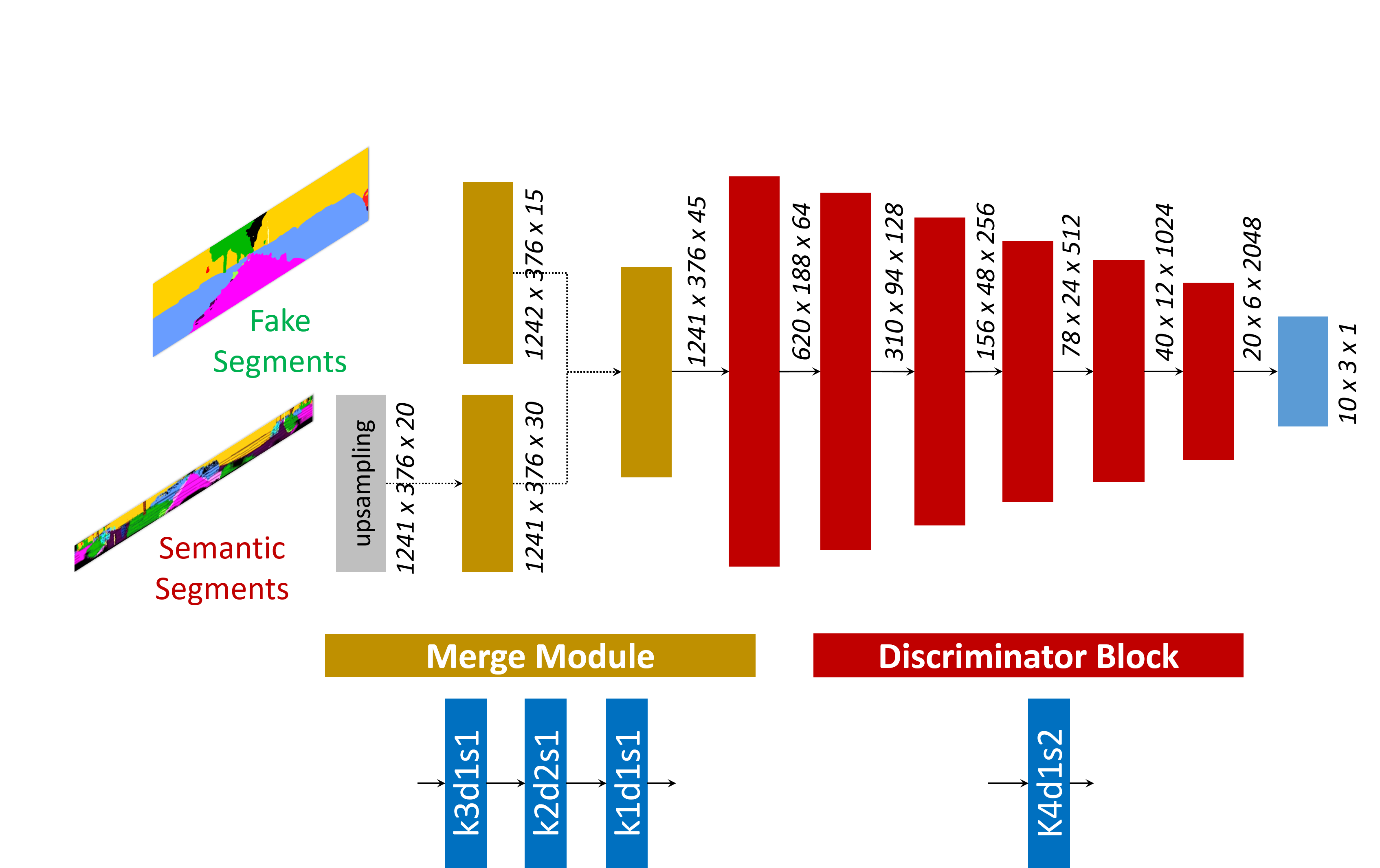}
   \caption{TITAN-Net Discriminator. The abbreviations \textit{k} and \textit{d}   stand for the kernel size and dilation rate, respectively.}
   \label{fig:D}
\end{figure}   

\section{TITAN-Net Architecture}
As described in the main manuscript TITAN-Net is a conditional GAN model involving two units: \textit{Generator} and \textit{Discriminator}. 

\textbf{TITAN-Net Generator:}  Fig.~\ref{fig:G} shows the \textit{Generator} architecture, 
which is divided into four main components, described as follows: 

\begin{itemize}
\item \textbf{Merge Module:}  This module is responsible for the early fusion of LiDAR semantic segmentation maps and range-view projections.  Instead of a naive concatenation of both inputs, we, first, feed each of them to a $1 \times 1$ convolutional layer before concatenating and feeding to another $1 \times 1$ convolutional layer. This allows us to exploit the raw inputs more efficiently. The choice of $1 \times 1$ kernel also comes from the fact that we pretend to combine the inputs from a local perspective without considering the surrounding pixels.

\item \textbf{Contextual Module:} The \textit{Generator} has a contextual module which learns the global context information, \eg complex correlations between segment classes  by large receptive fields. 
More precisely, the contextual module has a set of residual dilated convolution operations fusing a large receptive field ($3 \times 3$ )  with a smaller one ($1 \times 1$) through a skip connection to aggregate the context cues in different regions.  This way, the  \textit{Generator} captures fine detailed spatial information while extracting the global context.

\item \textbf{U-Net architecture:} The main skeleton of the \textit{Generator} relies on a U-Net like encoder-decoder architecture. The encoder unit has  blocks of dilated convolutions (see Block I in Fig.~\ref{fig:G})  with gradually increasing receptive fields of  $3 \times 3$, $5 \times 5$, and $7 \times 7$. Each block first passes the received feature map through a $1 \times 1$ layer that will be the primary skip connection after the inner block. 
The decoder employs \textit{pixel-shuffle} layers (see Block III in Fig.~\ref{fig:G}) that exploit the learnt feature maps to upsample the spatial dimension. Unlike conventional  transpose convolutions which are   prone to checkerboard artifacts, \textit{pixel-shuffle} layers have less parameters  and  force the learnt feature maps to retain more  information.
The decoder involves another blocks of  dilated convolutions (Block V in Fig.~\ref{fig:G}) to extract more descriptive features.

\item \textbf{Output:} As shown in Fig.~\ref{fig:G}, we finally perform a bilinear upsampling at the end of the network to generate segmentation maps in the camera image space while avoiding visual artifacts from the start. 
 
\end{itemize}

\textbf{TITAN-Net Discriminator}: The \textit{Discriminator} follows the same central idea as in the \textit{Generator}. Both inputs are fused using a convolutional block as illustrated in Fig.\ref{fig:D}. In this case, the conditional input (the LiDAR segmentation map) is upsampled to meet the original camera image dimension  before the concatenation. 

The \textit{Discriminator} model follows the   PatchGAN structure. This type of \textit{Discriminator} focuses on penalising structures at a local patch scale. Instead of mapping an entire image to a single scalar, we   have a final and smaller representation of the inputs, representing the realism of the original image's different and independent regions. This can be seen as if we manually divide  the original image into smaller patches and pass  each of them through the \textit{Discriminator}. The advantage of this method relies on the fact that we do not need to apply a preprocessing to the inputs - to divide them into patches - while the \textit{Generator} is still able to operate on the full input instead of patches that can hurt the performance. 

As depicted in Fig.\ref{fig:D}, the TITAN-Net \textit{Discriminator} has six strided convolutional blocks that   yields a final patch size of $10 \times 3$. No normalisation is used. 

%
\begin{table}[!b]
\scalebox{0.93}{
\centering
\begin{adjustbox}{max width=0.5\textwidth}
\begin{tabular}{cl|lllllllllllllll}
\multirow{22}{*}{\begin{sideways}\textbf{SemanticKitti~~~~~~~~~~~~~~~~~~~~~~~~~~~~~~~~~~~~~~~~~~~~~~~~~~~~~~~~~~}\end{sideways}} & \multicolumn{1}{l}{} & \multicolumn{15}{c}{\textbf{Cityscapes}} \\
 &  & \multicolumn{1}{c}{\begin{sideways}Unlabeled\end{sideways}} & \multicolumn{1}{c}{\begin{sideways}Car\end{sideways}} & \multicolumn{1}{c}{\begin{sideways}Bicycle\end{sideways}} & \multicolumn{1}{c}{\begin{sideways}Motorcycle\end{sideways}} & \multicolumn{1}{c}{\begin{sideways}Truck\end{sideways}} & \multicolumn{1}{c}{\begin{sideways}Other-Vehicle\end{sideways}} & \multicolumn{1}{c}{\begin{sideways}Person\end{sideways}} & \multicolumn{1}{c}{\begin{sideways}Road\end{sideways}} & \multicolumn{1}{c}{\begin{sideways}Sidewalk\end{sideways}} & \multicolumn{1}{c}{\begin{sideways}Building\end{sideways}} & \multicolumn{1}{c}{\begin{sideways}Fence\end{sideways}} & \multicolumn{1}{c}{\begin{sideways}Vegetation\end{sideways}} & \multicolumn{1}{c}{\begin{sideways}Terrain\end{sideways}} & \multicolumn{1}{c}{\begin{sideways}Pole\end{sideways}} & \multicolumn{1}{c}{\begin{sideways}Traffic-Sign\end{sideways}} \\ 
\cline{2-17}
 & Unlabeled & $\mathit{X}$ &  &  &  &  &  &  &  &  &  &  &  &  &  &  \\
 & Car &  & $\mathit{X}$ &  &  &  &  &  &  &  &  &  &  &  &  &  \\
 & Bicycle &  &  & $\mathit{X}$ &  &  &  &  &  &  &  &  &  &  &  &  \\
 & Motorcycle &  &  &  & $\mathit{X}$ &  &  &  &  &  &  &  &  &  &  &  \\
 & Truck &  &  &  &  & $\mathit{X}$ &  &  &  &  &  &  &  &  &  &  \\
 & Other-Vehicle &  &  &  &  &  & $\mathit{X}$ &  &  &  &  &  &  &  &  &  \\
 & Person &  &  &  &  &  &  & $\mathit{X}$ &  &  &  &  &  &  &  &  \\
 & Bicyclist &  &  &  &  &  &  & $\mathit{X}$ &  &  &  &  &  &  &  &  \\
 & Motorcyclist &  &  &  &  &  &  & $\mathit{X}$ &  &  &  &  &  &  &  &  \\
 & Road &  &  &  &  &  &  &  & $\mathit{X}$ &  &  &  &  &  &  &  \\
 & Parking &  &  &  &  &  &  &  &  & $\mathit{X}$ &  &  &  &  &  &  \\
 & Sidewalk &  &  &  &  &  &  &  &  & $\mathit{X}$ &  &  &  &  &  &  \\
 & Other-Ground & $\mathit{X}$ &  &  &  &  &  &  &  &  &  &  &  &  &  &  \\
 & Building &  &  &  &  &  &  &  &  &  & $\mathit{X}$ &  &  &  &  &  \\
 & Fence &  &  &  &  &  &  &  &  &  &  & $\mathit{X}$ &  &  &  &  \\
 & Vegetation &  &  &  &  &  &  &  &  &  &  &  & $\mathit{X}$ &  &  &  \\
 & Trunk &  &  &  &  &  &  &  &  &  &  &  & $\mathit{X}$ &  &  &  \\
 & Terrain &  &  &  &  &  &  &  &  &  &  &  &  & $\mathit{X}$ &  &  \\
 & Pole &  &  &  &  &  &  &  &  &  &  &  &  &  & $\mathit{X}$ &  \\
 & Traffic-Sign &  &  &  &  &  &  &  &  &  &  &  &  &  &  & $\mathit{X}$
\end{tabular}
\end{adjustbox}
}
  \caption{Mapping between the labels available on the Cityscapes and SemanticKITTI datasets. }
  \label{tab:labelmap}
\end{table}

\section{Mapping Between Different Datasets}

As already described in the main manuscript, \snx and SD-Net are trained on the \sk and Cityscapes   datasets, respectively. Both dataset have different   classes  with unique class  labels. To cope with the incompatibilities between these differences in   two   datasets, we define Table~\ref{tab:labelmap} returning $14$ unique class labels matched in both datasets.
This mapping allows us to have a better alignment between LiDAR and RGB labels used for training of TITAN-Net.

\begin{table*}[!t]
\centering
\resizebox{0.9\hsize}{!}{
\begin{tabular}{l||cccccccccccccc|c}
Approach &  \rotatebox{90}{Car} & \rotatebox{90}{Bicycle} & \rotatebox{90}{Motorcycle} & \rotatebox{90}{Truck}    & \rotatebox{90}{Other-Vehicle} & \rotatebox{90}{Person} & \rotatebox{90}{Road} & \rotatebox{90}{Sidewalk} & \rotatebox{90}{Building} & \rotatebox{90}{Fence} & \rotatebox{90}{Vegetation} & \rotatebox{90}{Terrain}  & \rotatebox{90}{Pole} & \rotatebox{90}{Traffic-Sign} & mIoU $\uparrow$ \\ 
\hline
Pix2Pix 					         & 8.8                                & 0                                 & 0                                 & 0                                  & 0                               & 0                                 & 57.7                               & 15.7                               & 32.8                               & 12.5                               & 32.7                               & 14.8                               & 0.5                             & 0                                  & 12.5                                  \\
TITAN-Net (Ours) &   \multicolumn{1}{c}{\textbf{68.2 }} & \multicolumn{1}{c}{\textbf{9.9 }} & \multicolumn{1}{c}{\textbf{7.6 }} & \multicolumn{1}{c}{\textbf{7.6 }} & \multicolumn{1}{c}{0\textbf{ }} & \multicolumn{1}{c}{\textbf{7.3 }} & \multicolumn{1}{c}{75.4} & \multicolumn{1}{c}{\textbf{48.3 }} & \multicolumn{1}{c}{\textbf{62.9 }} & \multicolumn{1}{c}{\textbf{33.6 }} & \multicolumn{1}{c}{\textbf{60.1 }} & \multicolumn{1}{c}{\textbf{49.9 }} & \multicolumn{1}{c}{\textbf{2 }} & \multicolumn{1}{c|}{\textbf{3.7 }} & \textbf{31.1 }
\\
TITAN-Net (w/o $\mathcal{L}_{ls}$)           & 60.5                                & 0                                 & 0                                 & 0.3                                  & 0                               & 0.05                                 & \textbf{78.0}                               & 43.6                              & 56.9                               & 33.0                               & 52.8                               & 42.9                              & 0.08                             & 0.07                                  & 26.2                                  \\
\end{tabular}
}
\caption{Quantitative results for the generated semantic segment  images on the test sequences.  
 $\uparrow$ denotes that   higher is better.  }
   \label{tab:gen_sem_seg}
\end{table*}
\section{Ablation Study}
As described in the main manuscript, the final TITAN-Net loss function has two components: Wasserstein GAN with Gradient Penalty   ($\mathcal{L}_{wgan-gp}$)    and  \ls   ($\mathcal{L}_{ls}$). 
We, here, ablate the \ls loss term to diagnose the overall contribution in the network performance.  
As reported in the last row of Table ~\ref{tab:gen_sem_seg},  we observe that the term $\mathcal{L}_{ls}$ has a certain contribution to the generation of the semantic segment maps. Including both loss terms lead to the best results (see the second row in Table ~\ref{tab:gen_sem_seg}), which suggests that both losses    regularize  the network in a complementary manner.

\section{Qualitative Results}
In the following, we provide more qualitative results compared to the other baseline models on the test dataset.
The following figures show different qualitative results:
\begin{itemize}
\item Fig.~\ref{fig:segmap} shows   sample  segment maps and   RGB  images generated by our framework  (\ie TITAN-Net $\rightarrow$ Vid2Vid) in comparison with      Pix2Pix    and Vid2Vid (\ie Pix2Pix $\rightarrow$ Vid2Vid), 
\item Fig.~\ref{fig:base} presents sample  images synthesized directly from the raw projected point cloud images without employing the segmentation maps.
\item Fig.~\ref{fig:360} shows sample panoramic images synthesized by our TITAN-Net model on the \sk test set.
\item Fig.~\ref{fig:city} depicts different variants of the original camera image, generated by   our framework  (\ie TITAN-Net $\rightarrow$ Vid2Vid) using the \sk and Cityscapes datasets.
 
\end{itemize}

\section{Video}

We also provide  three videos showing the performance of TITAN-Net on the validation and test splits of the \sk dataset. 
All three videos are available\footnote{\href{https://youtu.be/He6fKkF88IE}{https://youtu.be/He6fKkF88IE}}~\footnote{\href{https://youtu.be/k59zmVhsKVI}{https://youtu.be/k59zmVhsKVI}}~\footnote{\href{https://youtu.be/zR6Ix6YUhwI}{https://youtu.be/zR6Ix6YUhwI}}.
Note that  in our proposed framework  each LiDAR and camera data is treated individually. Thus, there is no  temporal  consistency  between synthesized images.

\setlength{\columnsep}{0.1cm}
\begin{figure*}[!t]
\begin{multicols}{3}
    \includegraphics[width=1.0\linewidth]{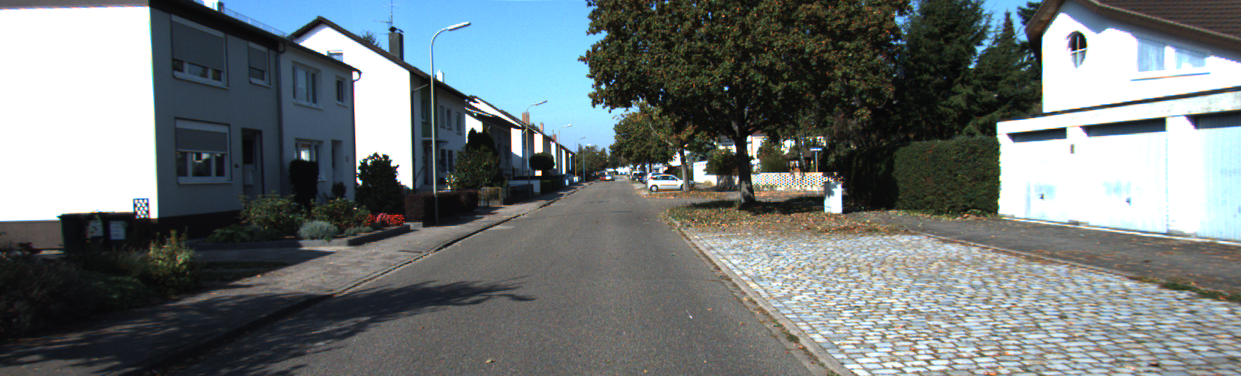}\par
    \includegraphics[width=1.0\linewidth]{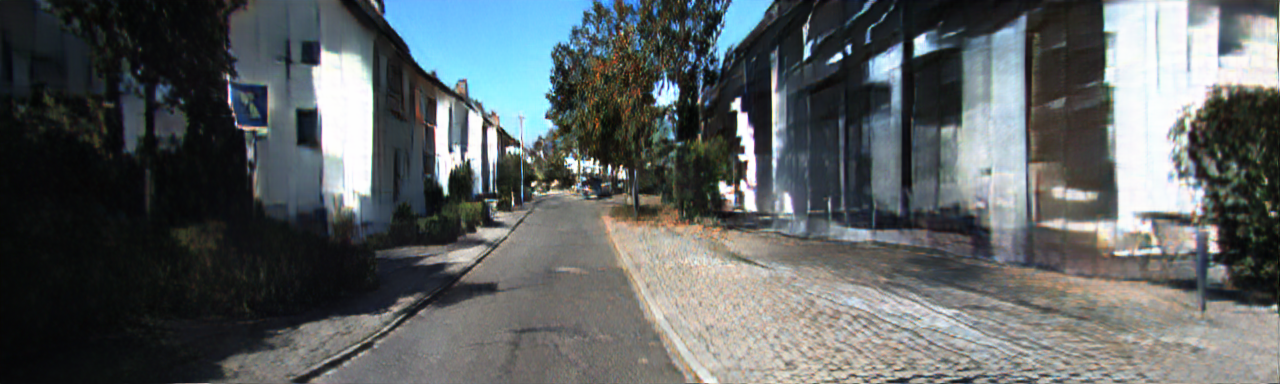}\par
    \includegraphics[width=1.0\linewidth]{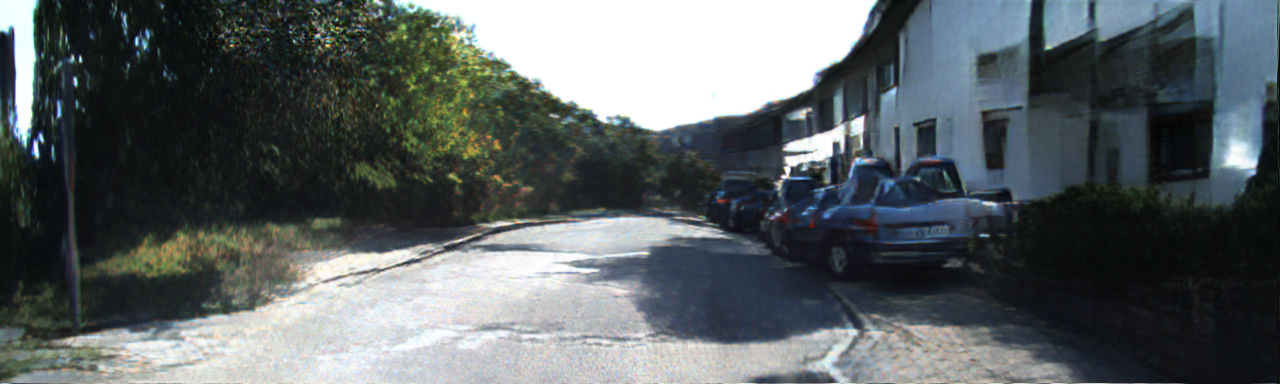}\par
\end{multicols}
\vspace{-0.95cm}
\begin{multicols}{3}
   \includegraphics[width=1.0\linewidth]{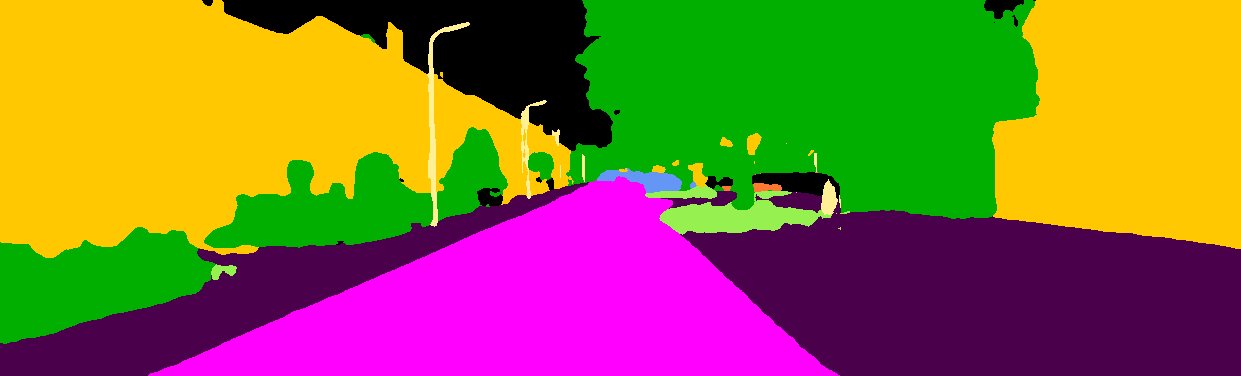} \par
    \includegraphics[width=1.0\linewidth]{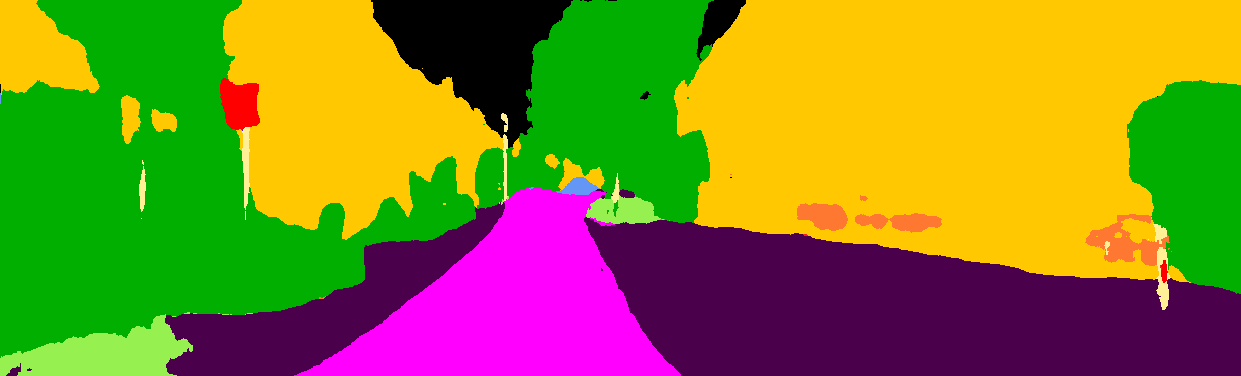}\par
    \includegraphics[width=1.0\linewidth]{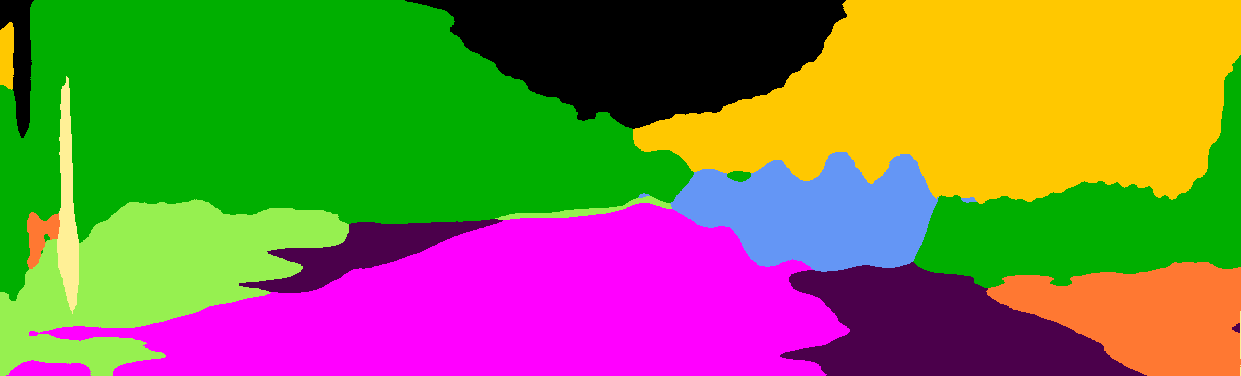}\par
\end{multicols}

\vspace{-0.8cm}

\begin{multicols}{3}
    \includegraphics[width=1.0\linewidth]{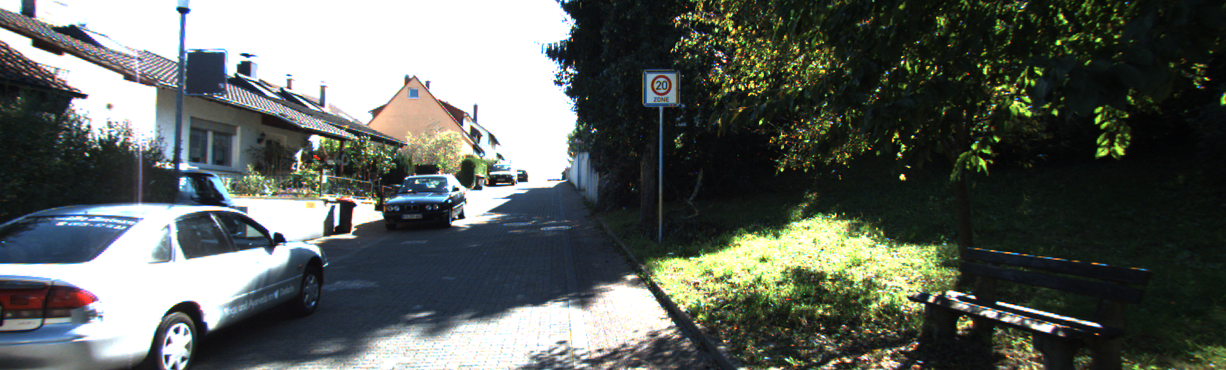}\par
    \includegraphics[width=1.0\linewidth]{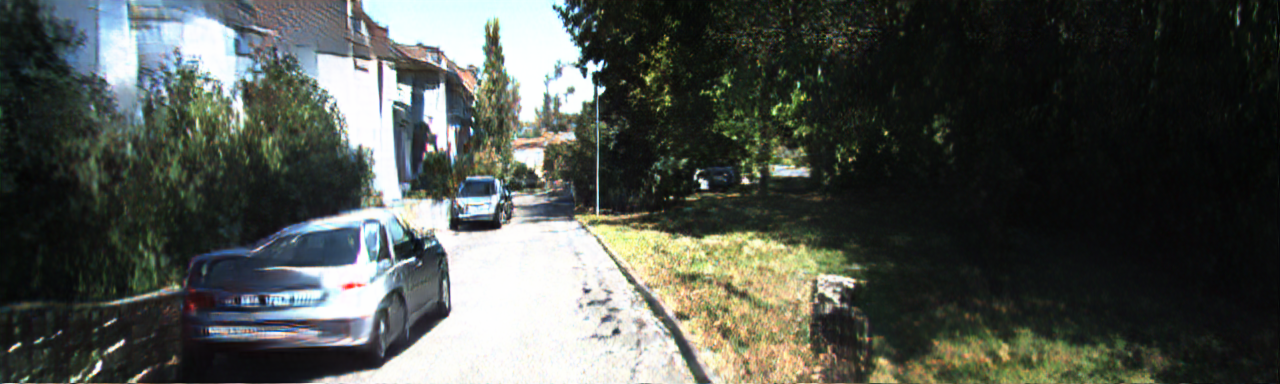}\par
    \includegraphics[width=1.0\linewidth]{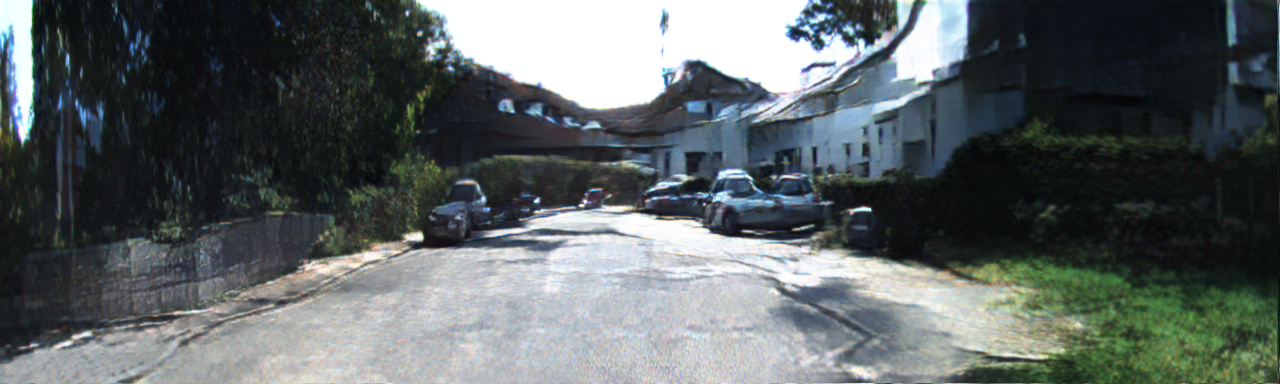}\par
\end{multicols}
\vspace{-0.95cm}
\begin{multicols}{3}
   \includegraphics[width=1.0\linewidth]{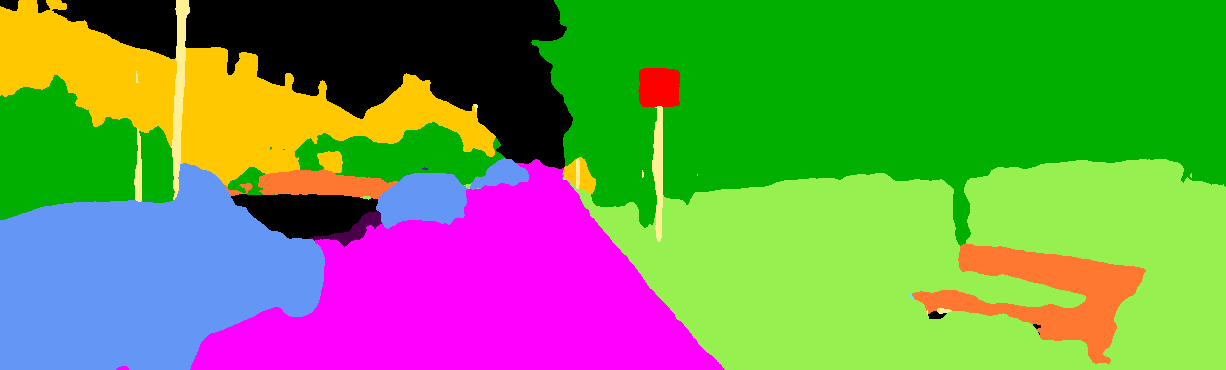} \par
    \includegraphics[width=1.0\linewidth]{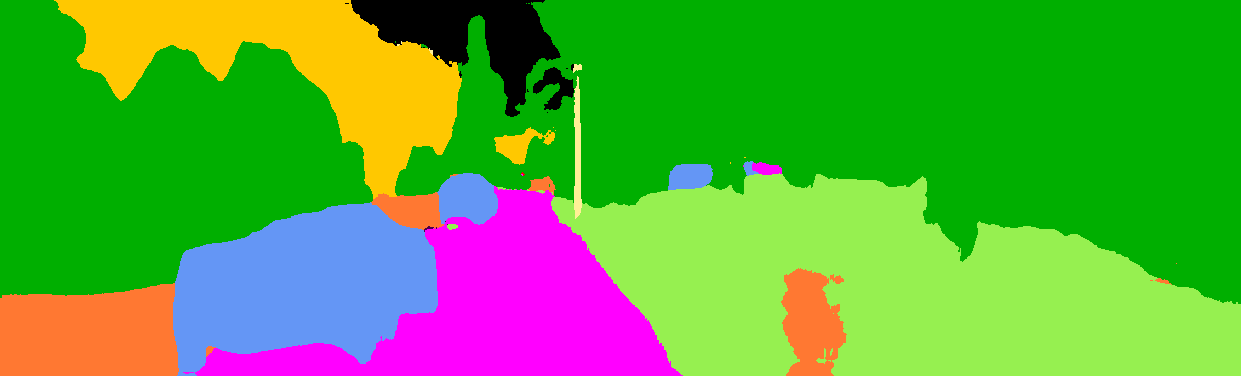}\par
    \includegraphics[width=1.0\linewidth]{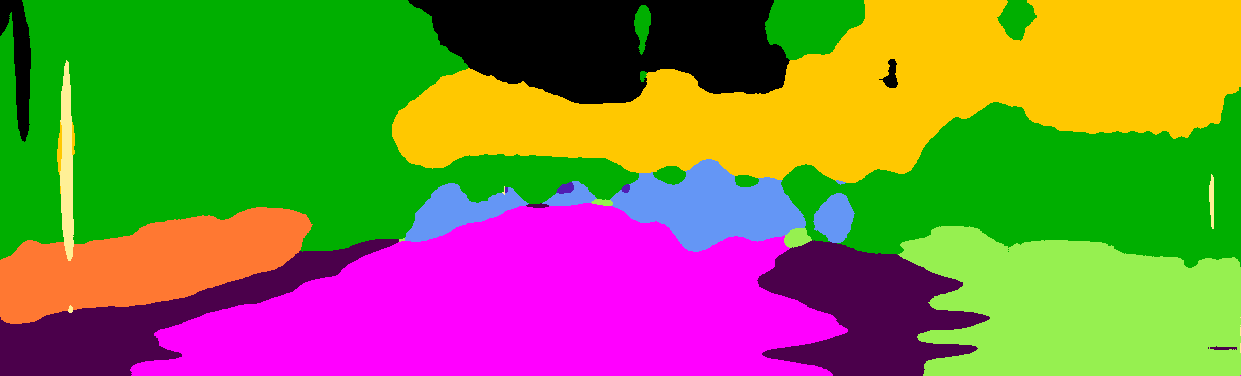}\par
\end{multicols}

\vspace{-0.8cm}

\begin{multicols}{3}
    \includegraphics[width=1.0\linewidth]{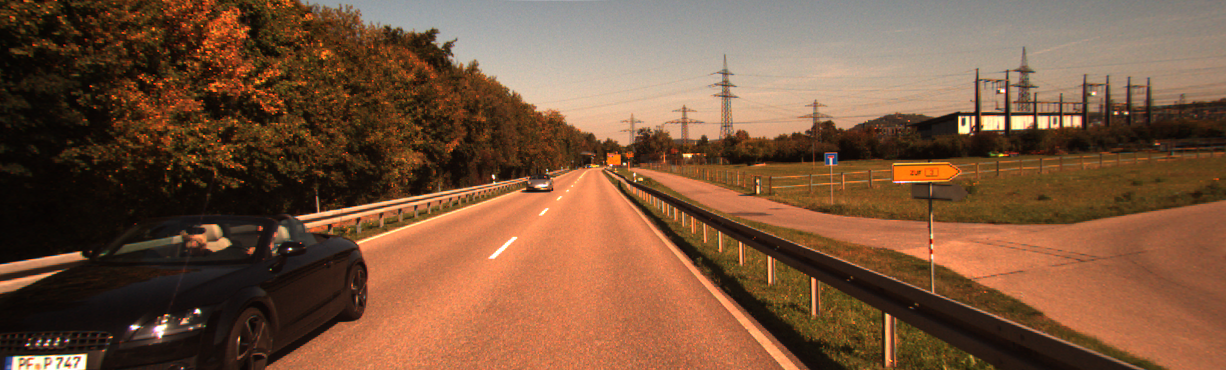}\par
    \includegraphics[width=1.0\linewidth]{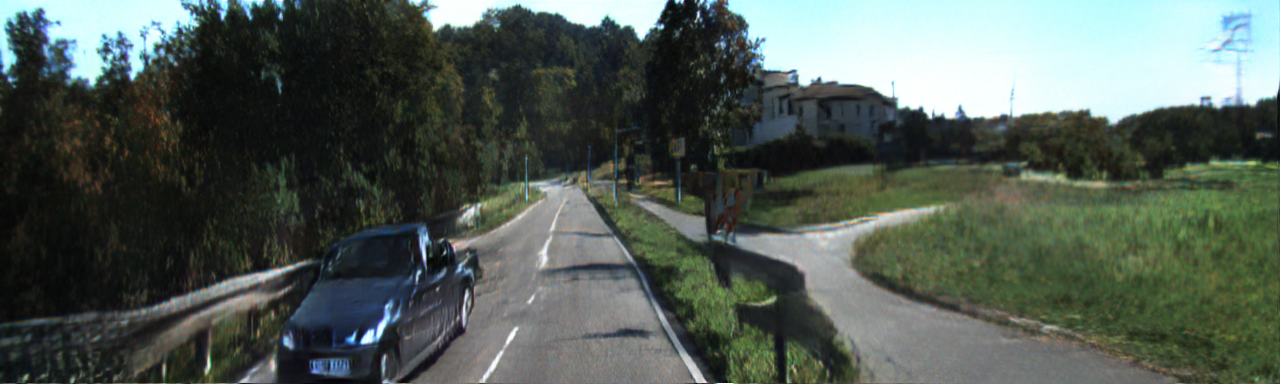}\par
    \includegraphics[width=1.0\linewidth]{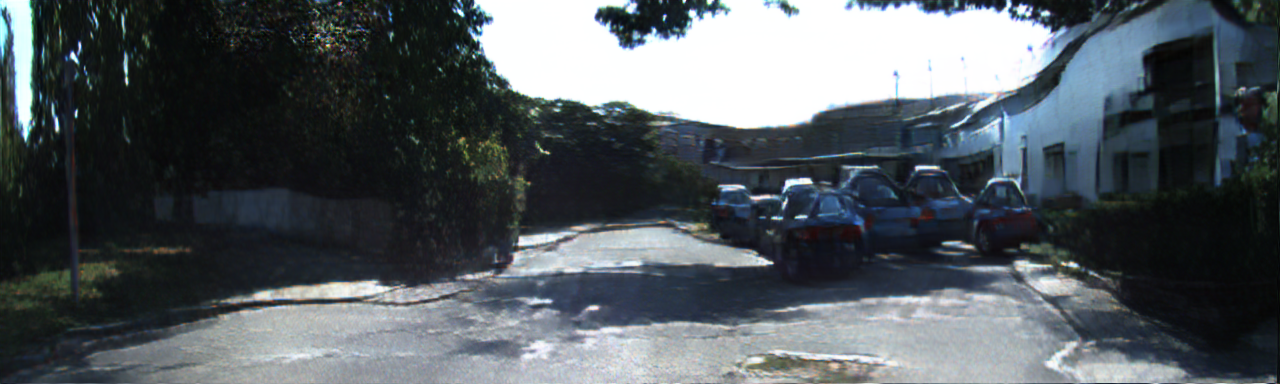}\par
\end{multicols}
\vspace{-0.95cm}
\begin{multicols}{3}
   \includegraphics[width=1.0\linewidth]{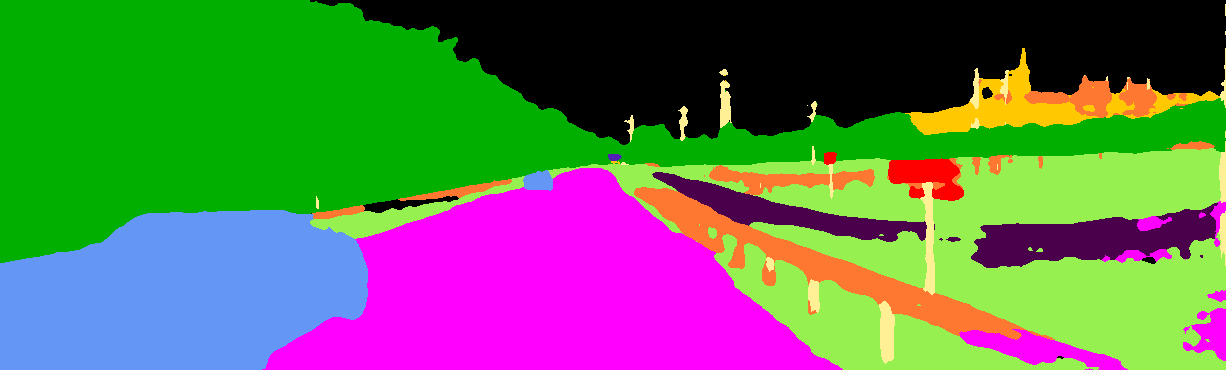} \par
    \includegraphics[width=1.0\linewidth]{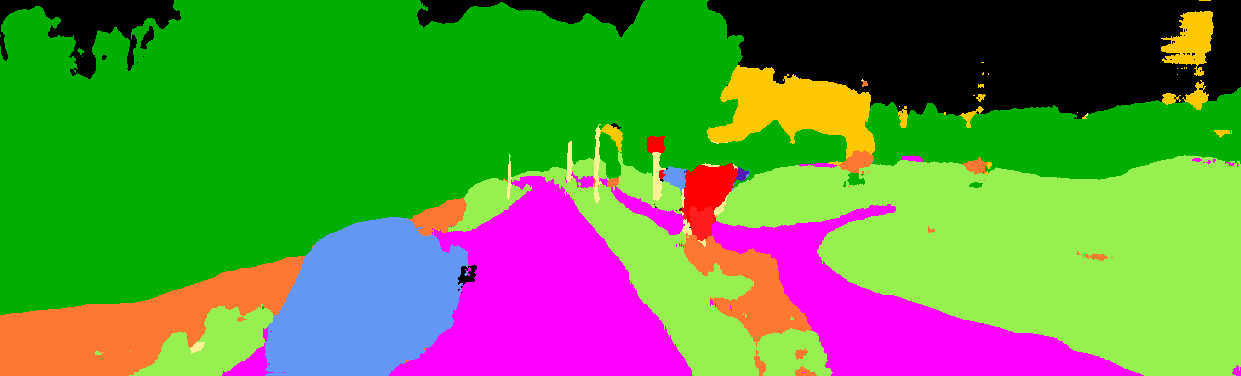}\par
    \includegraphics[width=1.0\linewidth]{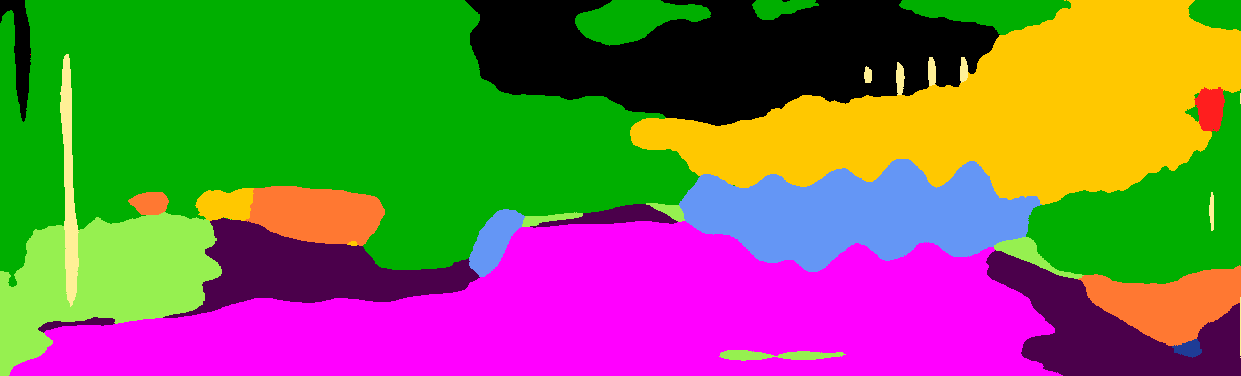}\par
\end{multicols}
 
\vspace{-0.8cm}

\begin{multicols}{3}
    \includegraphics[width=1.0\linewidth]{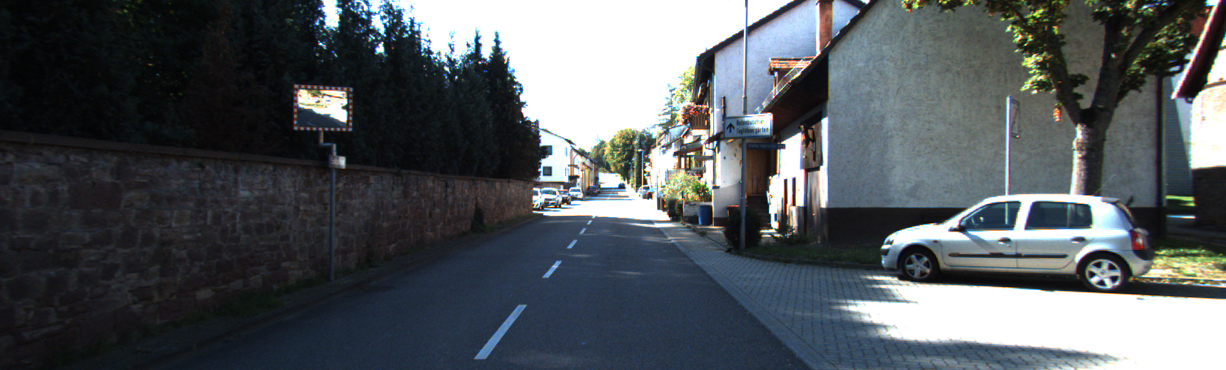}\par
    \includegraphics[width=1.0\linewidth]{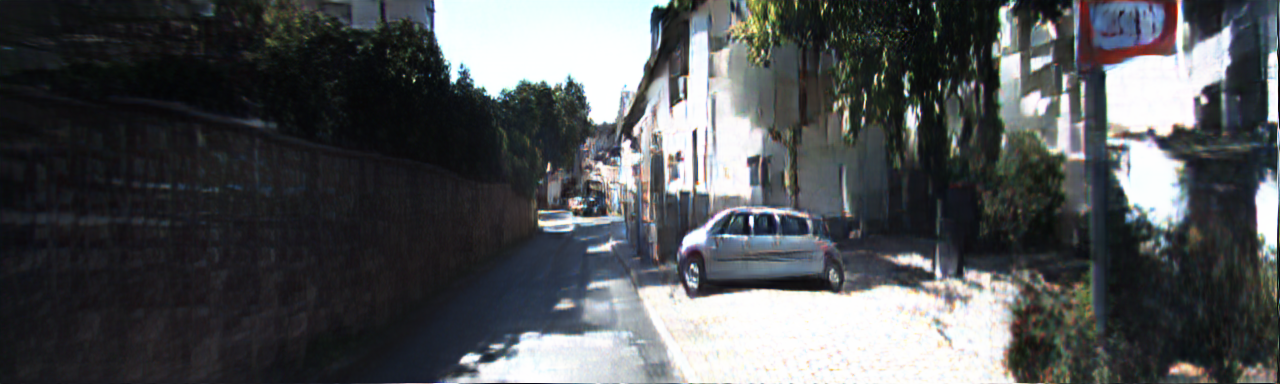}\par
    \includegraphics[width=1.0\linewidth]{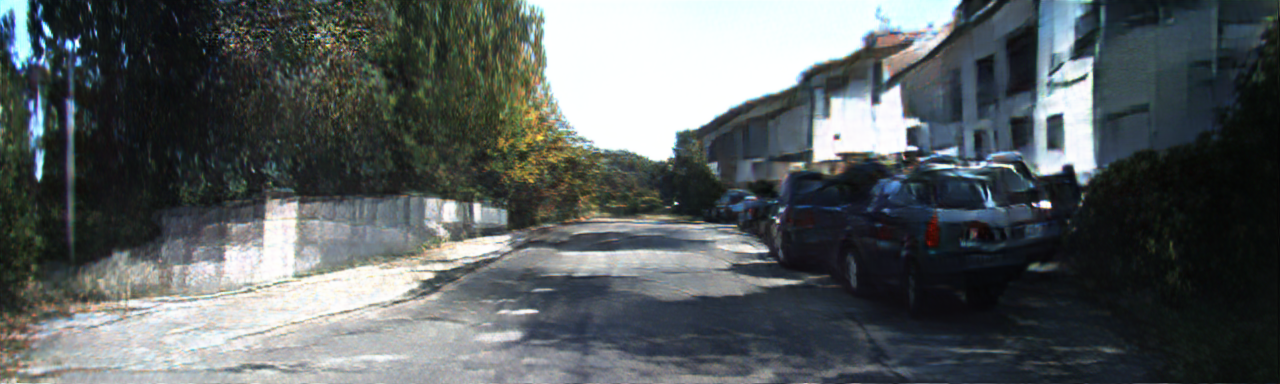}\par
\end{multicols}
\vspace{-0.95cm}
\begin{multicols}{3}
   \includegraphics[width=1.0\linewidth]{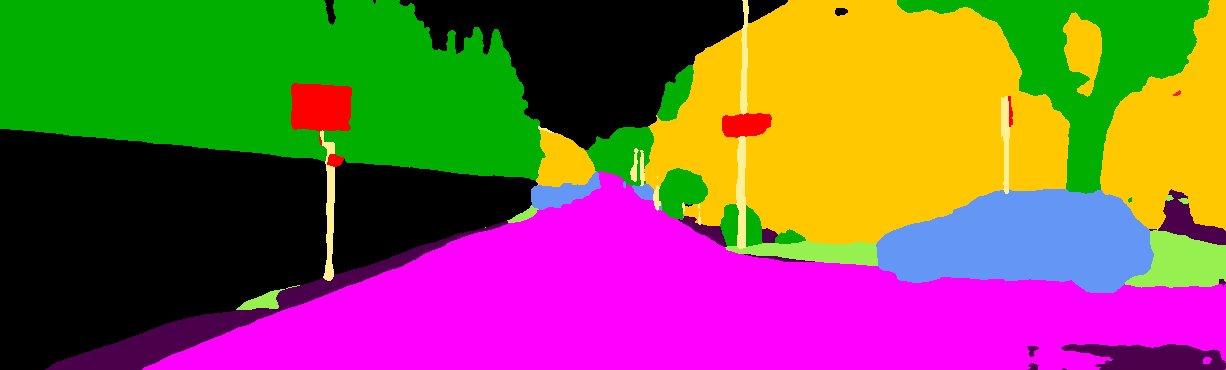} \par
    \includegraphics[width=1.0\linewidth]{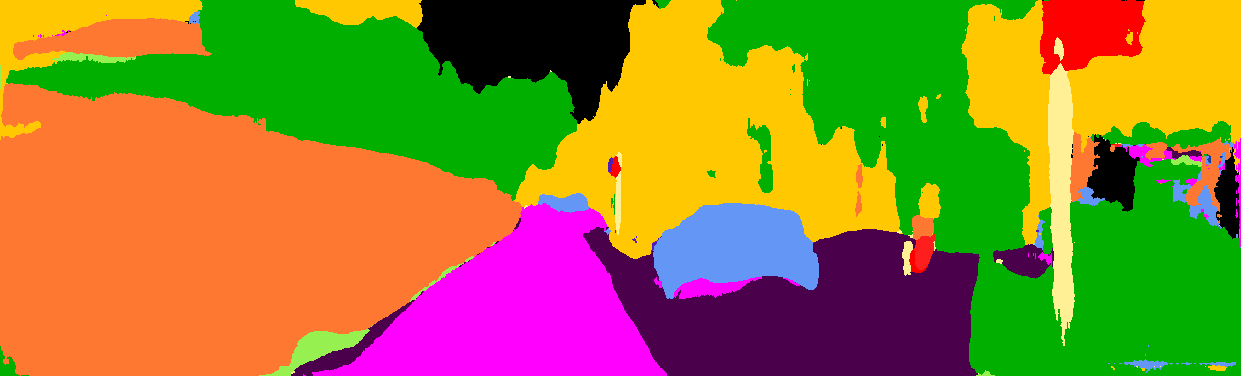}\par
    \includegraphics[width=1.0\linewidth]{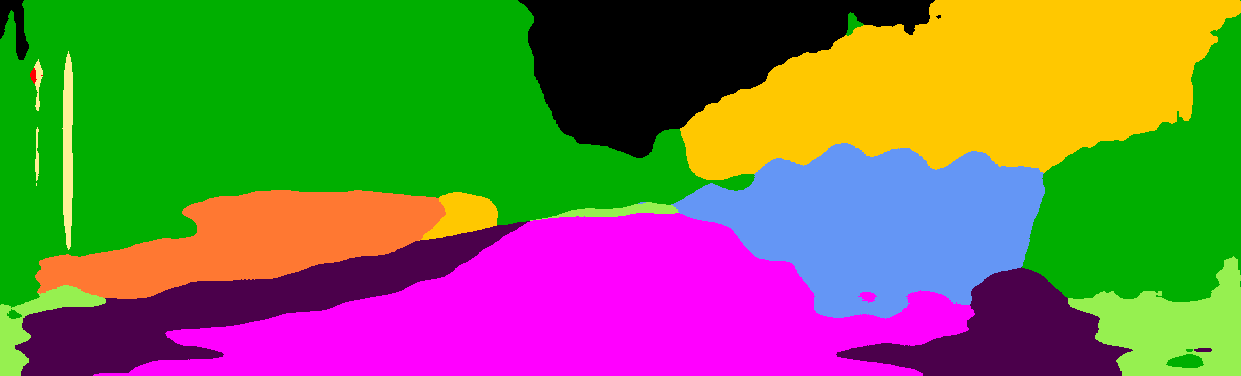}\par
\end{multicols}
\vspace{-0.8cm}

\begin{multicols}{3}
    \includegraphics[width=1.0\linewidth]{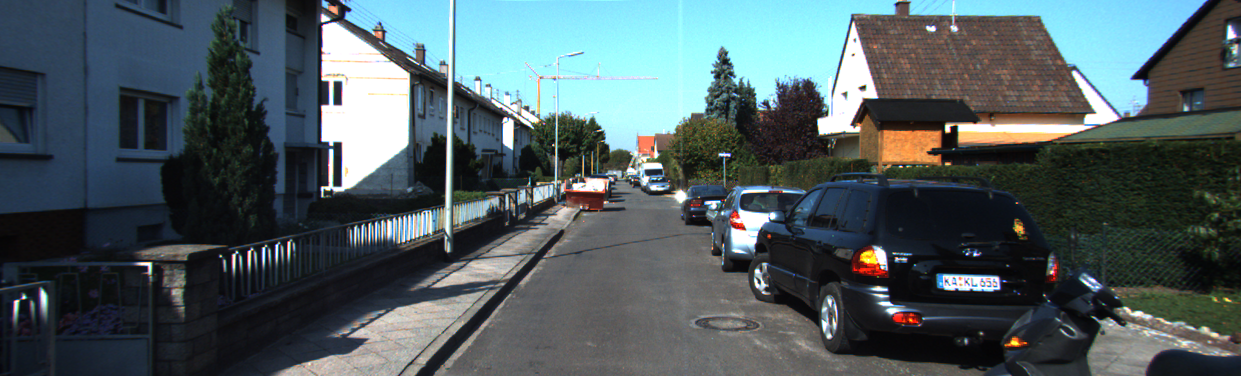}\par
    \includegraphics[width=1.0\linewidth]{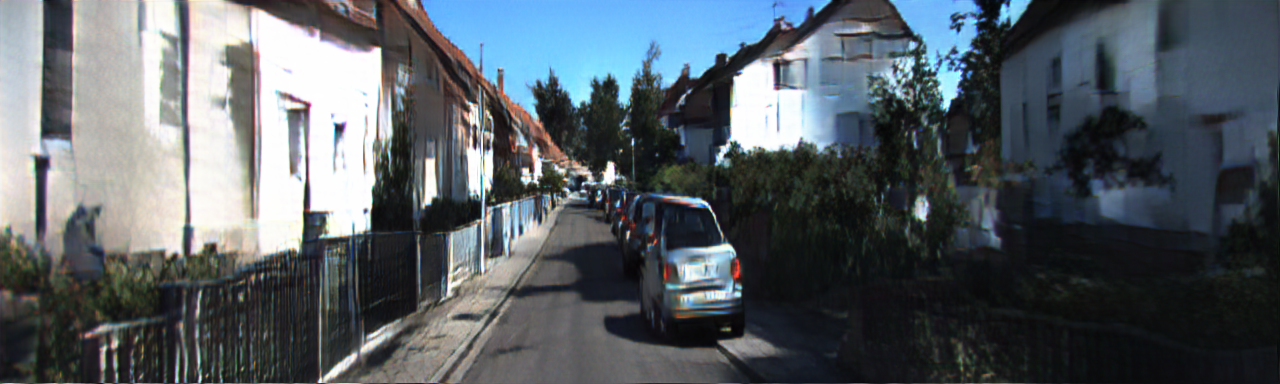}\par
    \includegraphics[width=1.0\linewidth]{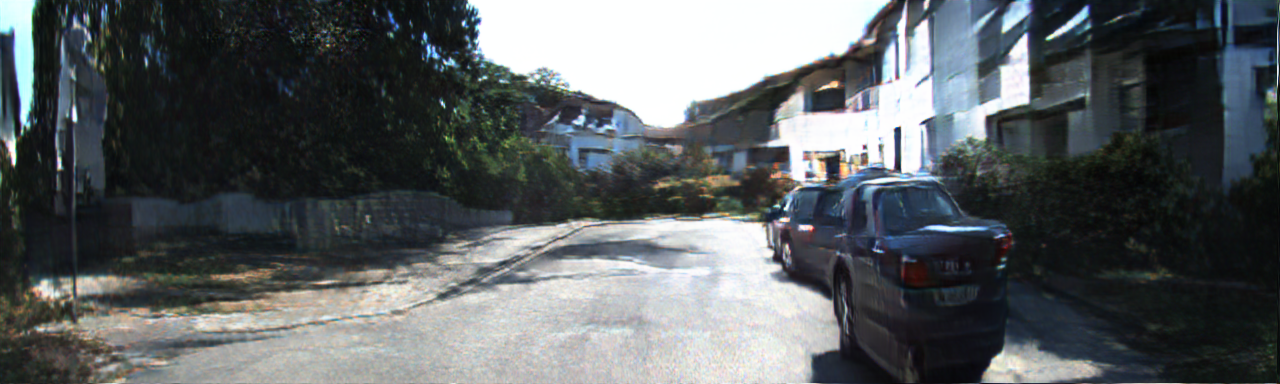}\par
\end{multicols}
\vspace{-0.95cm}
\begin{multicols}{3}
   \includegraphics[width=1.0\linewidth]{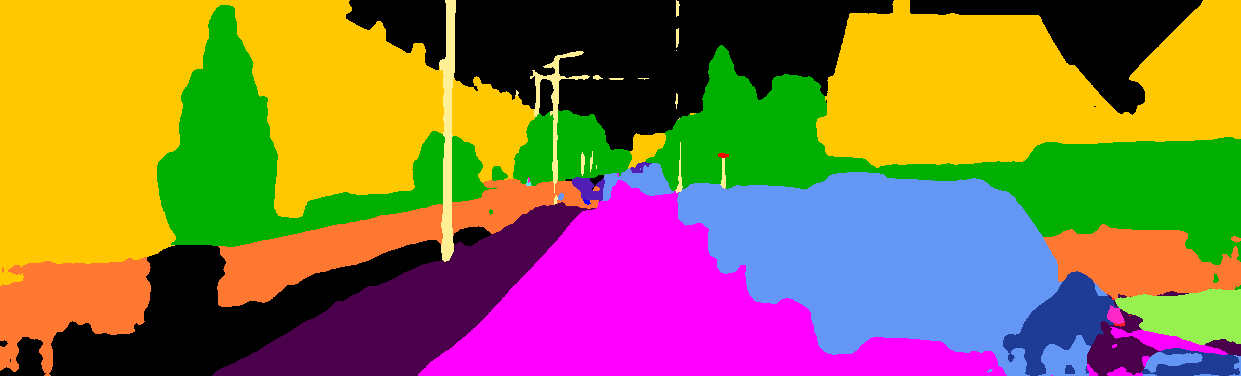} \par
    \includegraphics[width=1.0\linewidth]{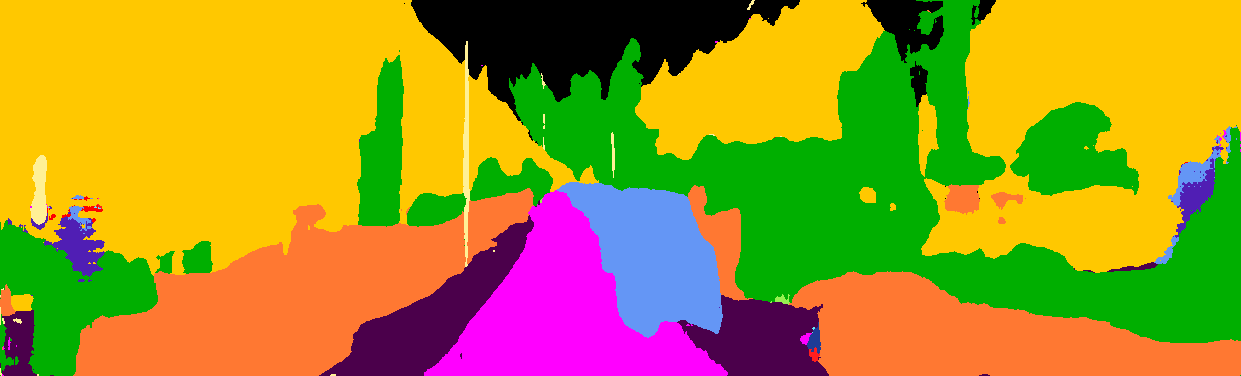}\par
    \includegraphics[width=1.0\linewidth]{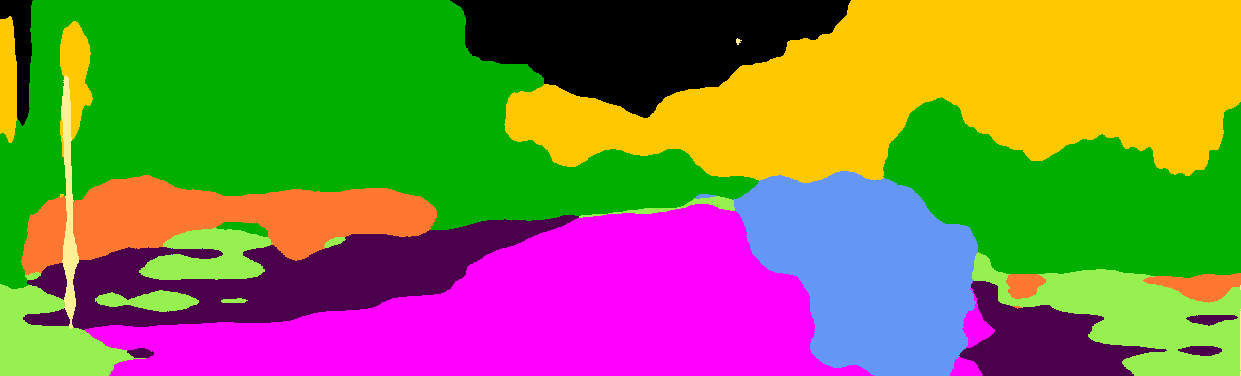}\par
\end{multicols}

\caption{Qualitative results on the \sk test set. The synthesized images are shown  at the top and the corresponding generated   segment maps are depicted at the bottom. From left to right, we have the ground-truth images, the TITAN-Net results (Ours)  and the Pix2Pix  outputs. Note that TITAN-Net and Pix2Pix are combined with Vid2Vid to translate segments to RGB images. 
}
\label{fig:segmap}
\end{figure*}

\begin{figure*}[!t]
\begin{multicols}{4}
    \includegraphics[width=1.0\linewidth]{./images/qualitative/test_set/real/realrgb/seq11_000030.png}\par 
    \includegraphics[width=1.0\linewidth]{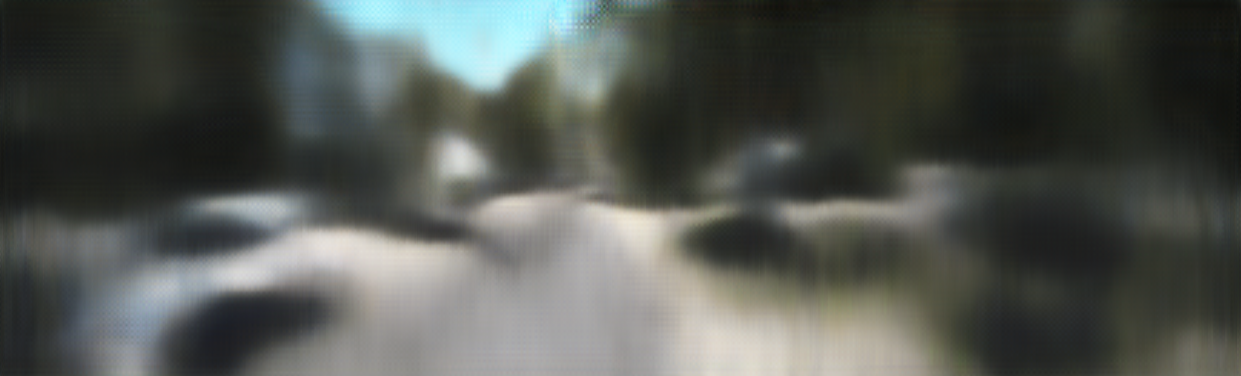}\par               
    \includegraphics[width=1.0\linewidth]{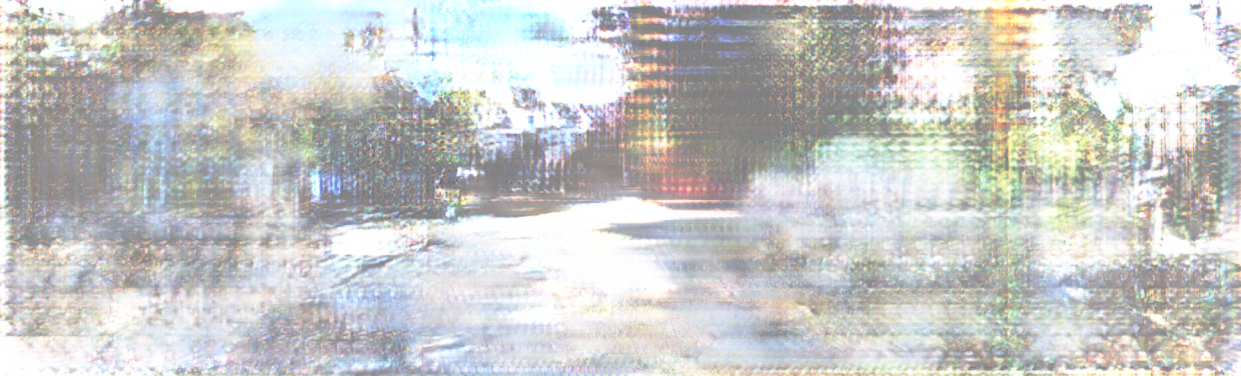}\par 
    \includegraphics[width=1.0\linewidth]{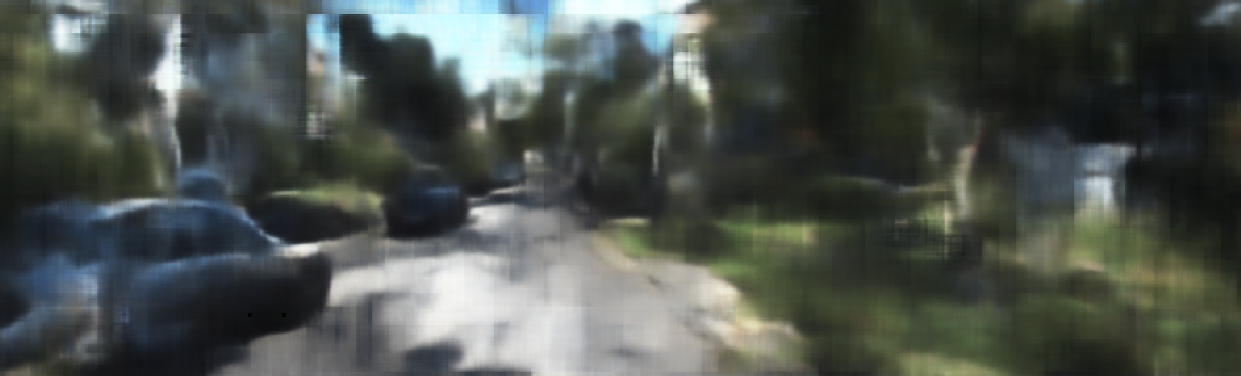}\par
\end{multicols}


\vspace{-0.8cm}
\begin{multicols}{4}
    \includegraphics[width=1.0\linewidth]{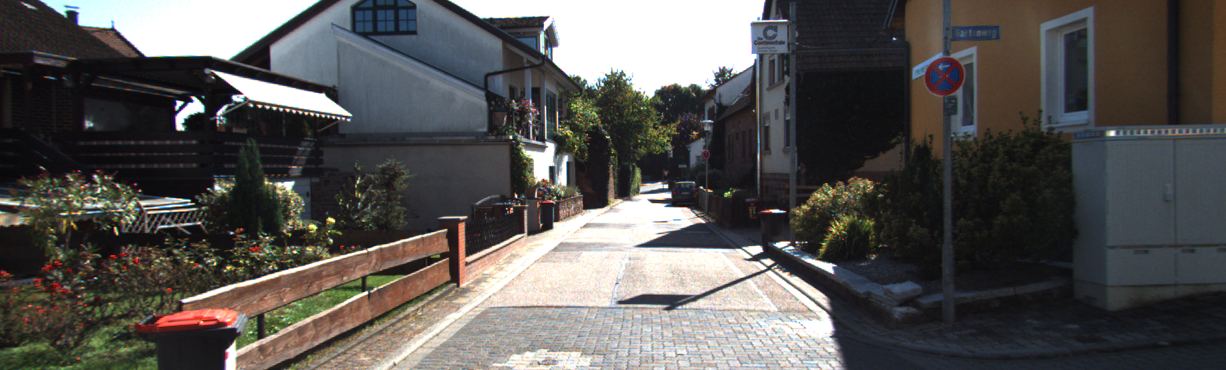}\par 
    \includegraphics[width=1.0\linewidth]{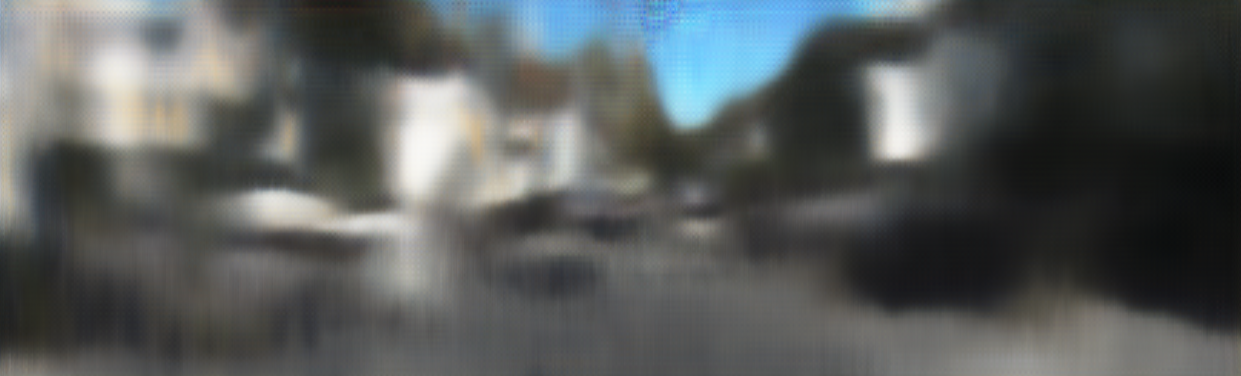}\par               
    \includegraphics[width=1.0\linewidth]{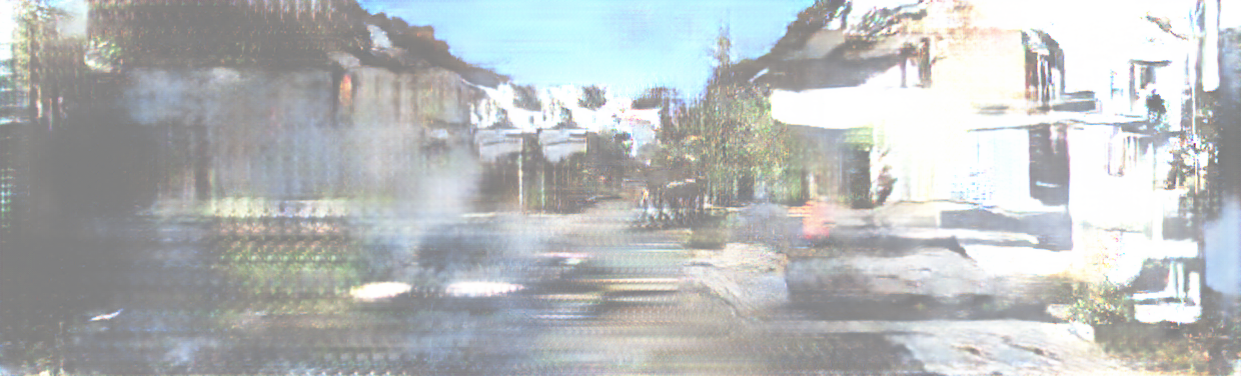}\par 
    \includegraphics[width=1.0\linewidth]{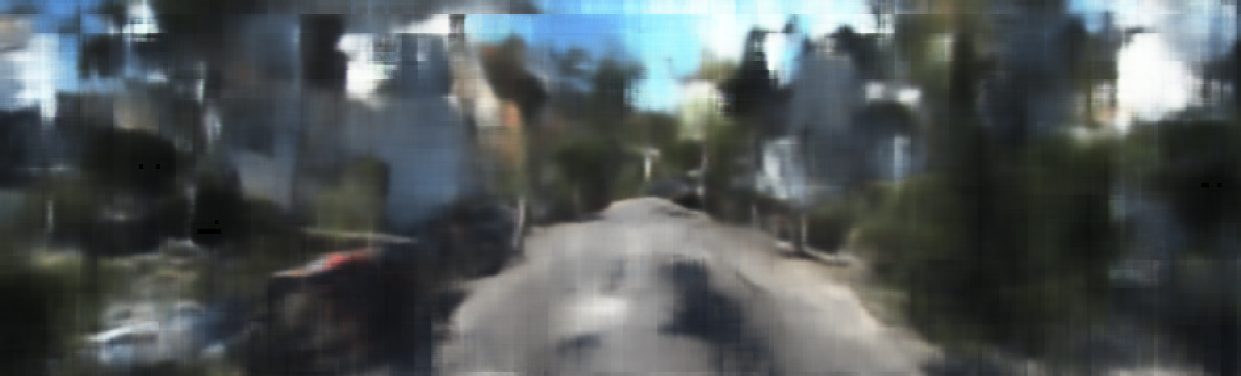}\par
\end{multicols}

\vspace{-0.8cm}
\begin{multicols}{4}
    \includegraphics[width=1.0\linewidth]{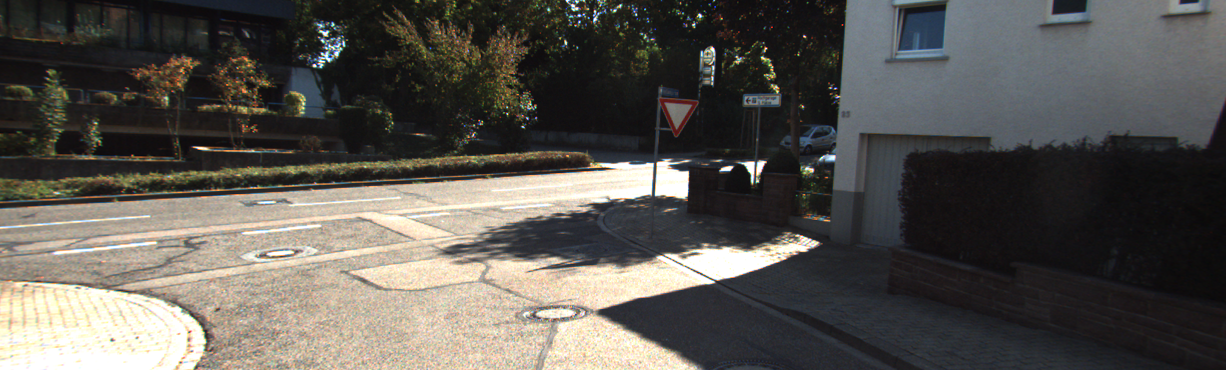}\par 
    \includegraphics[width=1.0\linewidth]{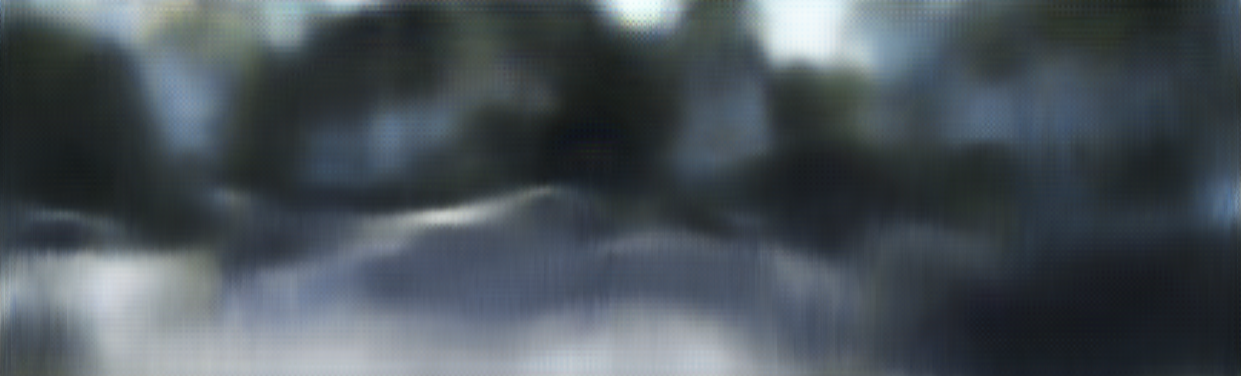}\par               
    \includegraphics[width=1.0\linewidth]{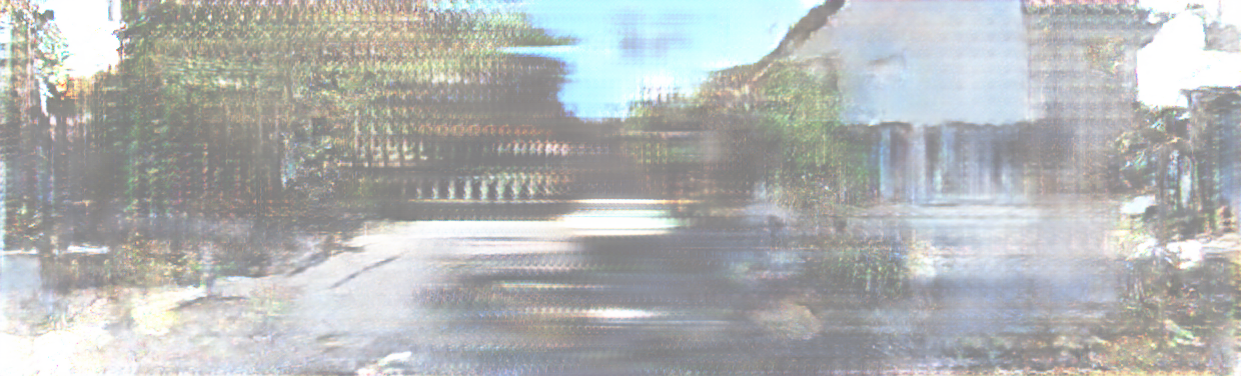}\par 
    \includegraphics[width=1.0\linewidth]{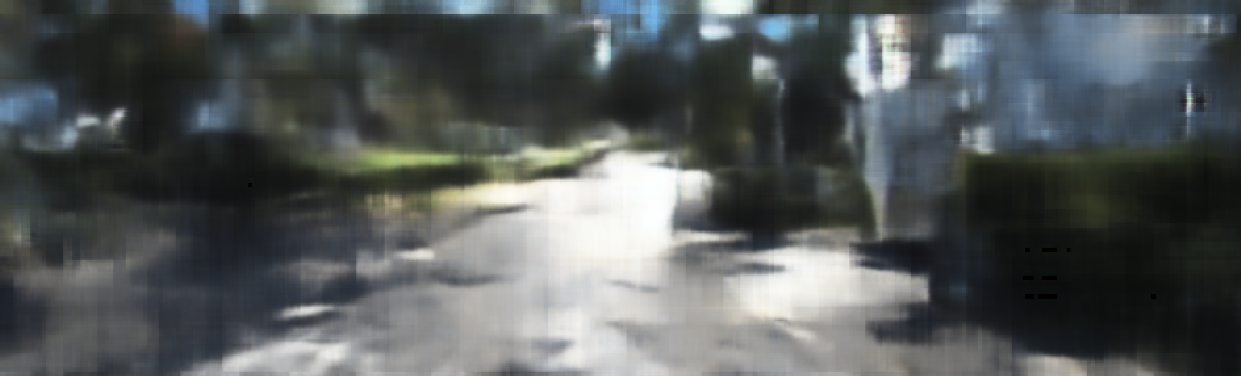}\par
\end{multicols}

\vspace{-0.8cm}
\begin{multicols}{4}
    \includegraphics[width=1.0\linewidth]{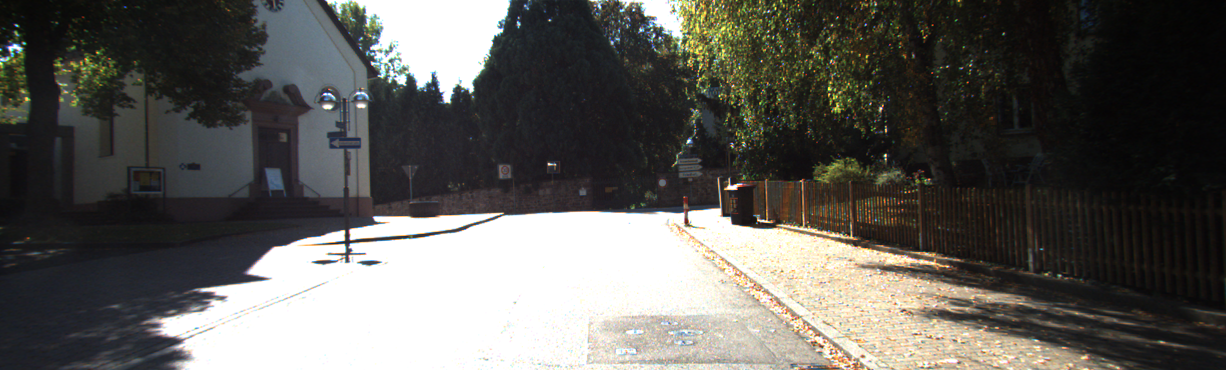}\par 
    \includegraphics[width=1.0\linewidth]{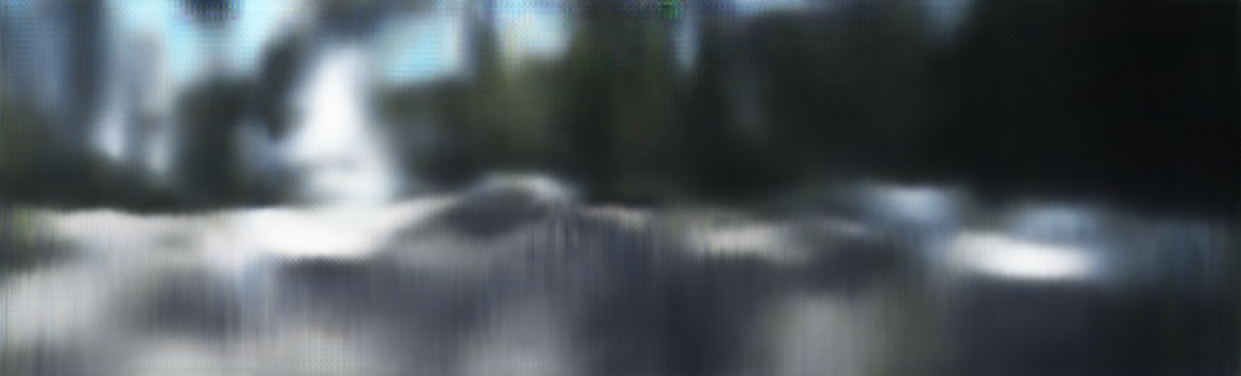}\par               
    \includegraphics[width=1.0\linewidth]{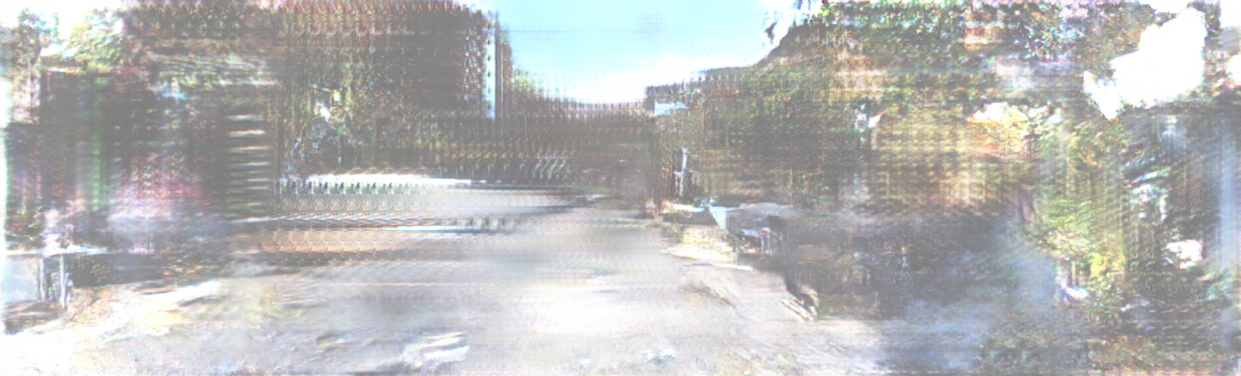}\par 
    \includegraphics[width=1.0\linewidth]{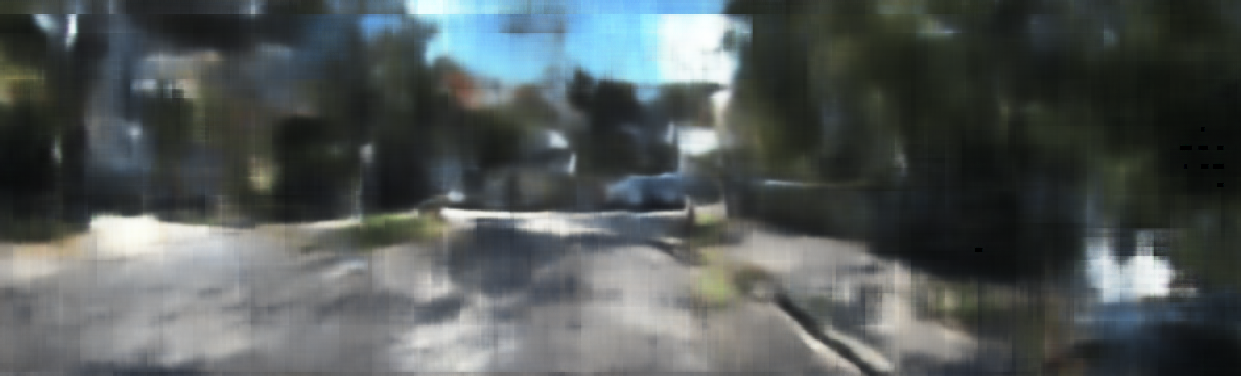}\par
\end{multicols}

\vspace{-0.8cm}
\begin{multicols}{4}
    \includegraphics[width=1.0\linewidth]{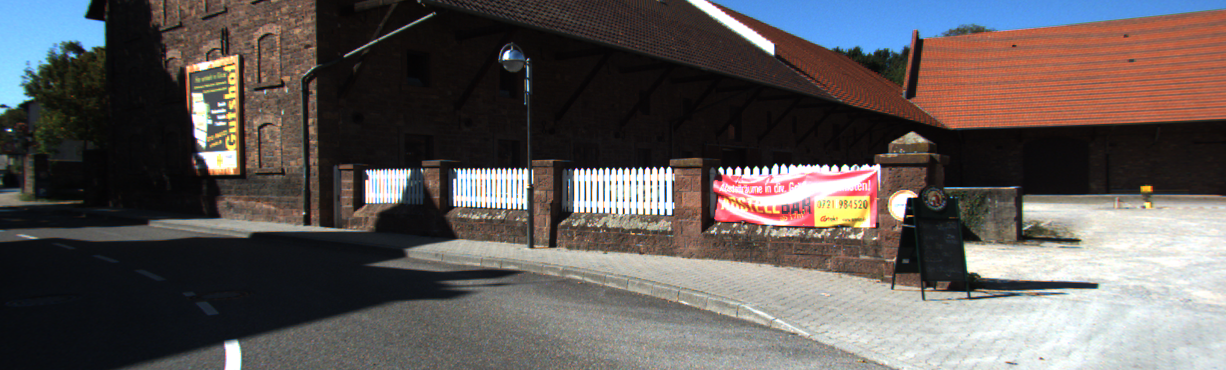}\par 
    \includegraphics[width=1.0\linewidth]{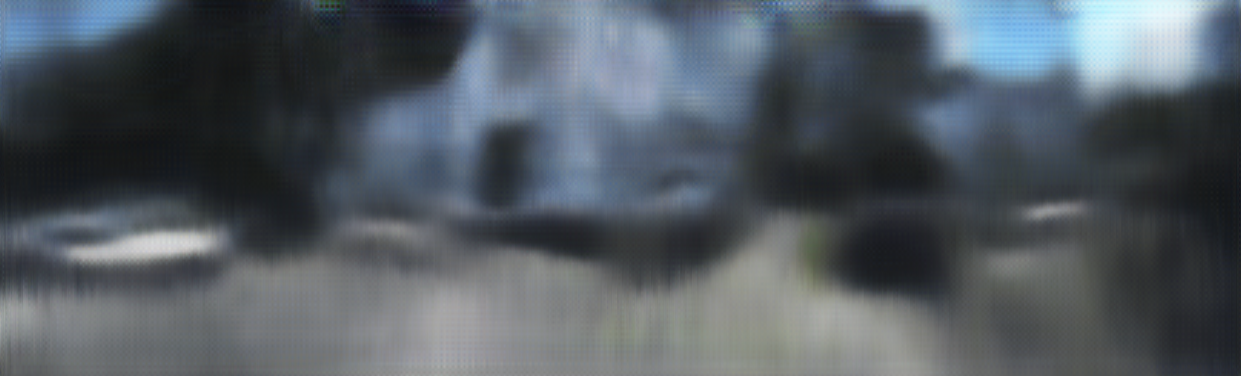}\par               
    \includegraphics[width=1.0\linewidth]{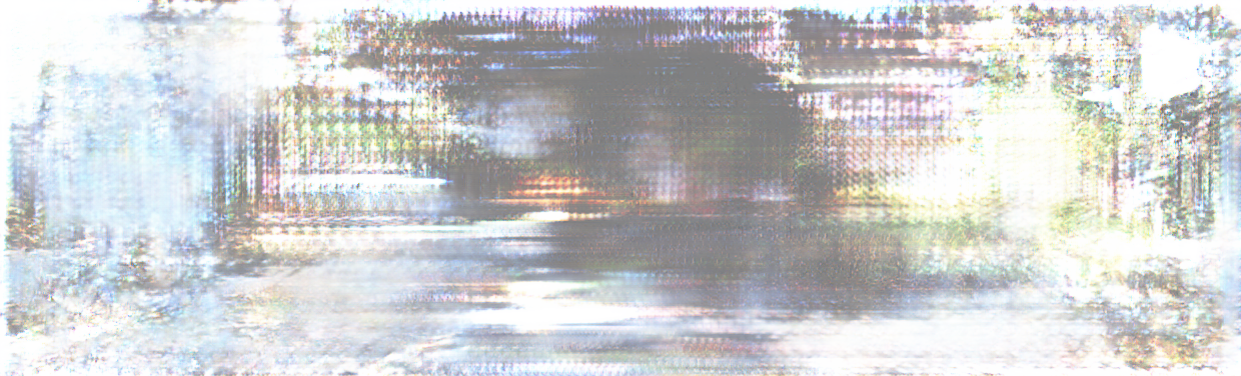}\par 
    \includegraphics[width=1.0\linewidth]{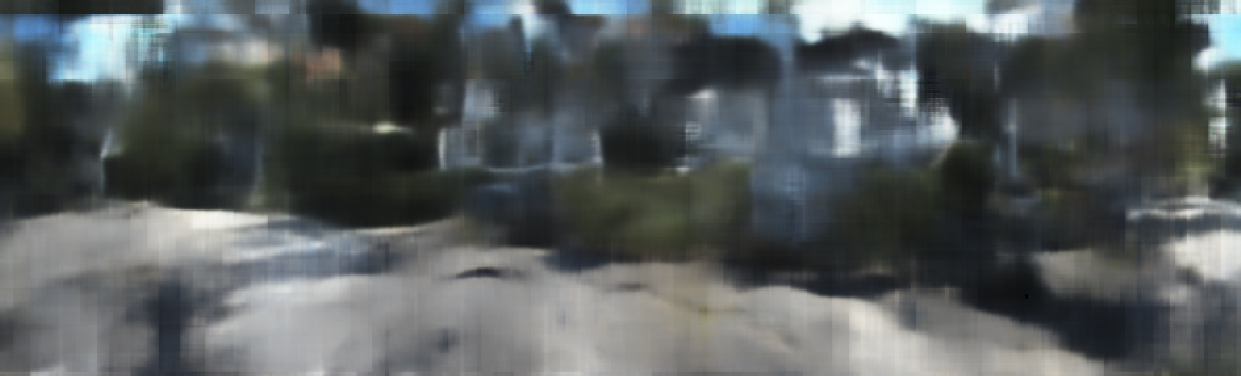}\par
\end{multicols}

\vspace{-0.8cm}
\begin{multicols}{4}
    \includegraphics[width=1.0\linewidth]{./images/qualitative/test_set/real/realrgb/seq11_000683.png}\par 
    \includegraphics[width=1.0\linewidth]{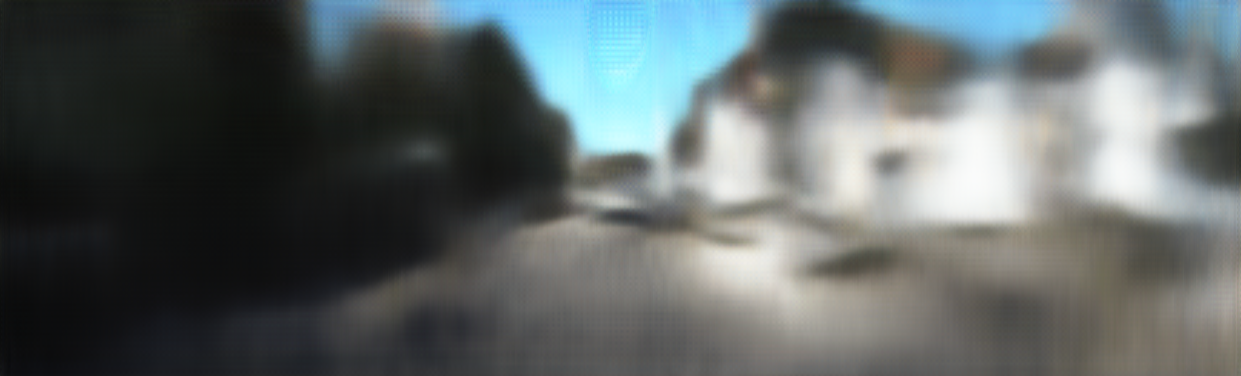}\par               
    \includegraphics[width=1.0\linewidth]{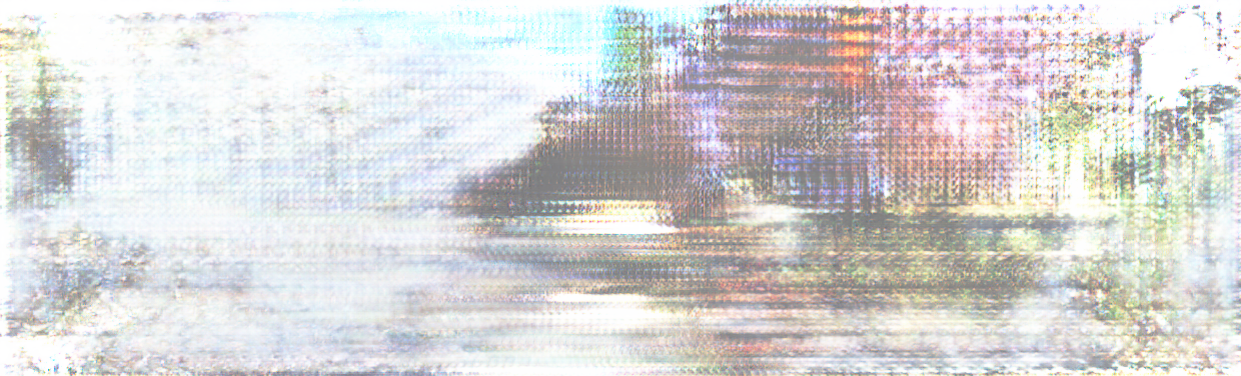}\par 
    \includegraphics[width=1.0\linewidth]{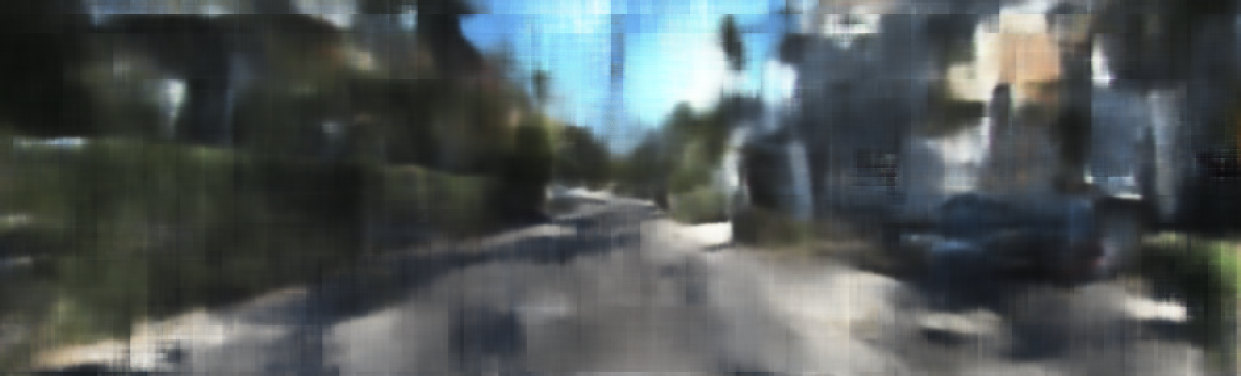}\par
\end{multicols}

\vspace{-0.8cm}
\begin{multicols}{4}
    \includegraphics[width=1.0\linewidth]{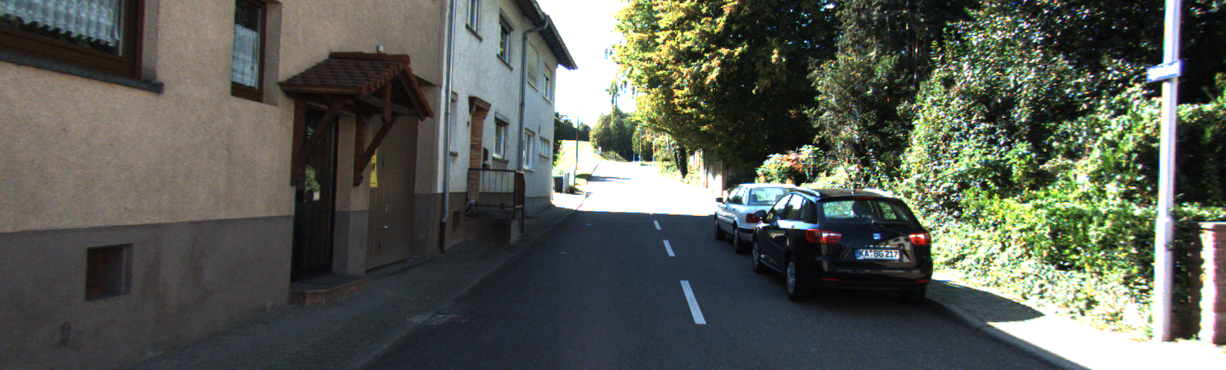}\par 
    \includegraphics[width=1.0\linewidth]{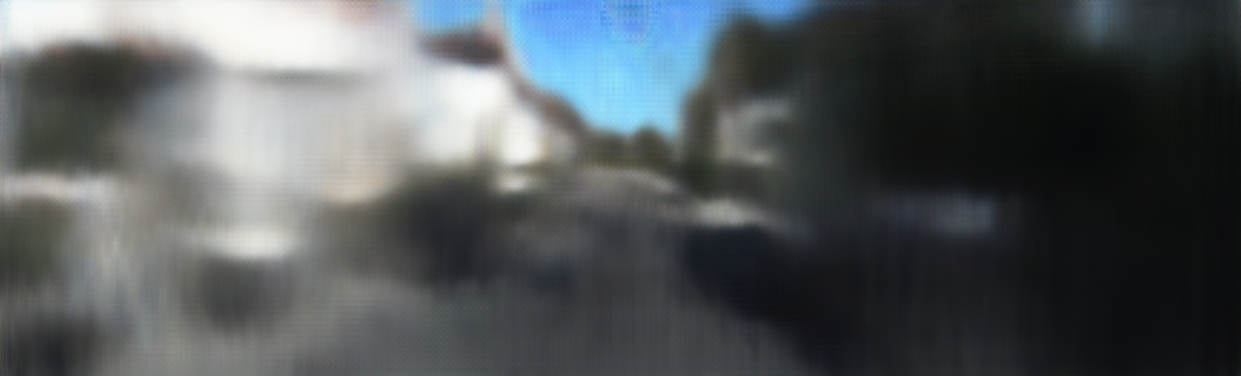}\par               
    \includegraphics[width=1.0\linewidth]{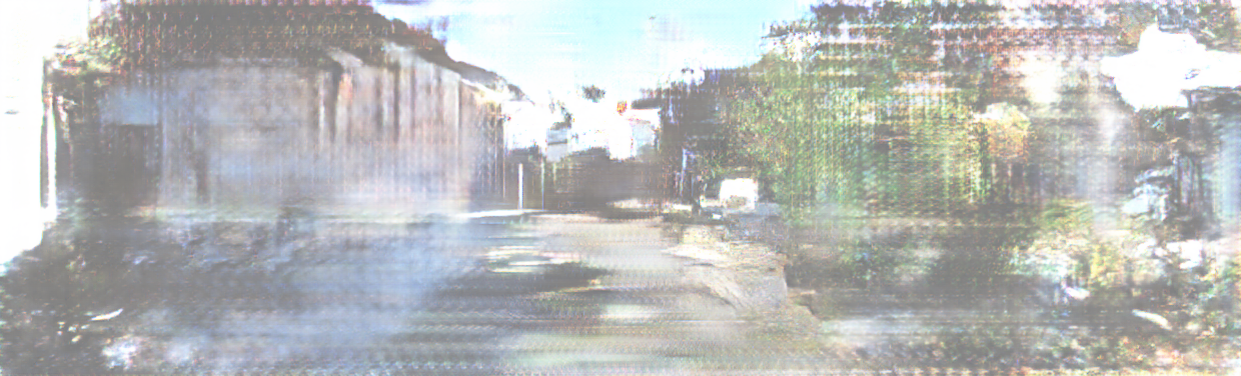}\par 
    \includegraphics[width=1.0\linewidth]{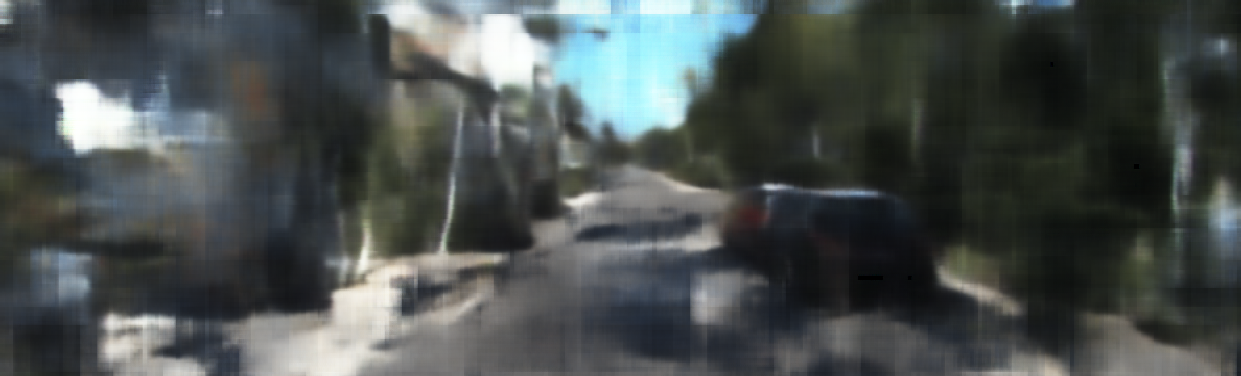}\par
\end{multicols}

\vspace{-0.8cm}
\begin{multicols}{4}
    \includegraphics[width=1.0\linewidth]{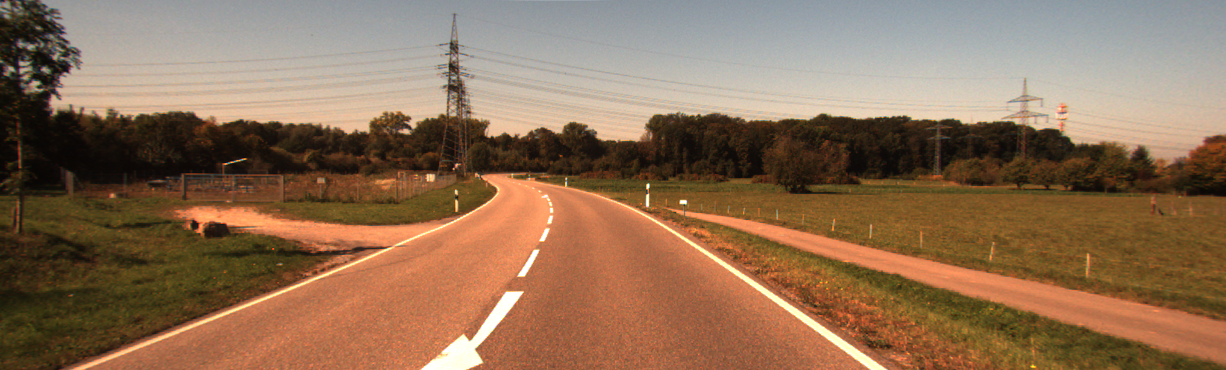}\par 
    \includegraphics[width=1.0\linewidth]{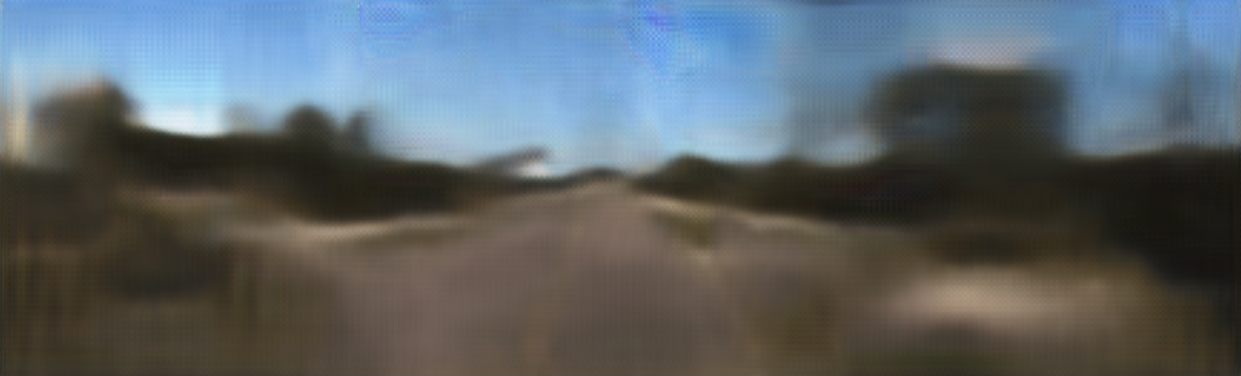}\par               
    \includegraphics[width=1.0\linewidth]{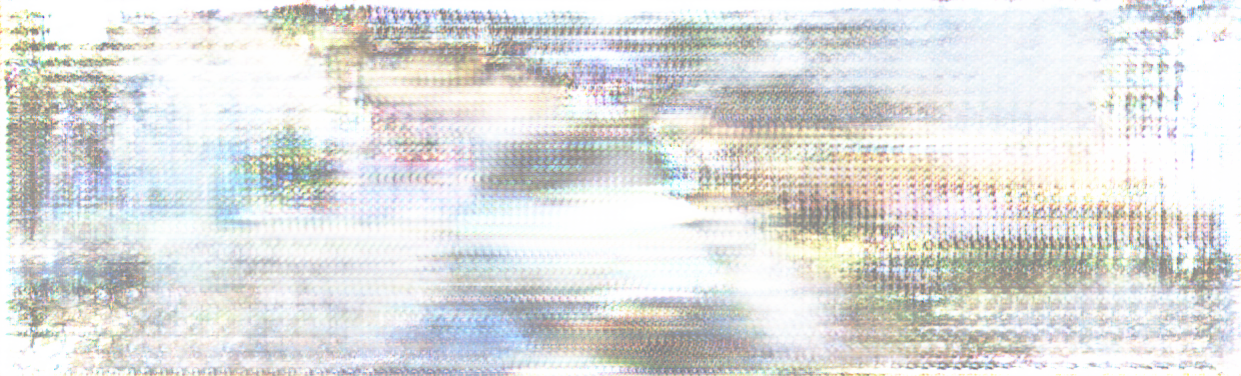}\par 
    \includegraphics[width=1.0\linewidth]{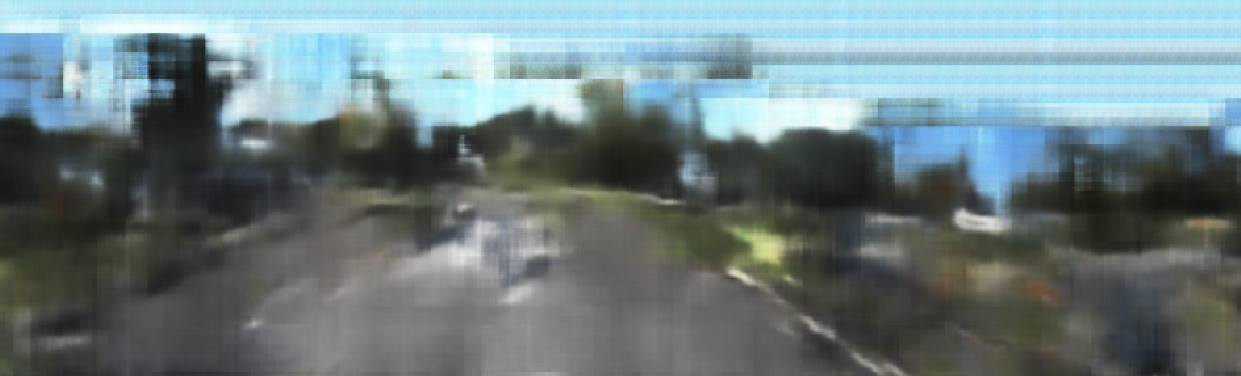}\par
\end{multicols}

\vspace{-0.8cm}
\begin{multicols}{4}
    \includegraphics[width=1.0\linewidth]{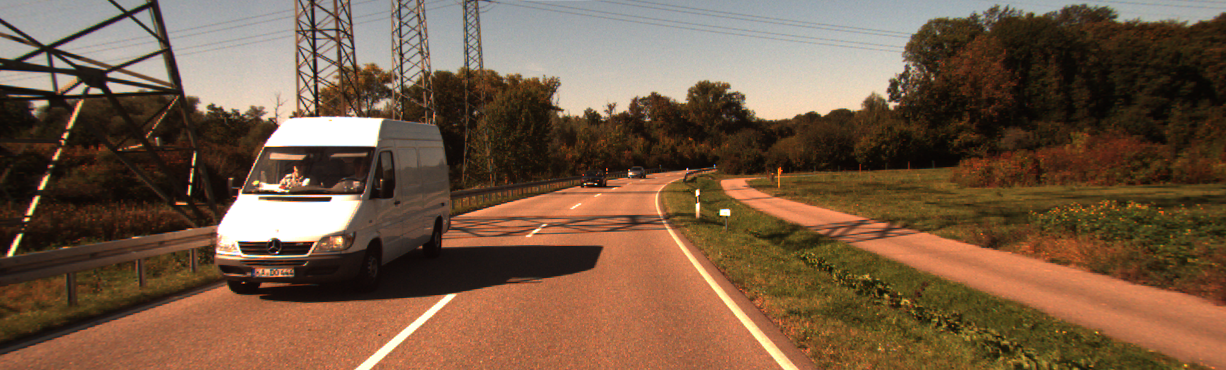}\par 
    \includegraphics[width=1.0\linewidth]{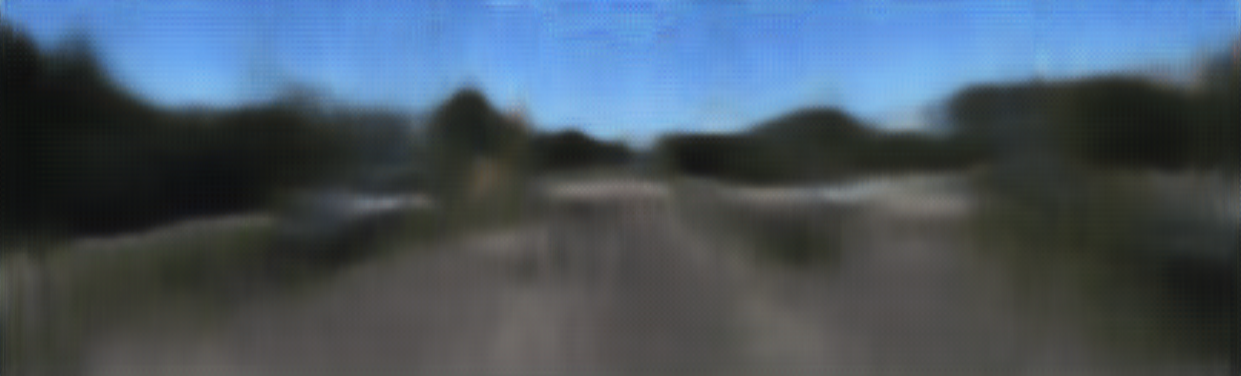}\par               
    \includegraphics[width=1.0\linewidth]{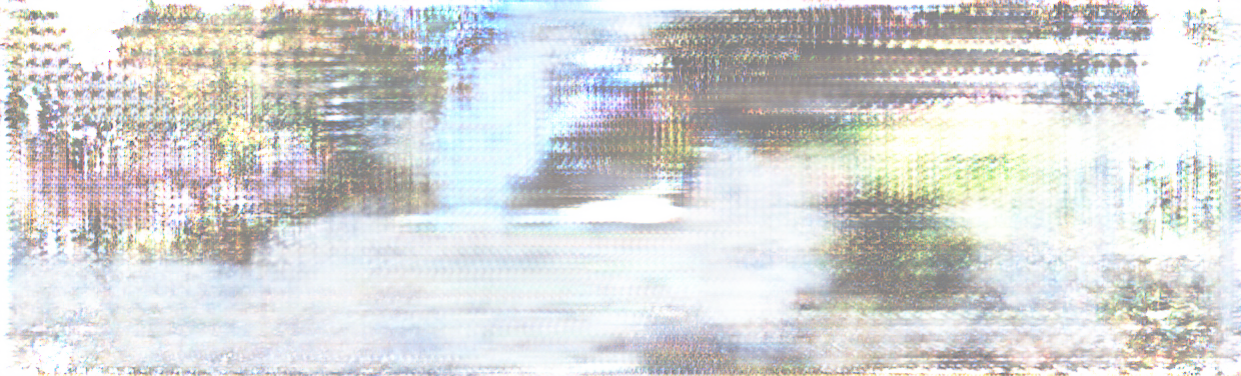}\par 
    \includegraphics[width=1.0\linewidth]{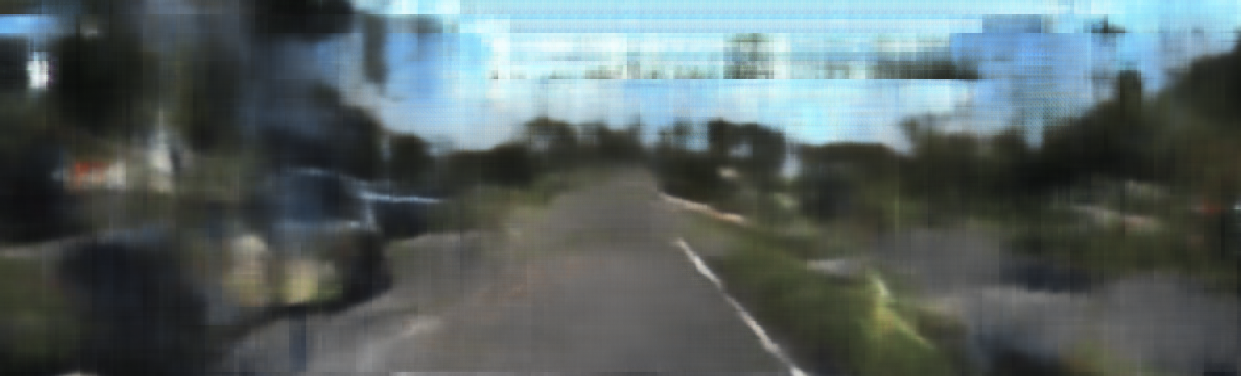}\par
\end{multicols}


\vspace{-0.8cm}
\begin{multicols}{4}
    \includegraphics[width=1.0\linewidth]{./images/qualitative/test_set/real/realrgb/seq12_000826.png}\par 
    \includegraphics[width=1.0\linewidth]{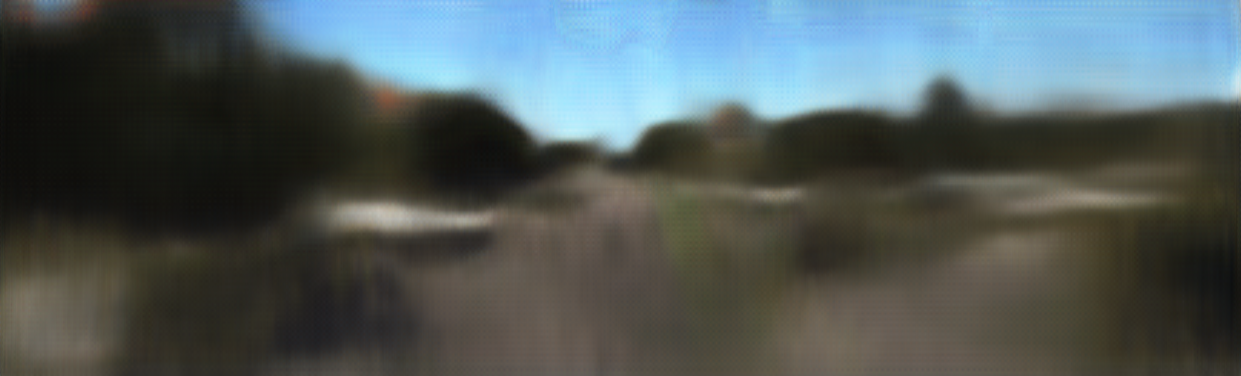}\par               
    \includegraphics[width=1.0\linewidth]{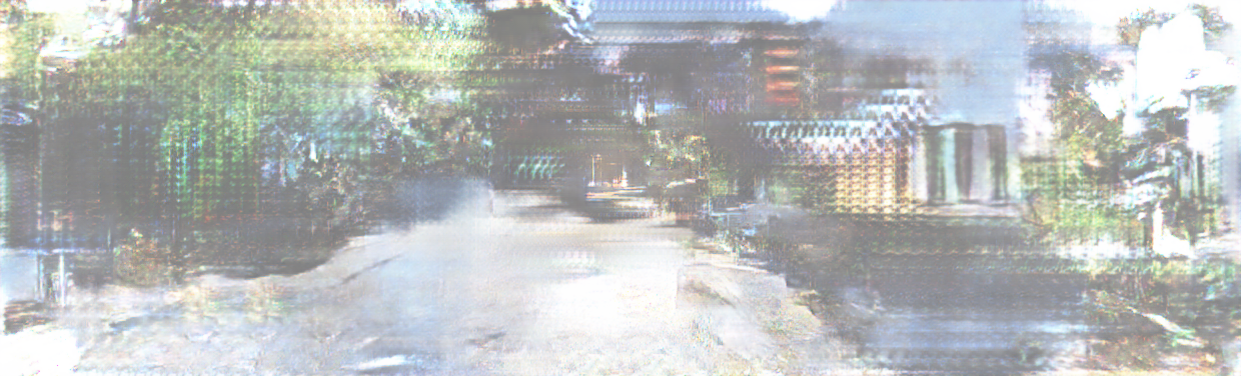}\par 
    \includegraphics[width=1.0\linewidth]{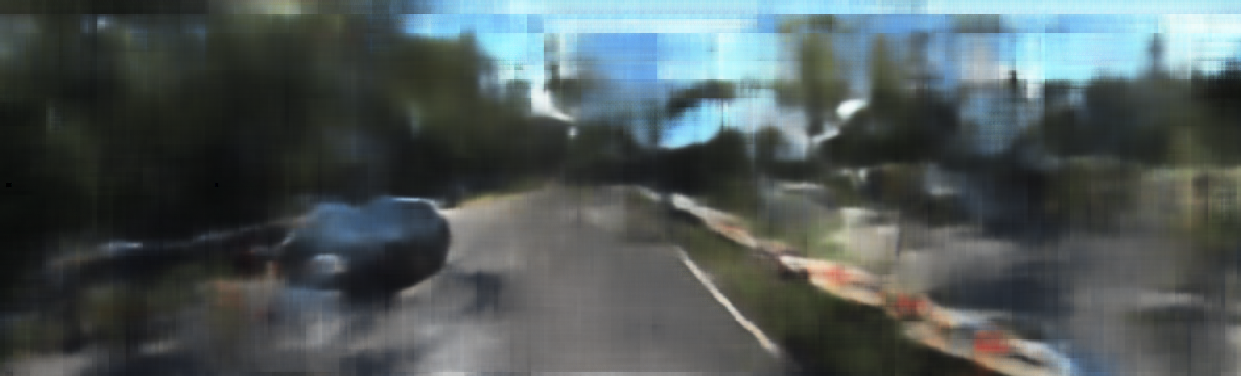}\par
\end{multicols}

\vspace{-0.8cm}
\begin{multicols}{4}
    \includegraphics[width=1.0\linewidth]{./images/qualitative/test_set/real/realrgb/seq15_000009.png}\par 
    \includegraphics[width=1.0\linewidth]{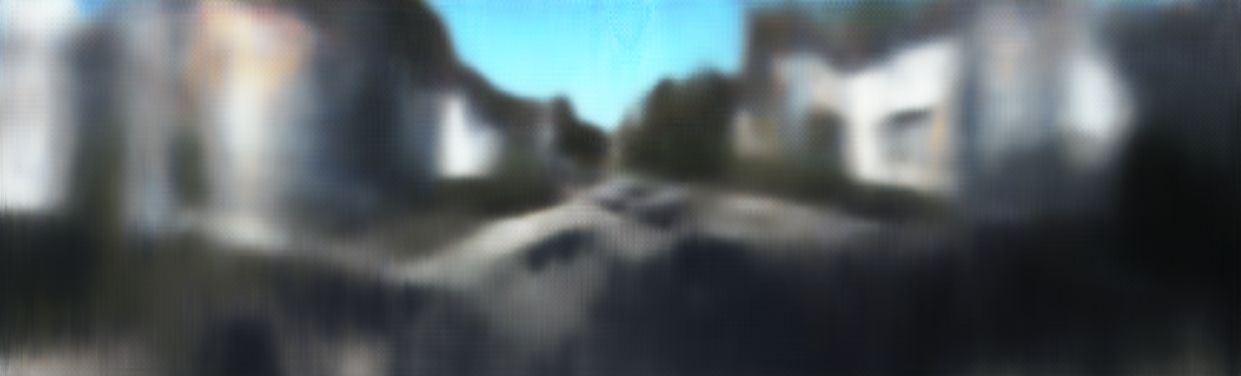}\par               
    \includegraphics[width=1.0\linewidth]{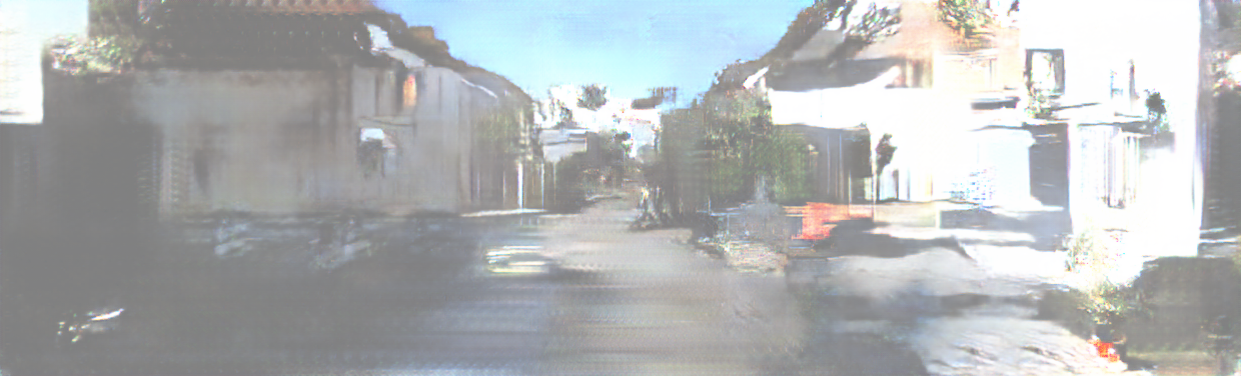}\par 
    \includegraphics[width=1.0\linewidth]{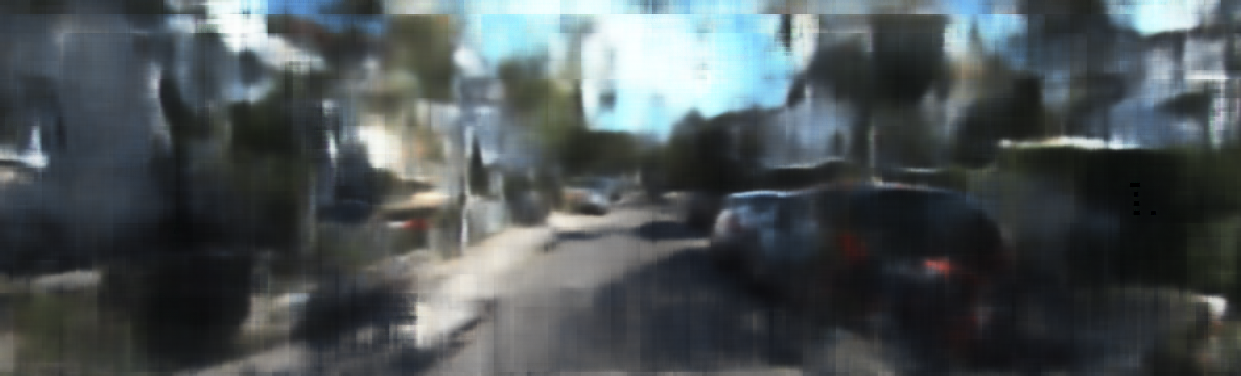}\par
\end{multicols}

\vspace{-0.8cm}
\begin{multicols}{4}
    \includegraphics[width=1.0\linewidth]{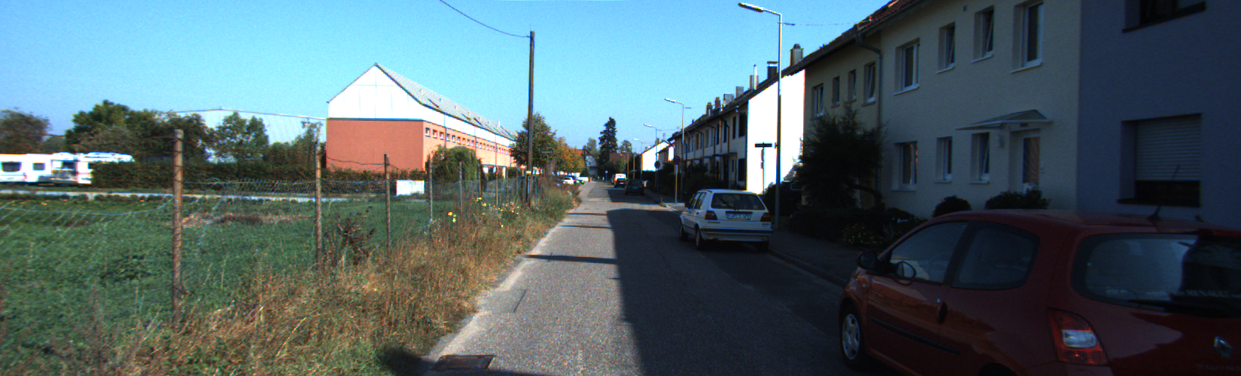}\par 
    \includegraphics[width=1.0\linewidth]{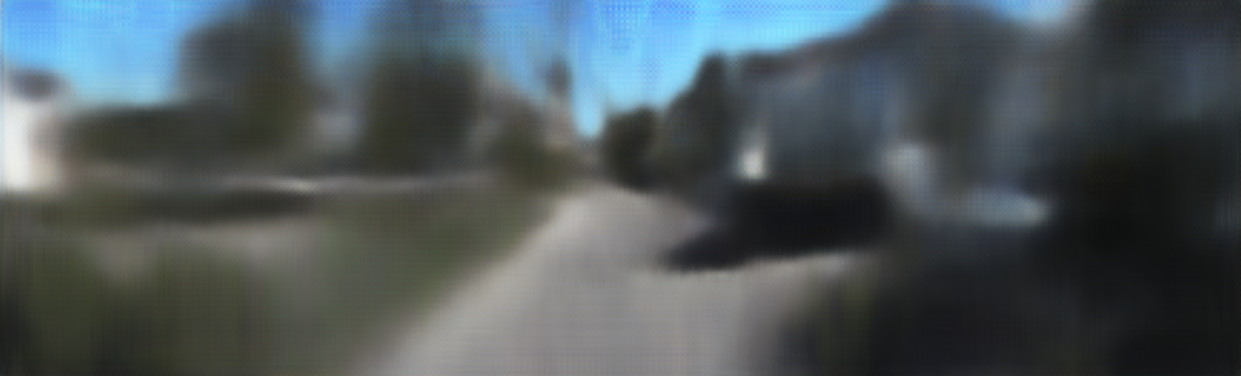}\par               
    \includegraphics[width=1.0\linewidth]{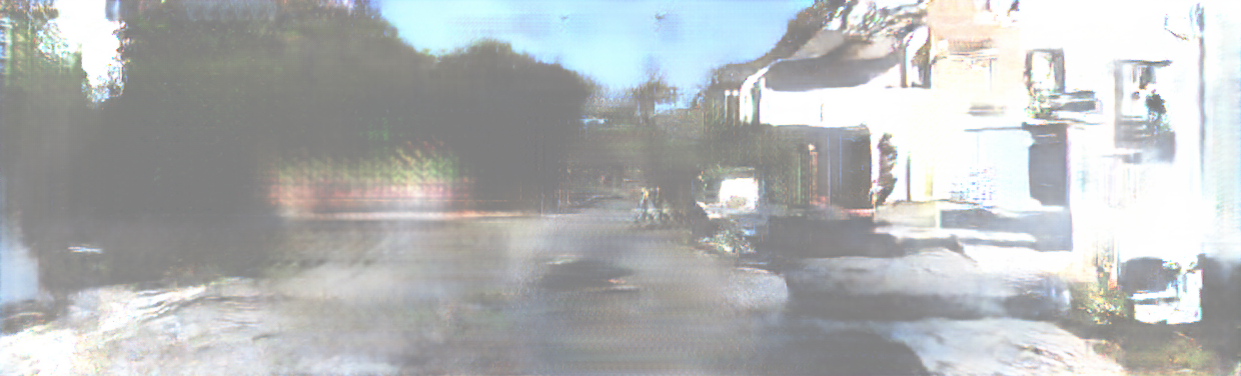}\par 
    \includegraphics[width=1.0\linewidth]{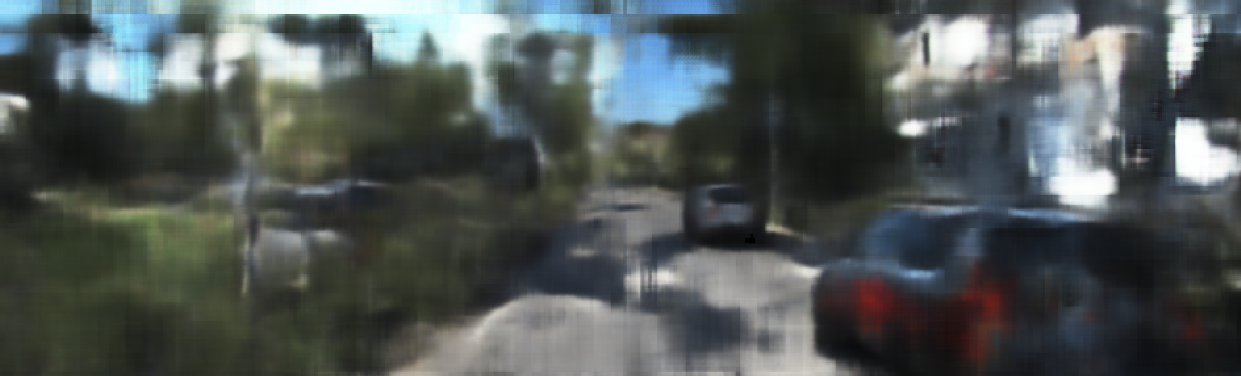}\par
\end{multicols}

\vspace{-0.8cm}
\begin{multicols}{4}
    \includegraphics[width=1.0\linewidth]{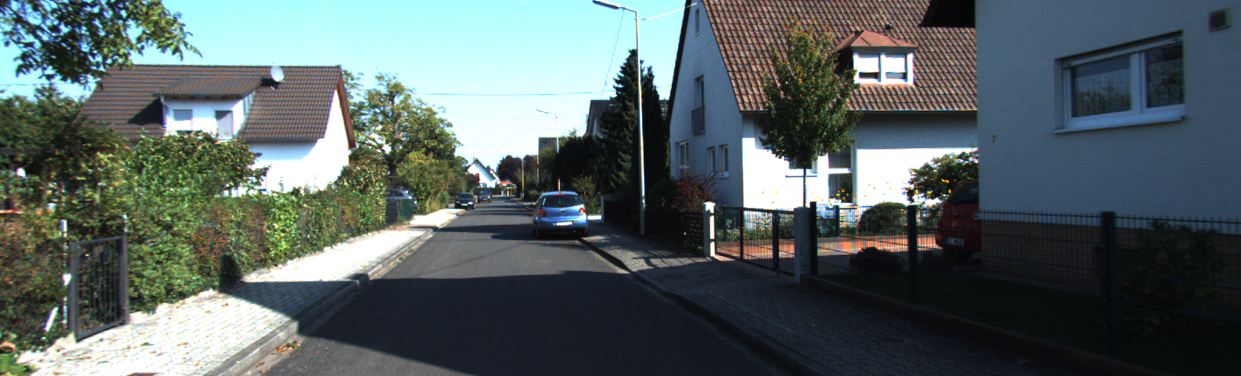}\par 
    \includegraphics[width=1.0\linewidth]{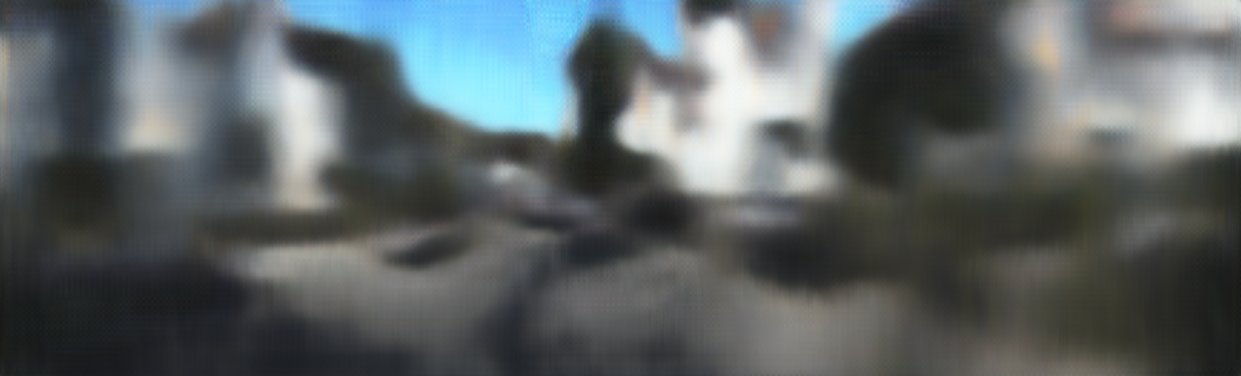}\par               
    \includegraphics[width=1.0\linewidth]{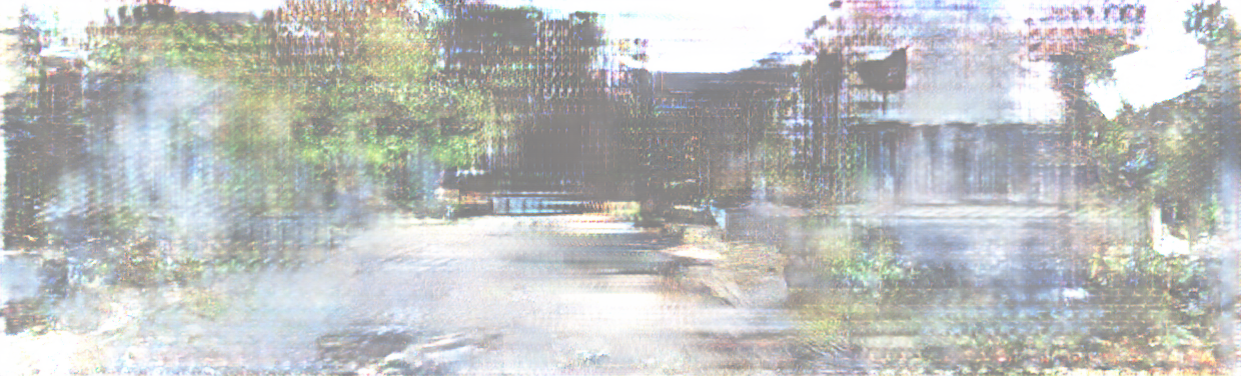}\par 
    \includegraphics[width=1.0\linewidth]{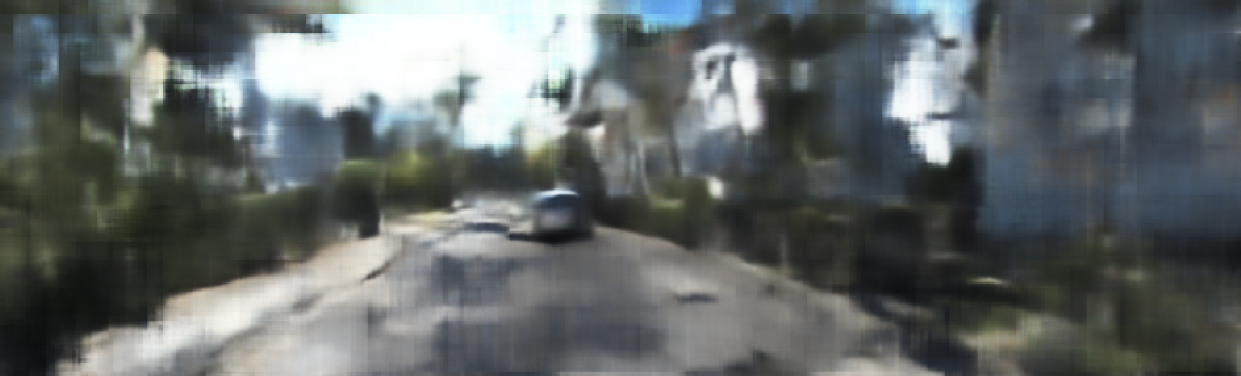}\par
\end{multicols}

\caption{Sample images generated directly from the projected point cloud images without employing the segmentation maps. From left to right we have the ground-truth image and the results from TITAN-Net (Ours), Pix2Pix, and SC-UNET.  
}
\label{fig:base}
\end{figure*}

\begin{figure*}[!t]
\centering 
\includegraphics[width=1.0\linewidth, height=0.12\linewidth]{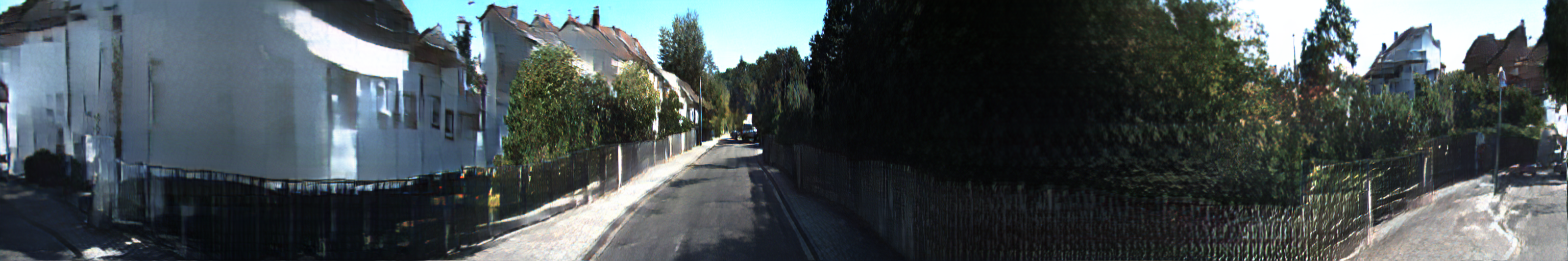}\par
\vspace{0.05cm}
\includegraphics[width=1.0\linewidth, height=0.12\linewidth]{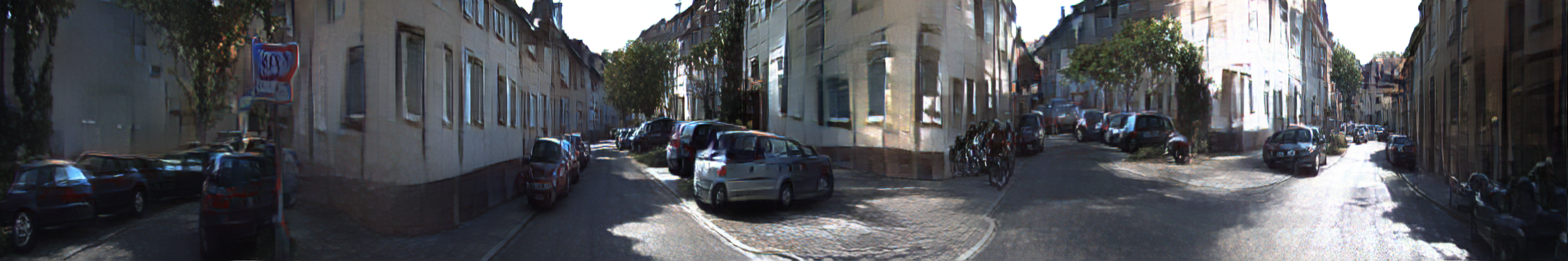}\par
\vspace{0.05cm}
\includegraphics[width=1.0\linewidth, height=0.12\linewidth]{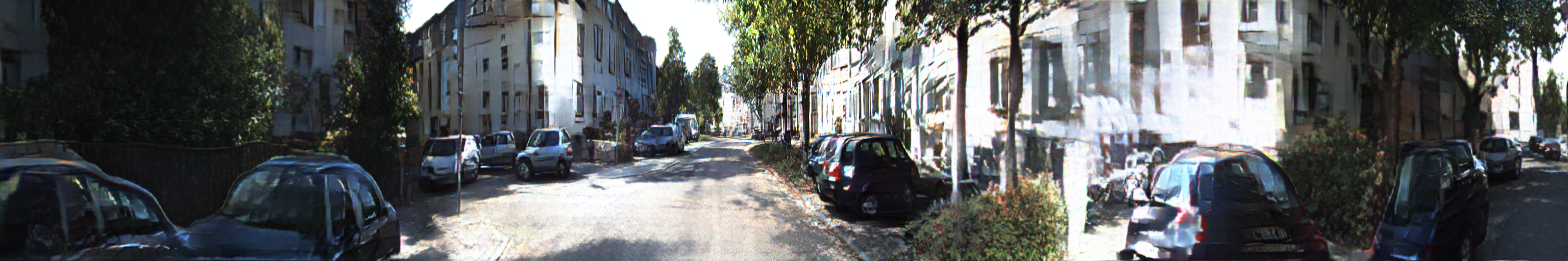}\par   
\vspace{0.05cm}
\includegraphics[width=1.0\linewidth, height=0.12\linewidth]{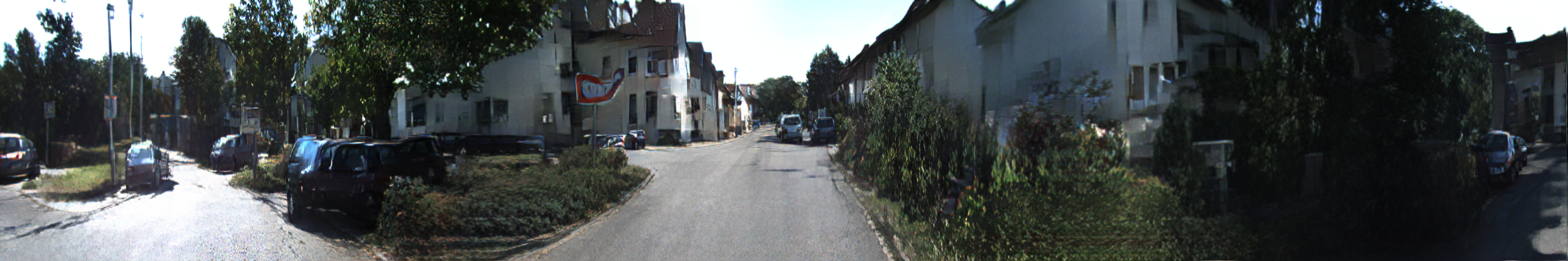}\par    
\vspace{0.05cm}
\includegraphics[width=1.0\linewidth, height=0.12\linewidth]{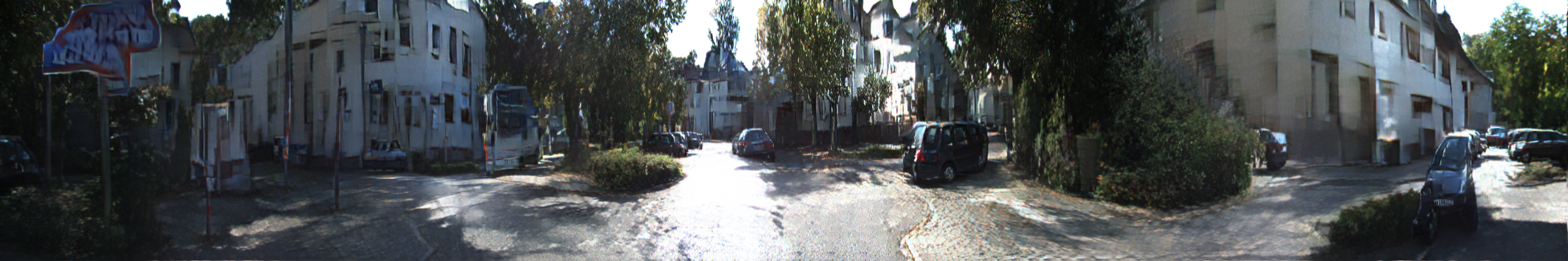}\par  
\vspace{0.05cm}
\includegraphics[width=1.0\linewidth, height=0.12\linewidth]{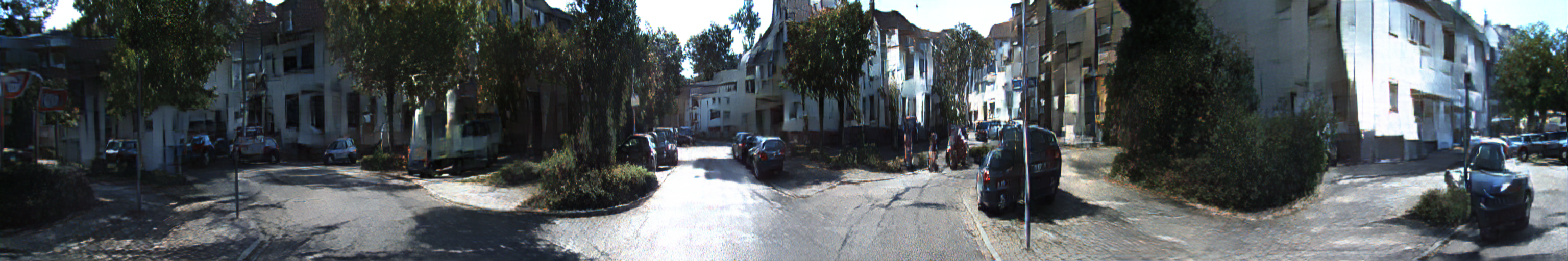}\par     
\vspace{0.05cm}
\includegraphics[width=1.0\linewidth, height=0.12\linewidth]{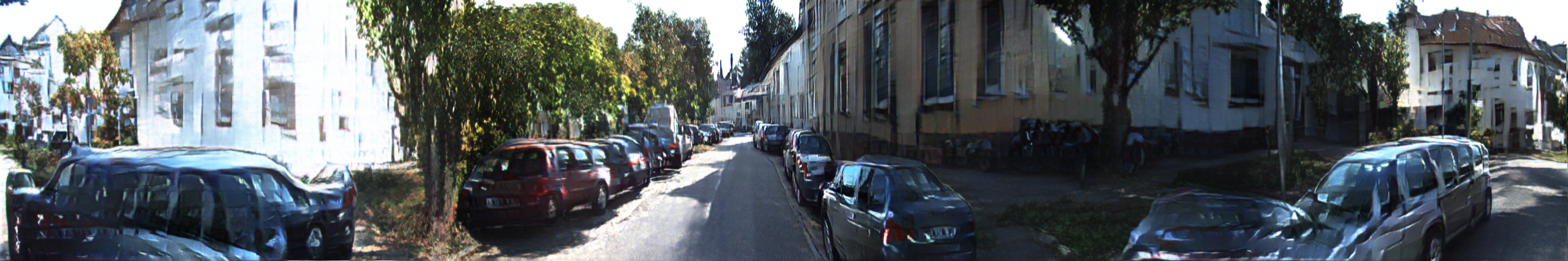}\par
\vspace{0.05cm}
\includegraphics[width=1.0\linewidth, height=0.12\linewidth]{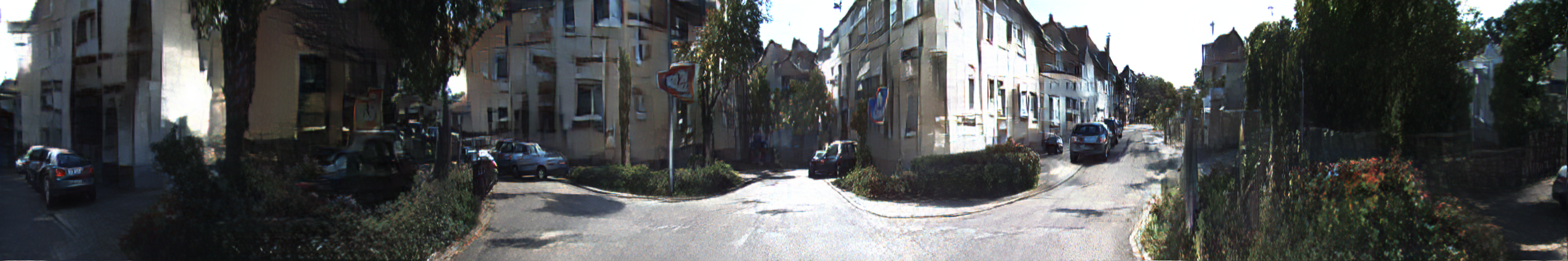}\par    
\vspace{0.05cm}
\includegraphics[width=1.0\linewidth, height=0.12\linewidth]{./images/qualitative/seq13_360/001962.png}\par   
\vspace{0.05cm}
\includegraphics[width=1.0\linewidth, height=0.12\linewidth]{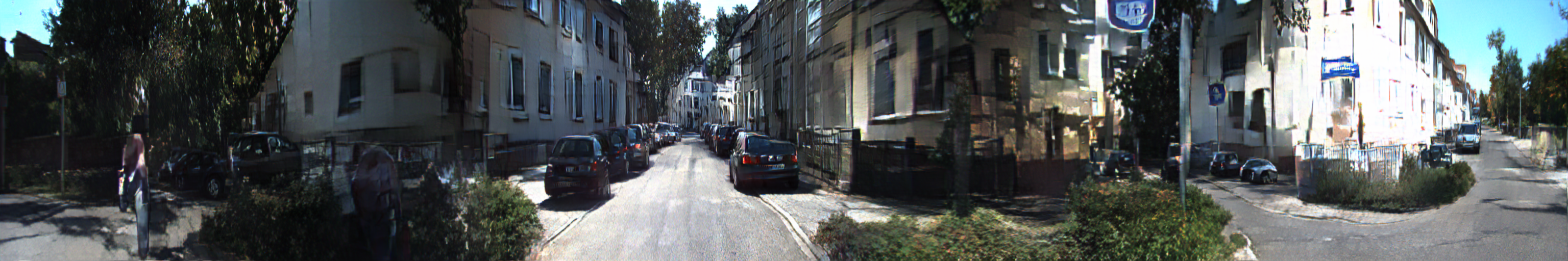}\par    
\caption{Panoramic images  synthesized by our proposed TITAN-Net model on the \sk test set.}
\label{fig:360}
\end{figure*}

\begin{figure*}[!b]
\centering

   \vspace{-0.8cm}
\begin{multicols}{3}
\includegraphics[width=1\linewidth]{./images/qualitative/test_set/real/realrgb/seq16_001594.png}\par 
\includegraphics[width=1\linewidth]{./images/qualitative/test_set/titan/gen/seq16_001594.png}\par     
\includegraphics[width=1\linewidth]{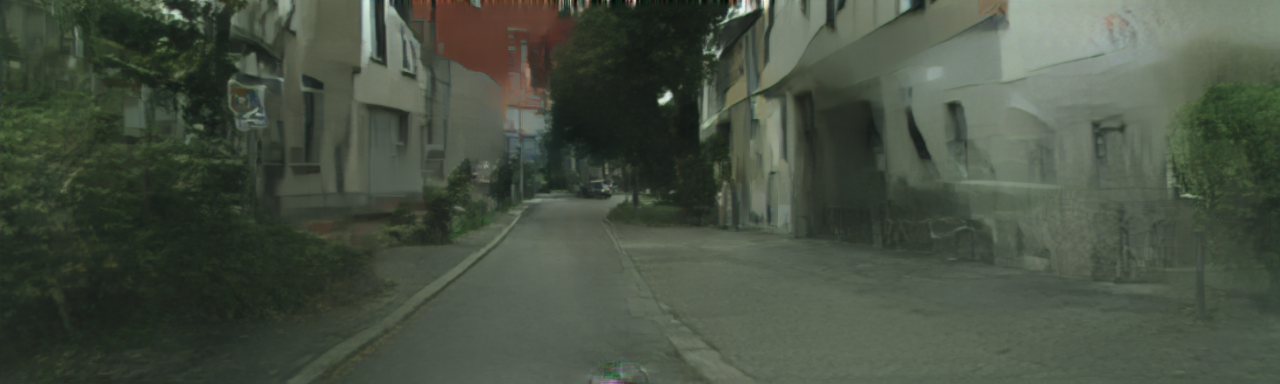}\par 
  \end{multicols}

   \vspace{-0.8cm}
\begin{multicols}{3}
\includegraphics[width=1\linewidth]{./images/qualitative/test_set/real/realrgb/seq15_000009.png}\par 
\includegraphics[width=1\linewidth]{./images/qualitative/test_set/titan/gen/seq15_000009.png}\par     
\includegraphics[width=1\linewidth]{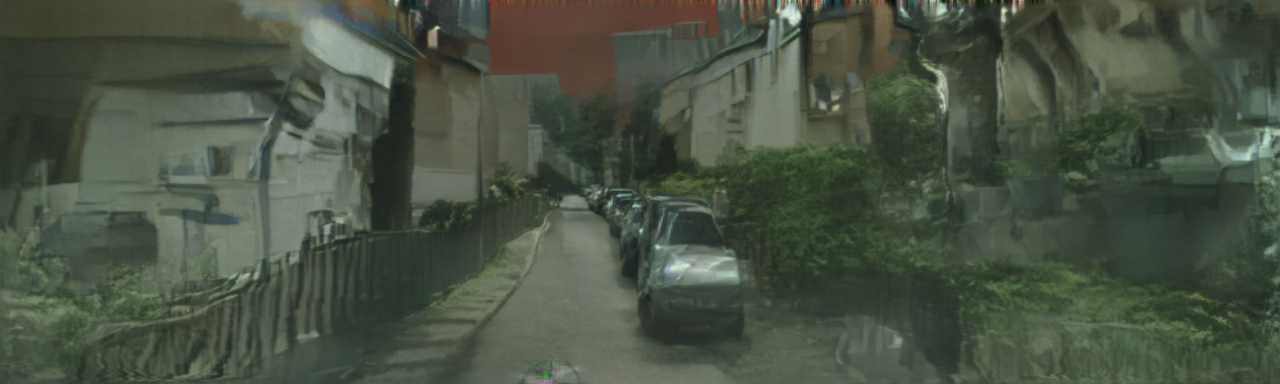}\par 
  \end{multicols}

   \vspace{-0.8cm}
\begin{multicols}{3}
\includegraphics[width=1\linewidth]{./images/qualitative/test_set/real/realrgb/seq12_000826.png}\par 
\includegraphics[width=1\linewidth]{./images/qualitative/test_set/titan/gen/seq12_000826.png}\par     
\includegraphics[width=1\linewidth]{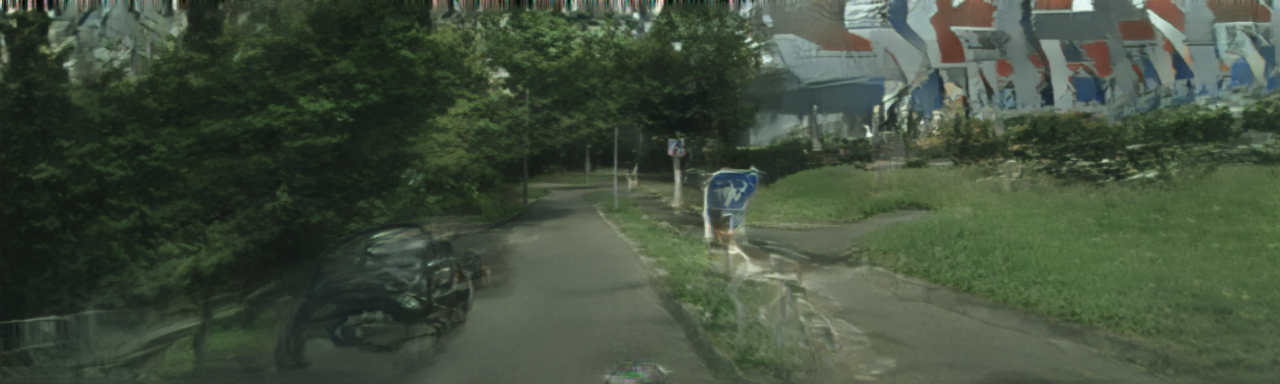}\par 
  \end{multicols}

    \vspace{-0.8cm}
\begin{multicols}{3}
\includegraphics[width=1\linewidth]{./images/qualitative/test_set/real/realrgb/seq11_000683.png}\par 
\includegraphics[width=1\linewidth]{./images/qualitative/test_set/titan/gen/seq11_000683.png}\par     
\includegraphics[width=1\linewidth]{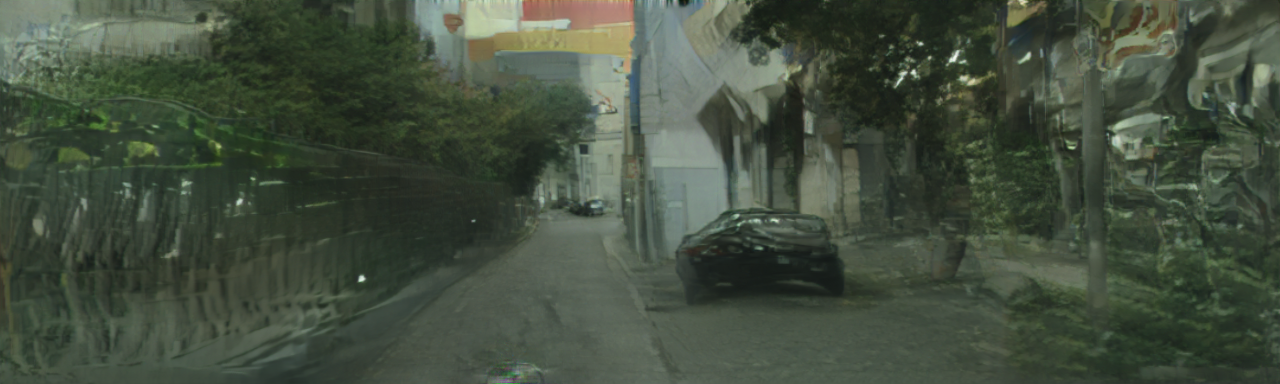}\par 
  \end{multicols}

  \vspace{-0.8cm}
\begin{multicols}{3}
\includegraphics[width=1\linewidth]{./images/qualitative/test_set/real/realrgb/seq11_000030.png}\par 
\includegraphics[width=1\linewidth]{./images/qualitative/test_set/titan/gen/seq11_000030.png}\par     
\includegraphics[width=1\linewidth]{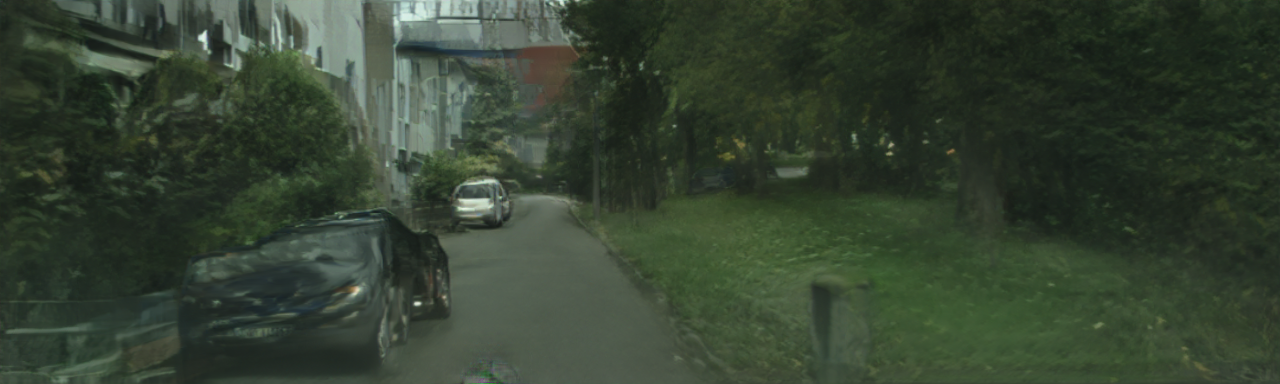}\par 
\end{multicols}

 \vspace{-0.8cm}
\begin{multicols}{3}
\includegraphics[width=1\linewidth]{./images/qualitative/test_set/real/realrgb/seq11_000812.png}\par 
\includegraphics[width=1\linewidth]{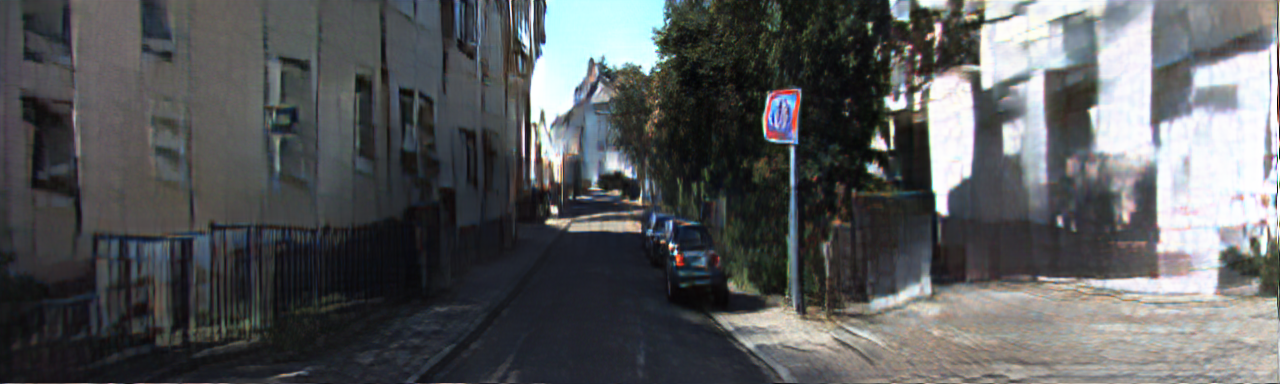}\par     
\includegraphics[width=1\linewidth]{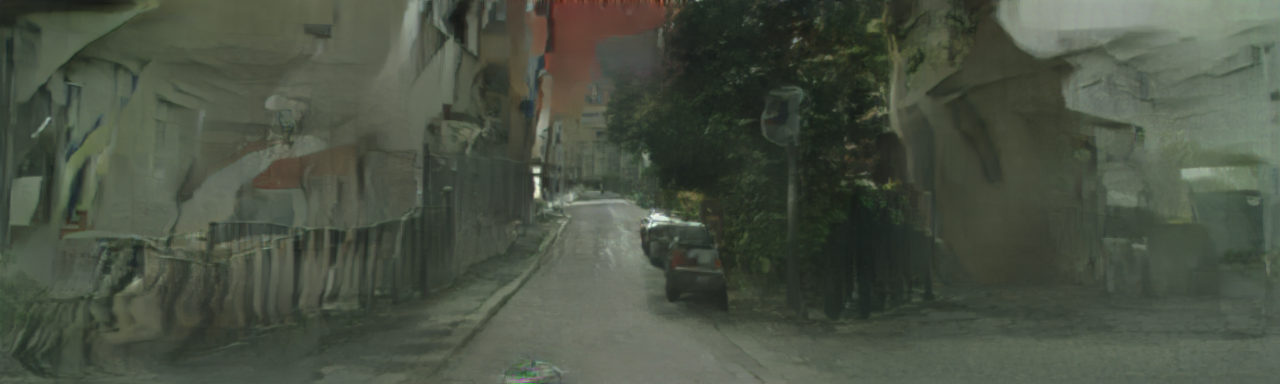}\par 
  \end{multicols}


\vspace{-0.8cm}
\begin{multicols}{3}
\includegraphics[width=1\linewidth]{./images/qualitative/test_set/real/realrgb/seq15_000401.png}\par 
\includegraphics[width=1\linewidth]{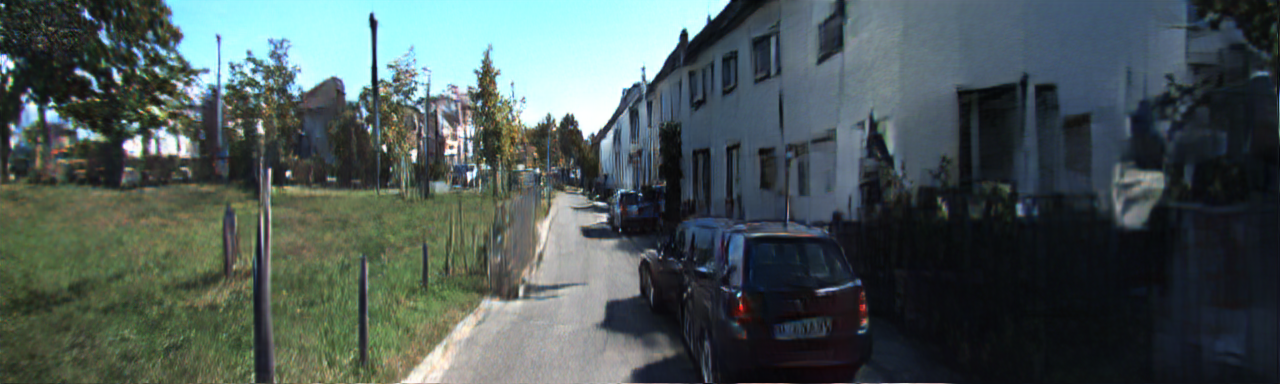}\par     
\includegraphics[width=1\linewidth]{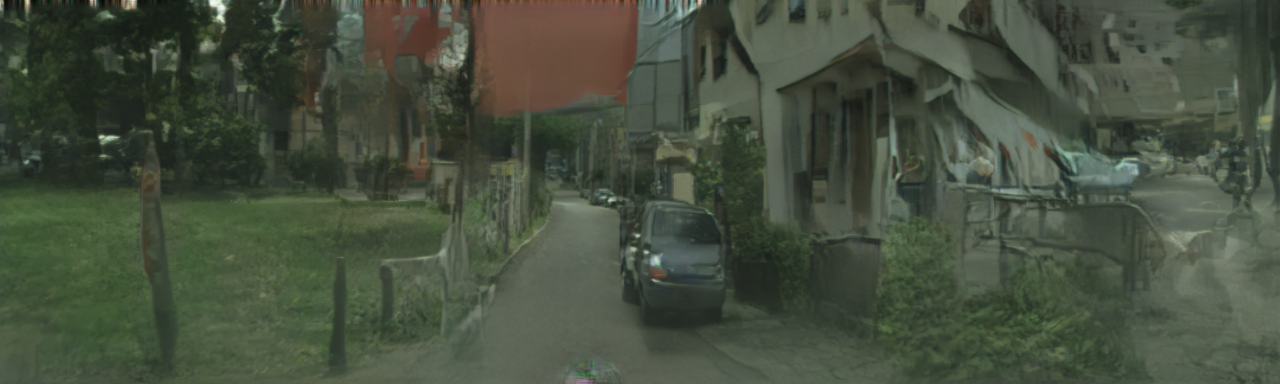}\par 
\end{multicols}

\vspace{-0.8cm}
\begin{multicols}{3}
\includegraphics[width=1\linewidth]{./images/qualitative/test_set/real/realrgb/seq15_000694.png}\par 
\includegraphics[width=1\linewidth]{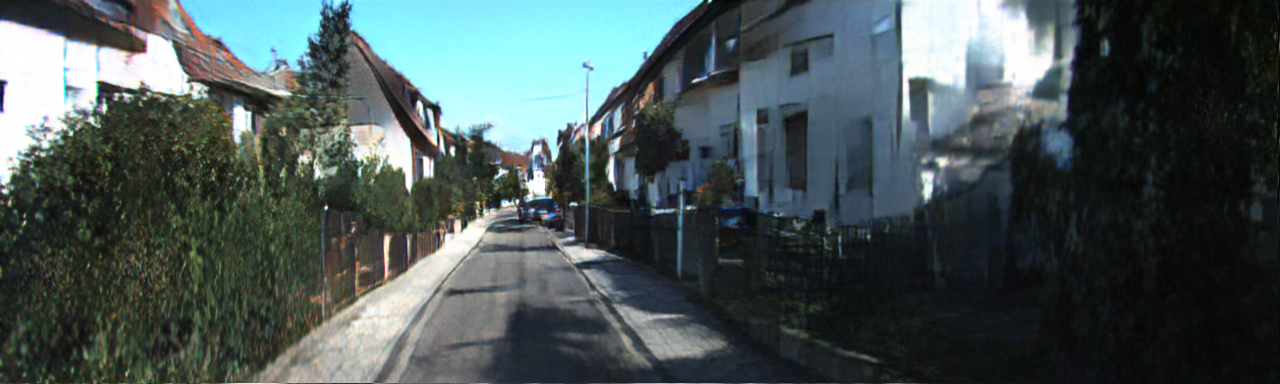}\par     
\includegraphics[width=1\linewidth]{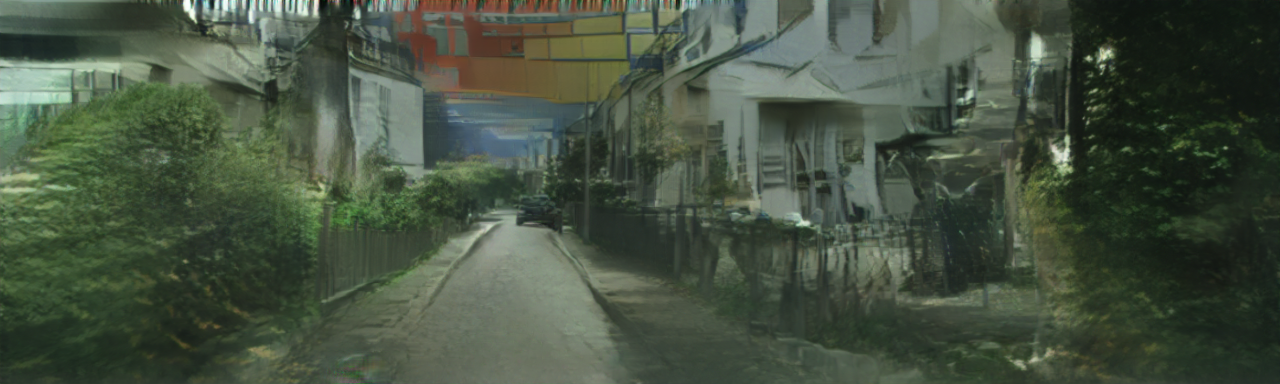}\par 
\end{multicols}

\vspace{-0.8cm}
\begin{multicols}{3}
\includegraphics[width=1\linewidth]{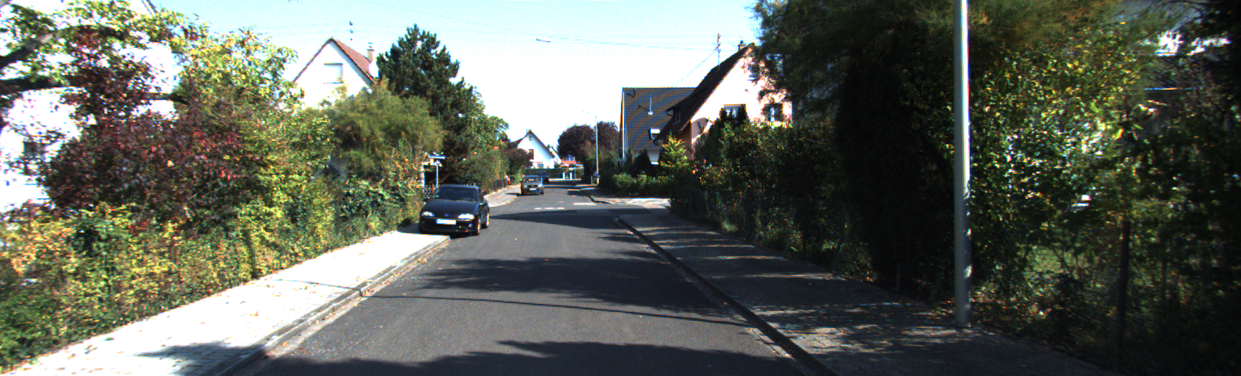}\par 
\includegraphics[width=1\linewidth]{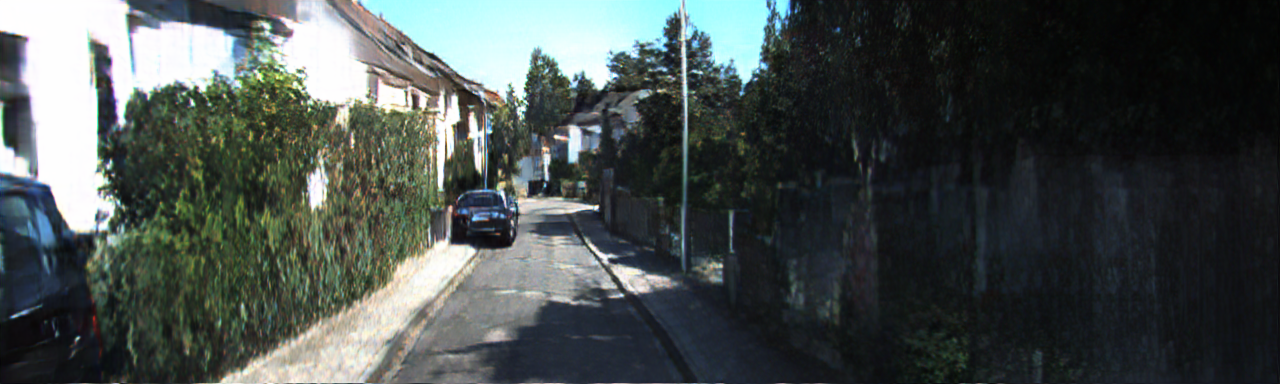}\par     
\includegraphics[width=1\linewidth]{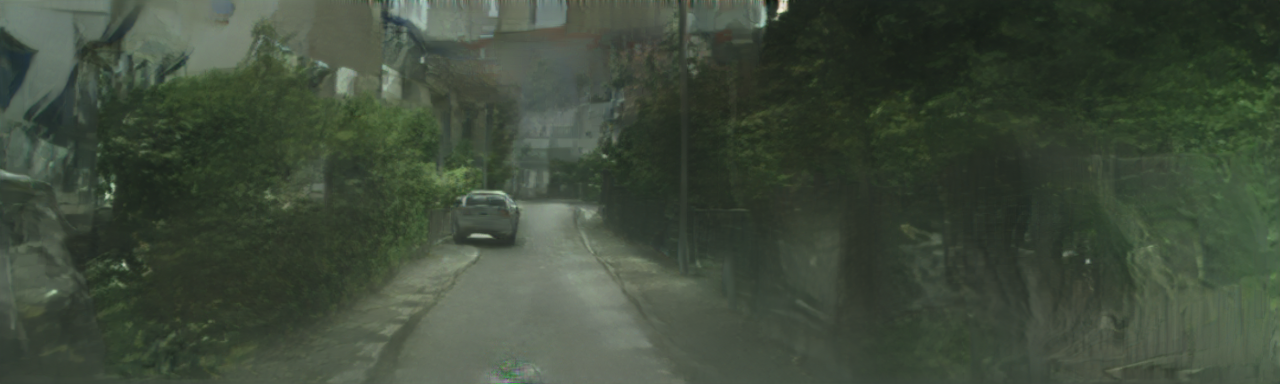}\par 
\end{multicols}

\vspace{-0.8cm}
\begin{multicols}{3}
\includegraphics[width=1\linewidth]{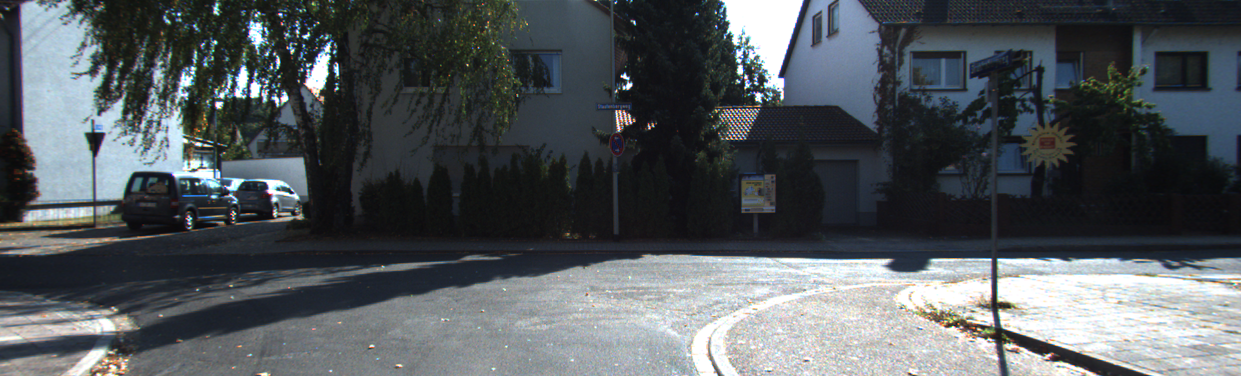}\par 
\includegraphics[width=1\linewidth]{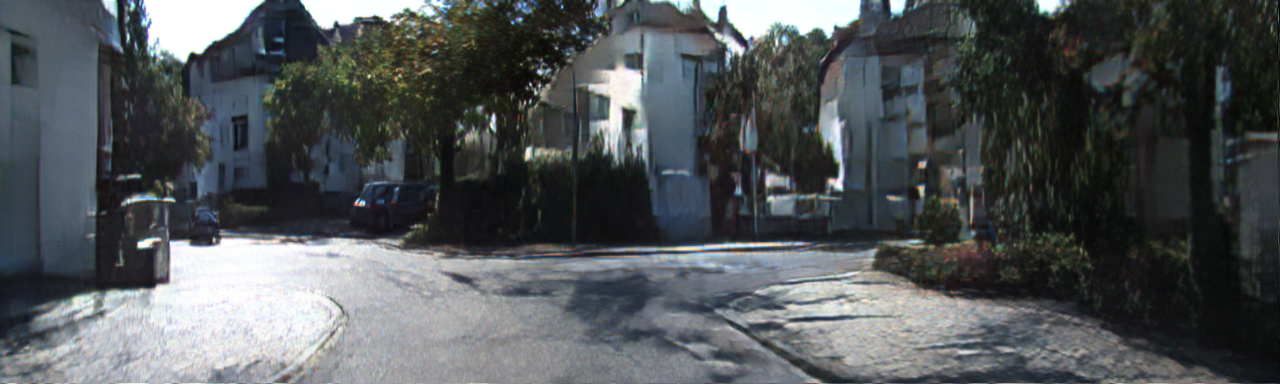}\par     
\includegraphics[width=1\linewidth]{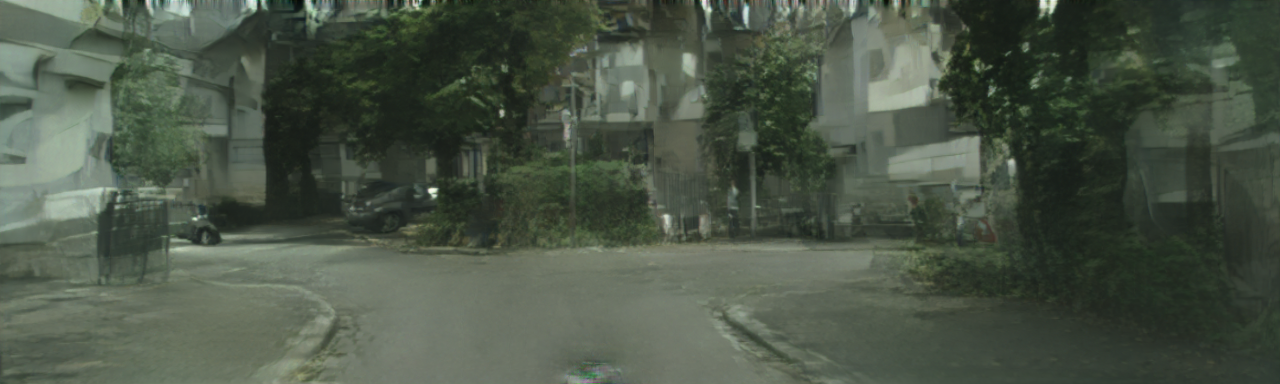}\par 
\end{multicols}

\vspace{-0.8cm}
\begin{multicols}{3}
\includegraphics[width=1\linewidth]{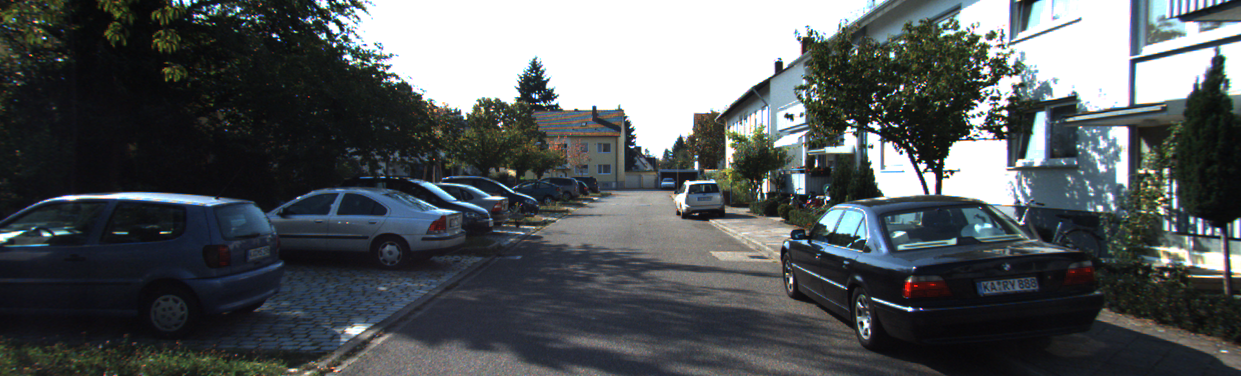}\par 
\includegraphics[width=1\linewidth]{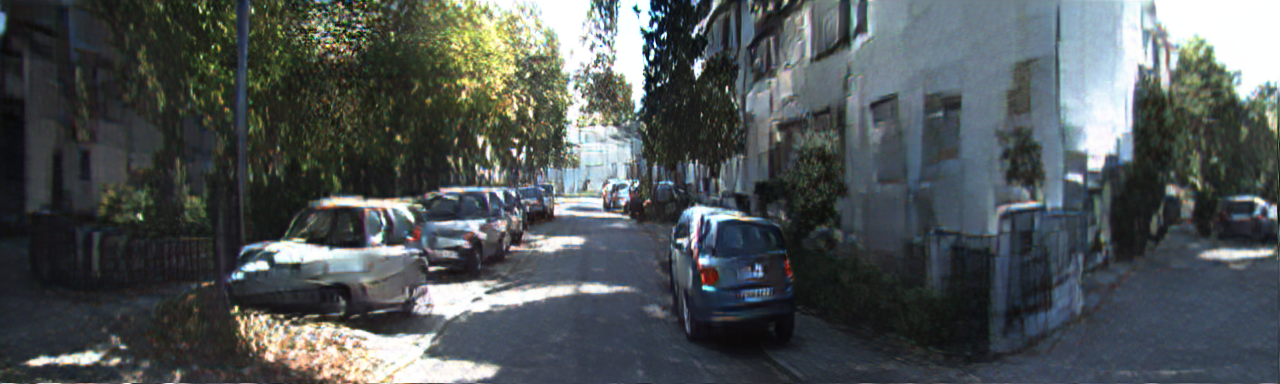}\par     
\includegraphics[width=1\linewidth]{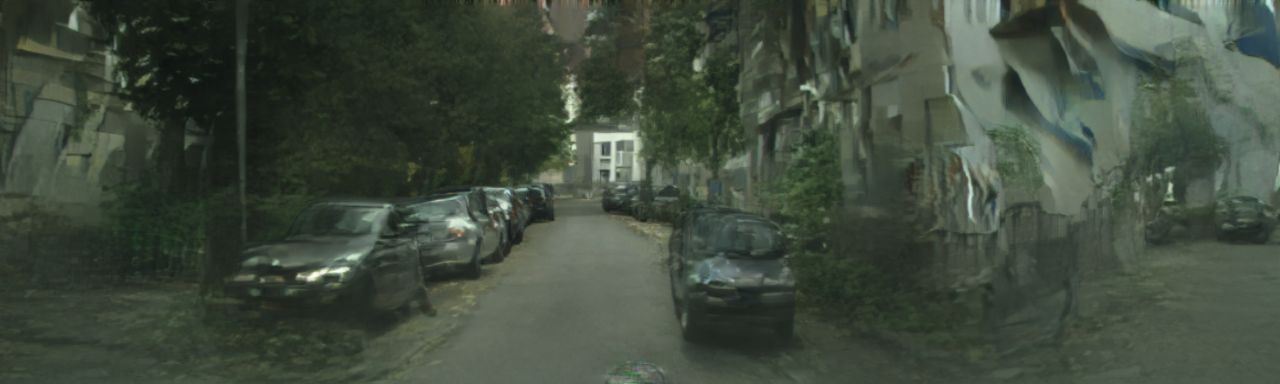}\par 
\end{multicols}

\caption{Different variants generated by our framework on the test set. From left to right, they are the ground truth camera image,  generated image by TITAN-Net $\rightarrow$ Vid2Vid trained on the \sk dataset, and synthesized image by TITAN-Net $\rightarrow$ Vid2Vid trained on     Cityscapes, respectively.}
\label{fig:city}
\end{figure*}

\end{document}